\pdfoutput=1

\documentclass{article} 
\newif\ifpreprintstyle\preprintstyletrue
\ifpreprintstyle
  \PassOptionsToPackage{preprint}{corl_2026}
\else
  \PassOptionsToPackage{final}{corl_2026}
\fi
\usepackage{corl_2026}
\usepackage{adjustbox}
\usepackage{enumitem}
\usepackage{booktabs}
\usepackage{wrapfig}
\usepackage{subcaption}
\usepackage{graphicx}
\usepackage{caption}
\usepackage{makecell}
\usepackage{titletoc} %

\usepackage[utf8x]{inputenc}
\usepackage{amssymb}
\usepackage{booktabs} 
\usepackage{amsmath}
\usepackage{placeins}
\usepackage{longtable}
\usepackage{array}
\usepackage{float}

\usepackage{graphics, graphicx}
\usepackage{natbib}
\usepackage{color}
\usepackage{algorithm}
\usepackage{comment}
\usepackage{hyperref}
\usepackage[symbol*]{footmisc}
\usepackage{xcolor}
\usepackage{caption}
\usepackage{subcaption}
\usepackage{tikz}
\usepackage{algpseudocode}
\usepackage{adjustbox}
\usetikzlibrary{arrows.meta}
\usepackage{booktabs}
\usepackage{bm}

\usepackage{hyperref}
\usepackage{url}
\usepackage{xcolor}

\usepackage{listings}
\definecolor{promptbg}{RGB}{246,248,251}
\definecolor{promptframe}{RGB}{170,180,198}
\definecolor{promptarrow}{RGB}{150,160,178}
\DeclareTextSymbol{\textquotedbl}{OT1}{34}
\lstdefinestyle{promptstyle}{
  basicstyle=\ttfamily\scriptsize,
  backgroundcolor=\color{promptbg},
  frame=single,
  framerule=0.4pt,
  rulecolor=\color{promptframe},
  framesep=5pt,
  xleftmargin=12pt,
  xrightmargin=4pt,
  breaklines=true,
  breakindent=0pt,
  breakautoindent=false,
  postbreak=\mbox{\textcolor{promptarrow}{$\hookrightarrow$}\space},
  columns=fullflexible,
  keepspaces=true,
  showstringspaces=false,
  upquote=true,
  aboveskip=6pt,
  belowskip=4pt,
}
\lstnewenvironment{promptbox}[1][]{\lstset{style=promptstyle, title={\small\textbf{#1}}}}{}

\newif\ifappendix\appendixtrue %
\definecolor{gaingreen}{RGB}{0,140,0}
\newcommand{\up}[1]{{\scriptsize\textcolor{gaingreen}{(+#1)}}}
\newcommand{\dn}[1]{{\scriptsize\textcolor{red}{($-$#1)}}}
\newcommand{\best}[1]{\textbf{#1}} %

\newcommand{\model}{\text{SkillWeaver}}

\newcommand*{\etc}{%
    \@ifnextchar{.}%
        {\textit{etc}}%
        {\textit{etc.}\@\xspace}%
}

\usepackage{tikz}
\usepackage[scaled=0.95]{helvet}
\definecolor{cardbg}{HTML}{F5F6F7}
\definecolor{gradA}{HTML}{2DD4BF}
\definecolor{gradB}{HTML}{7C83F7}
\definecolor{gradC}{HTML}{FFB020}
\definecolor{inkdark}{HTML}{141414}
\definecolor{tgradA}{HTML}{0FA396}  %
\definecolor{tgradB}{HTML}{5B62E8}  %
\definecolor{tgradC}{HTML}{C07A06}  %
\definecolor{affilgray}{HTML}{6E7277}
\definecolor{linkblue}{HTML}{3B5BDB}
\newcommand{\gradname}{%
  \textcolor{tgradA}{S}\textcolor{tgradA!80!tgradB}{k}\textcolor{tgradA!60!tgradB}{i}%
  \textcolor{tgradA!40!tgradB}{l}\textcolor{tgradA!20!tgradB}{l}\textcolor{tgradB}{W}%
  \textcolor{tgradB!80!tgradC}{e}\textcolor{tgradB!60!tgradC}{a}\textcolor{tgradB!40!tgradC}{v}%
  \textcolor{tgradB!20!tgradC}{e}\textcolor{tgradC}{r}}
\newcommand{\paperkeywords}{Agentic Robot Data Generation, Embodied Exploration, Neural Interaction Skills}
\newcommand{\papertitlebody}{Agentic Exploration over Neural Interaction Skills for Scalable Robot Data Generation}
\newif\iffirstmodel
\newcommand{\gradrule}[1]{\textcolor{gradB}{\rule{#1}{1.1pt}}}
\newlength{\cardinner}
\newlength{\cardwidth}
\newlength{\cardtext}

\ifpreprintstyle
\usepackage{fancyhdr}
\renewcommand{\headrule}{\nointerlineskip\vskip3pt\hbox to\headwidth{\tikz[baseline=0pt]\shade[left color=gradA,right color=gradC,middle color=gradB] (0,0) rectangle (\headwidth,0.5pt);}\vskip-3.5pt}
\AtBeginDocument{%
  \setlength{\headwidth}{\textwidth}%
  \addtolength{\topmargin}{-6pt}\addtolength{\headheight}{6pt}}
\fi

\title{\papertitlebody}

\author{
He Zhu$^{1}$ \quad Lusen Zhao$^{2}$ \quad Kwan Man Cheng$^{1}$ \quad Su Li$^{1}$ \quad Katerina Fragkiadaki$^{1}$ \\[3pt]
{\normalfont $^{1}$Carnegie Mellon University \qquad $^{2}$University of Illinois Urbana-Champaign} \\[2pt]
{\normalfont \texttt{\{hez2,ncheng2,suli\}@andrew.cmu.edu} \quad \texttt{lusenz2@illinois.edu} \quad \texttt{katef@cs.cmu.edu}}
}

\begin{document}

\ifpreprintstyle\thispagestyle{empty}\else\maketitle\fi
\addtocontents{toc}{\protect\setcounter{tocdepth}{-5}}

\ifpreprintstyle%
\noindent\hspace*{\dimexpr(\textwidth-\cardwidth)/2\relax}\begin{tikzpicture}
\node[fill=cardbg,rounded corners=9mm,inner sep=\cardinner,text width=\cardtext,align=left]{%
  \setlength{\parindent}{0pt}%
  \firstmodeltrue\renewcommand{\model}{\iffirstmodel\gradname\global\firstmodelfalse\else\text{SkillWeaver}\fi}%
  {\sffamily\bfseries\fontsize{17.5}{20.5}\selectfont\color{inkdark}\gradname: \papertitlebody\par}
  \vspace{9pt}
  {\sffamily\bfseries\fontsize{10.5}{13}\selectfont\color{inkdark}
   He Zhu\textsuperscript{1} \quad Lusen Zhao\textsuperscript{2} \quad
   Kwan Man Cheng\textsuperscript{1} \quad Su Li\textsuperscript{1} \quad Katerina Fragkiadaki\textsuperscript{1}\par}
  \vspace{2pt}
  {\sffamily\fontsize{9}{12}\selectfont\color{affilgray}
   \textsuperscript{1}Carnegie Mellon University \quad
   \textsuperscript{2}University of Illinois Urbana-Champaign\par}
  \vspace{5pt}
  \gradrule{\cardtext}\par
  \vspace{4pt}
  {\fontsize{9.6}{13.4}\selectfont\color{inkdark}
Large-scale demonstrations have driven unprecedented progress in robot learning, 
yet collecting robot data through teleoperation is expensive and difficult to scale to diverse environments and long-horizon tasks. 
Simulation offers a scalable alternative, 
but existing data-generation pipelines often rely on open-loop controllers, scripted skill sequences, or task-specific programs. 
We introduce \model{}, an agentic framework that autonomously generates robot experience by exploring over Neural Interaction Skills (NIS): 
reusable, parameterized, closed-loop policies that expose learned physical interaction capabilities to a reasoning agent. 
Given a task and a simulated environment, a VLM agent reasons about what to do next, 
invokes and parameterizes NIS to interact with the environment, 
observes their outcomes, 
and generates verification, reflection, and memory to guide subsequent exploration.
We instantiate NIS as reinforcement-learned policies for closed-loop, contact-rich manipulation and organize exploration as verifier-guided tree search, 
enabling the agent to discover successful long-horizon behaviors without relying on predetermined execution pipelines. 
\model{} scales autonomously to 39.1K demonstrations across 14.1K scenes, which we distill into visuomotor policies. 
Across simulation benchmarks and real-world manipulation, 
training on \model{}-generated experience substantially improves generalization to novel objects, spatial configurations, tasks, and environments, 
and enables zero- and few-shot sim-to-sim and sim-to-real transfer. 
Our results suggest \textbf{agentic exploration} over \textbf{neural interaction skills} as a scalable alternative for robot data generation.
\par}
  \vspace{5pt}
  {\sffamily\fontsize{9.2}{12}\selectfont\color{inkdark}\textbf{Keywords:}\quad\paperkeywords\par}
  \vspace{4pt}
  {\sffamily\fontsize{9.2}{12}\selectfont\textbf{Website:}\quad
   \textcolor{linkblue}{\texttt{\href{https://katefgroup.github.io/skillweaver/}{https://katefgroup.github.io/skillweaver/}}}\par}
};
\end{tikzpicture}\par\else\begin{abstract}

Robot learning increasingly relies on large-scale demonstrations, yet collecting robot data through teleoperation is expensive and difficult to scale to diverse environments and long-horizon tasks. Simulation offers a scalable alternative, but existing data-generation pipelines often rely on scripted controllers, fixed skill sequences, or task-specific programs. We introduce \model{}, an agentic framework that autonomously generates robot experience by exploring over Neural Interaction Skills (NIS): reusable, parameterized, closed-loop policies that expose learned physical interaction capabilities to a reasoning agent. Given a task and a simulated environment, a VLM agent reasons about what to do next, invokes and parameterizes NIS to interact with the environment, observes their outcomes, and uses  verification, reflection, and memory to guide subsequent exploration. We instantiate NIS as reinforcement-learned policies for contact-rich manipulation and organize exploration as verifier-guided tree search, enabling the agent to discover successful long-horizon behaviors rather than relying on predetermined execution pipelines. \model{} scales autonomously to 39.1K demonstrations across 14.1K scenes (11.0M frames), which we distill into visuomotor policies. Across simulation benchmarks and real-world manipulation, training on \model{}-generated experience substantially improves generalization to novel objects, spatial configurations, tasks, and environments, and enables zero- and few-shot sim-to-sim and sim-to-real transfer. Our results suggest \textbf{agentic exploration over learned physical tools as a scalable alternative to human demonstration collection for robot learning.}

\end{abstract}\fi

\ifpreprintstyle\else\fi 

\begin{center}
    \includegraphics[width=1.0\textwidth]{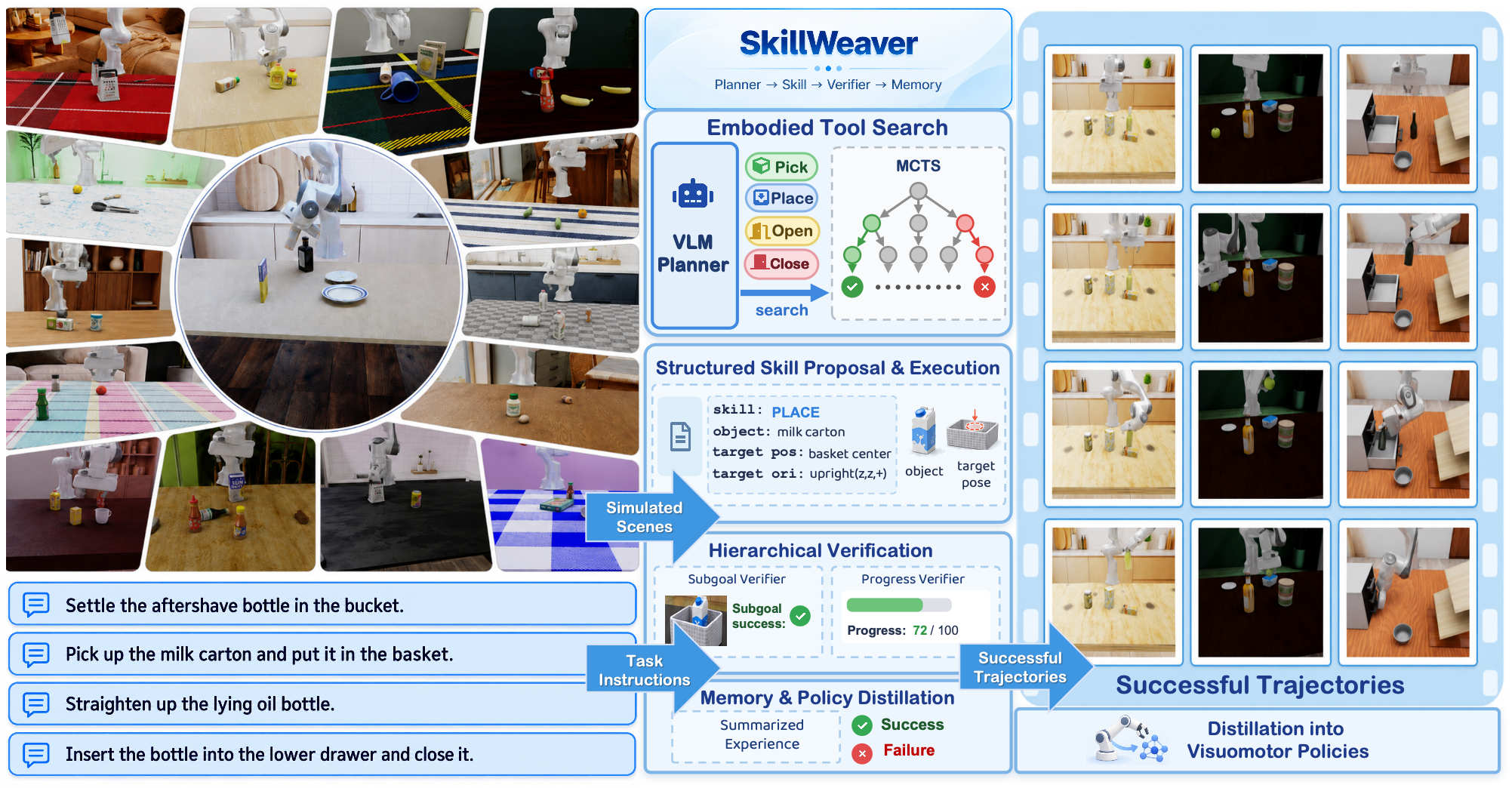}
    \captionof{figure}{We present \textbf{\model{}}, an agentic robot data engine that generates large-scale demonstrations in simulation.
    Given a simulated scene and a robot manipulation task, the agent composes 
    reusable neural interaction skills with verifier-guided embodied search,
    and retains successful experience as demonstrations.
    The generated demonstrations are distilled into visuomotor policies such as Vision-Language-Action models.
    }
    \label{fig:teaser}
\end{center}

\section{Introduction}

Robot manipulation has recently seen remarkable progress driven by imitation learning from
large-scale demonstrations collected through teleoperation~\cite{
chi2023diffusion,rt2,kim2024openvla,octo2024,black2024pi0,pi05}.
Despite these advances, robot data remains a fundamental bottleneck.
Collecting high-quality demonstrations is expensive, labor-intensive, and time-consuming,
requiring robot hardware, trained operators, environment resets, and substantial engineering effort.
More importantly, scaling human demonstration collection to cover the long tail of objects,
scene configurations, and long-horizon behaviors needed for broad generalization remains
challenging even for well-resourced laboratories.

Simulation offers an appealing alternative: once an interactive environment is available,
robot experience can in principle be generated at much larger scale and with richer supervision
than is practical in the real world~\cite{tian2025interndata,deshpande2026molmobot,genie3sim}.
Yet generating environments is only part of the problem.
Generating \emph{behavior} in those environments still commonly relies on engineered expert
pipelines---including scripted skill sequences, open-loop grasp proposals~\cite{murali2025graspgen,fang2023anygrasp} and motion planners~\cite{sundaralingam2023curobo},
or hand-designed waypoints.
Such pipelines can generate data efficiently for behaviors anticipated by their designers,
but are difficult to scale toward diverse, long-horizon tasks whose successful execution may
require reasoning about what to do next, recovering from failures, and discovering intermediate
steps through interaction.

At the same time, foundation models have enabled a new paradigm of \emph{agentic} problem
solving. LLMs and VLMs can reason over available tools, invoke them, observe their outcomes,
and iteratively revise their decisions. In robotics, this has enabled semantic skill
composition~\cite{ahn2022saycan} and agents that generate and debug executable robot
programs~\cite{liang2023code,zhang2026playful,lu2026aspire}.
However, software tools and generated programs can be limiting as interfaces for contact-rich physical interaction, 
where successful execution requires continuous closed-loop adaptation to the evolving physical state.

We ask whether the tool-use paradigm of reasoning agents can instead be extended to
\emph{learned physical interaction}.
To this end, we introduce \textbf{Neural Interaction Skills (NIS)}:
reusable, parameterized, closed-loop policies that expose learned physical capabilities to
a reasoning agent.
Rather than directly predicting low-level robot actions or generating task-specific control
programs, an agent can reason about \emph{which} interaction tool to invoke, parameterize \emph{where and
how} to apply it, execute the tool in the environment, observe the resulting state, and use
this feedback to decide the next step.
This abstraction combines the semantic flexibility of foundation-model agents with robust
learned physical interactions.

Building on this idea, we present \model{}, an agentic framework for scalable robot data
generation through exploration over Neural Interaction Skills.
Given a task instruction and a simulated environment, a VLM agent reasons over the current
scene and invokes parameterized NIS to manipulate it.
We instantiate NIS as model-free RL policies for contact-rich PICK, PLACE, OPEN, and CLOSE
behaviors, while retaining motion planning for collision-free free-space motion.
Rather than committing to a single predicted sequence, \model{} treats demonstration
generation as an embodied search problem and explores alternative compositions and
parameterizations of its tools using tree search.
A hierarchical verifier evaluates the physical consequences of each interaction, providing
intermediate rewards and reflective feedback for re-planning, while semantic memory distills
successful strategies and failure modes to guide future exploration.
Successful interaction traces are automatically verified and retained as robot demonstrations.

This turns simulation from a passive source of randomized training examples into an
environment in which an agent can actively \emph{search for its own training experience}.
Combined with scalable environment generation, \model{} produces \textbf{3.4K} frames of
multimodal embodied data per GPU hour and scales to 14.1K simulated scenes and 39.1K
manipulation demonstrations (11.0M frames).
We distill the generated experience into downstream visuomotor policies and evaluate them
across simulation benchmarks and real-world Franka manipulation tasks.
Training on \model{} data enables robust zero-shot sim-to-sim and sim-to-real
transfer and substantially improves generalization to novel objects, spatial configurations,
tasks, and environments.

Our contributions are threefold:
(1) We introduce Neural Interaction Skills, an interface that exposes learned
closed-loop physical interaction policies as parameterized tools to a reasoning agent, and
formulate scalable robot data generation as agentic exploration over these tools.
(2) We introduce \model{}, which combines VLM reasoning, verifier-guided tree search,
reflection and semantic memory with learned interaction tools and scalable simulation to
autonomously discover and verify long-horizon robot demonstrations.
(3) We generate 39.1K demonstrations across 14.1K scenes and show that the resulting data
substantially improves downstream visuomotor policy generalization and enables zero-shot transfer across simulators and to the real world, 
without additional teleoperation.

\section{Related Work}

\paragraph{Real-world Robot Datasets.}
While multiple industries pursue learning-from-demonstration (LfD) for robot training, 
such data remain largely proprietary due to the immense resources required for collection.
Open-sourced datasets such as DROID~\cite{khazatsky2024droid}, RoboNet~\cite{dasari2020robonet}, RH20T~\cite{fang2024rh20t}, Open X-Embodiment~\cite{openx2024}, and Agibot World~\cite{bu2025agibot} span diverse robots, tasks, and embodiments, but real demonstrations remain costly to scale.
Collecting such data typically requires robot hardware, human operators, repeated environment resets, and substantial engineering effort for calibration and maintenance.
These costs make it difficult to systematically cover the long tail of object configurations, scene variations, and long-horizon behaviors needed for broad policy generalization, and motivate complementary approaches that shift part of the data-generation burden to simulation.

\paragraph{Simulation-based Data Generation.}
Simulation provides controllable variation in tasks, scenes, and object configurations~\cite{gu2023maniskill2,james2020rlbench,mees2022calvin}, but generating successful robot behavior within these environments often still requires substantial human engineering.
Early systems rely on teleoperation or manually specified task programs~\cite{rlbench,chen2025robotwin2}, while recent work increasingly automates different stages of the data-generation process.
Gen2Sim~\cite{gen2sim}, RoboGen~\cite{robogen}, and GenSim~\cite{gensim} use foundation models to generate tasks, supervision, or executable programs.
RoboCasa~\cite{nasiriany2024robocasa} and MimicGen~\cite{mimicgen} scale demonstrations by adapting a small set of human demonstrations to new scene configurations.
More recent systems, including InternData-A1~\cite{tian2025interndata}, Genie Sim 3.0~\cite{genie3sim}, and MolmoBot~\cite{deshpande2026molmobot}, automate demonstration generation at substantially larger scales using engineered motion-planning and control pipelines, demonstrating the value of synthetic experience for downstream robot learning.
These approaches substantially reduce the cost of data collection, but behavior generation generally remains specified through source demonstrations, generated programs, or expert execution pipelines.
\model{} instead treats behavior generation itself as an embodied search problem, autonomously discovering long-horizon demonstrations through verifier-guided exploration over reusable learned interaction skills.

\paragraph{Foundation Models and Agentic Tool Use for Robotics.}
Foundation models have been used throughout the robot decision-making hierarchy.
At the task level, SayCan~\cite{ahn2022saycan} selects among predefined robot
skills, while TidyBot~\cite{wu2023tidybot} uses language models for semantic
household decisions.
Other approaches use foundation models to produce lower-level geometric
representations: LLMTrajGen~\cite{kwon2024llmtrajgen} predicts end-effector
trajectories, while CoPA~\cite{huang2024copa} predicts spatial constraints from
which manipulation poses are derived.
These approaches provide flexible semantic reasoning, but typically either
directly predict robot behavior or reason over a fixed set of engineered
robot capabilities.
A complementary line of work treats foundation models as agents that solve
problems by invoking tools, observing their outcomes, and iteratively revising
their decisions.
In robotics, Code as Policies~\cite{liang2023code} generates executable programs
that compose perception and control APIs, while recent systems such as
CaP-X~\cite{fu2026capx}, RATs~\cite{zhang2026playful}, and
ASPIRE~\cite{lu2026aspire} close the loop by allowing coding agents to execute,
verify, revise, or improve robot programs through interaction.
These works demonstrate the power of agentic reasoning, but primarily expose
robot capabilities through code, APIs, or generated programs.
Compared with program-based robot agents, 
\model{} offers a more scalable paradigm by 
extending the agentic tool-use paradigm from software tools to
\emph{parameterized, learned physical interaction}. 
These learned skills provide closed-loop control, 
while verifier-guided tree search enables more extensive exploration over alternative behaviors and skill parameterizations.

\section{\model{}}
\label{sec:result}

\begin{figure}[t]
    \centering
    \includegraphics[width=0.95\textwidth]{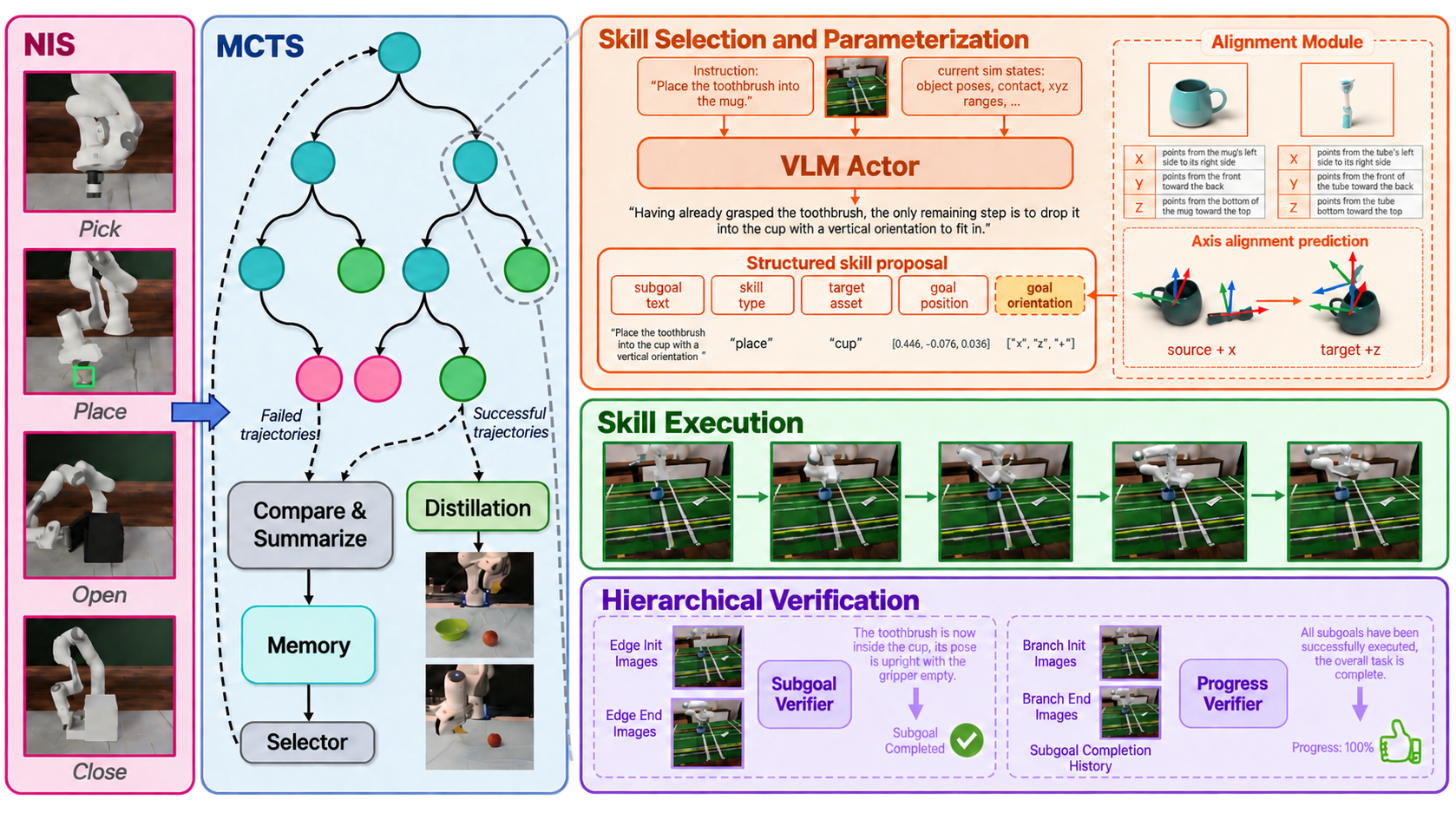}
 \caption{\textbf{Overview of \model{}.}
    \model{} generates robot experience through agentic exploration over
    Neural Interaction Skills (NIS).
    At each exploration step, a VLM agent
    selects and parameterizes an NIS, executes it in the environment, and
    observes the resulting state.
    A hierarchical verifier evaluates the physical outcome and provides
    progress rewards and reflection, while semantic memory transfers
    experience across search episodes.
    Verifier-guided tree search explores alternative interaction strategies,
    and successful trajectories are retained as multimodal robot training data.}
    \label{fig:skillweaver}
\end{figure}

We formulate autonomous robot data generation as \emph{agentic exploration
over learned physical interaction} (Fig.~\ref{fig:skillweaver}).
Given a task instruction and a simulated environment
(Sec.~\ref{sec:env_gen}), \model{} interacts with the environment through
a collection of Neural Interaction Skills (NIS)
(Sec.~\ref{sec:rl_train}).
At each step, a VLM agent reasons about the current world state, selects
a tool and its parameters, executes it, and observes its physical outcome.
Instead of predicting a single behavior sequence per episode, 
\model{} organizes these interactions into a search tree and explores alternative
strategies using verifier feedback, reflection, and memories distilled
from previous experience (Sec.~\ref{sec:mcts_agent}).
Successful interaction traces are automatically verified and transformed
into richly annotated robot training examples
(Sec.~\ref{sec:dataset}).

\subsection{Interactive Simulation Environments for Exploration}
\label{sec:env_gen}

Agentic data generation requires diverse, physically interactive environments
in which the agent can explore alternative manipulation strategies.
We therefore construct an automated simulation pipeline that generates
large-scale interactive scenes with diverse assets, layouts, and object
configurations.

We first build a simulatable asset database. Gemini~\cite{gemini2025} proposes common indoor object categories and synthesizes diverse object images, which are reconstructed into textured meshes using SAM3D-Objects~\cite{sam3d2025}. CoACD~\cite{wei2022coacd} generates collision geometry, 
and the resulting assets are packaged as IsaacLab-compatible USD objects~\cite{mittal2025isaac}. 
Each asset is further annotated with semantic and physical metadata—including the semantic meaning of canonical axes, approximate size, mass, friction, and so on to support planning.

The environment generator supports two complementary modes of data generation:
broad exploration across diverse synthetic environments and targeted exploration
within reconstructed real-world scenes.  
The \emph{broad exploration} mode procedurally composes diverse scenes by randomly sampling assets and layouts,
making scene generation highly efficient, where a single process generates \textbf{$\sim$340 scenes per minute}.
The \emph{targeted exploration} mode reconstructs environments from seed observations using 3D vision tools~\cite{sam3d2025,geng2025oneview,xu2025pixelperfectdepth}, 
instantiating them as physically interactive digital twins for targeted environments.

Finally, we randomize visual appearance to improve robustness. We perturb lighting, textures, backgrounds, and camera viewpoints, 
encouraging downstream policies to learn behaviorally relevant interaction cues rather than overfitting to specific simulation statistics.

\subsection{Neural Interaction Skills}
\label{sec:rl_train}

To interact with the physical environment, the reasoning agent requires
capabilities that are both reusable across tasks and robust to local variation
in physical state.
We introduce \textbf{Neural Interaction Skills (NIS)}, learned closed-loop
policies exposed to the agent through a structured, parameterized interface.
An NIS implements a reusable physical capability: the agent specifies
\emph{what} interaction to perform and its semantic/geometric parameters,
while the learned policy determines \emph{how} to realize that interaction
through closed-loop control.

This separation is particularly important for contact-rich manipulation.
While motion planners provide reliable collision-free motion through free
space, interactions such as grasp acquisition, stable placement, and
articulated-object manipulation require continuous adaptation to object
geometry, contact, and the evolving physical state.
We therefore use motion planning for free-space positioning and instantiate
NIS for contact-rich interaction using policies learned with model-free RL.

Concretely, we train reusable \textsc{Pick}, \textsc{Place}, \textsc{Open},
and \textsc{Close} NIS with PPO~\cite{schulman2017proximalppo} in
IsaacLab~\cite{mittal2025isaac}.
Each policy receives robot proprioception and privileged object-centric
observations, including target pose, goal residuals, and local
gripper--object geometry features, and outputs Cartesian end-effector deltas
and gripper commands.
Because effective interaction strategies vary across object geometries,
poses, and articulation types, each NIS contains specialized sub-policies
with tailored observations, rewards, and success conditions
\ifappendix; details are provided in Appendix~\ref{app:skill_details}\fi.

NIS training and deployment share the same local interaction interface.
During training, we randomize object poses, joint states, and robot
initialization within a local neighborhood of the target object or part.
At deployment, the agent parameterizes the selected NIS with the target
object and desired interaction outcome.
A motion planner first brings the gripper to a local pre-interaction
configuration, after which the corresponding learned policy performs the
contact-rich interaction.
This shared interface allows the same NIS to be invoked across objects,
layouts, tasks, and search episodes.

\subsection{Agentic Exploration over Neural Interaction Skills}
\label{sec:mcts_agent}

\paragraph{Tool Selection and Grounding.}
At each interaction step, the VLM agent observes the current environment
state and reasons about which physical interaction should be attempted next.
It predicts a structured action
\[
    a_t = (g_t, \tau_t, o_t, \theta_t),
\]
where $g_t$ denotes the semantic subgoal, $\tau_t$ the selected Neural
Interaction Skill, $o_t$ its target object or articulation part, and
$\theta_t$ the tool-specific geometric parameters, such as a target
position, orientation, or joint state.

For orientation-sensitive interactions, directly predicting unconstrained
3D rotations from a VLM is often unstable and geometrically inconsistent.
We instead parameterize orientation through \textbf{semantic axis alignment}.
Given pre-annotated canonical object axes, the agent predicts semantic axis
correspondences, from which the target rotation is computed analytically
(see App.~\ref{app:axis_align}).

Executing a tool invocation produces a physical state transition.
A motion planner~\cite{sundaralingam2023curobo} first moves the robot to a
local pre-interaction configuration near the selected object or target region.
The selected NIS then executes the contact-rich portion of the interaction
conditioned on its grounded parameters.
The resulting simulator state and execution observations are returned to
the agent, closing the loop between reasoning and physical interaction.

\paragraph{Embodied Tool Search.}

A single sequence of tool predictions can fail because of incorrect
high-level decisions, geometric grounding errors, or unexpected physical
outcomes.
Rather than committing to the first predicted sequence, \model{} explicitly
explores alternative tool invocations through interaction within the simulator.

We organize agent--environment interactions into a search tree.
Each node represents a simulator state together with its interaction history,
verifier feedback, and retrieved memory, while each outgoing edge represents
the execution of a parameterized NIS.
Expanding an edge therefore corresponds to an actual physical experiment:
the agent proposes a tool invocation, executes it in simulation, and observes
the resulting world state.

We use Monte Carlo Tree Search (MCTS) to allocate exploration toward promising
branches while preserving alternative strategies.
At each expansion, the VLM proposes candidate tool invocations conditioned
on the current state and previous feedback; the corresponding NIS are
executed, and the resulting states are inserted into the tree.
Successful prefixes can thus be reused while unsuccessful branches trigger
alternative tool choices or parameterizations.

\paragraph{Outcome Verification and Reflection.}

Agentic exploration requires feedback about whether an attempted physical
interaction succeeded and moved the system closer to the overall
task goal.
Standard MCTS estimates newly expanded nodes through downstream rollouts,
which would require repeatedly executing additional physical interactions
before receiving a terminal reward, which is expensive for long-horizon manipulation.
We therefore introduce a hierarchical verifier that directly evaluates the
outcome of each newly expanded node and provides intermediate feedback for
search.

Verification is decomposed into two levels to provide more reliable progress estimation.
A \textit{subgoal verifier} assesses whether the latest skill achieved its local objective, while a \textit{progress verifier} aggregates preceding subgoal outcomes to estimate overall task progress, producing a scalar score $(r_t \in [0,100])$ and a natural-language reflection on failures or corrective actions.
These reflections condition subsequent planner prompts for adaptive re-planning.
Reliable verification also requires accurate spatial and physical reasoning, 
yet VLMs remain brittle with embodied and spatial tasks from visual observations alone~\cite{luo2025robobench,feng2025seeing}.
We therefore complement rendered observations with privileged simulator states, 
including object poses, contacts, and spatial relations, providing explicit cues that may be ambiguous from RGB alone.

Together, these design choices allow the verifier to serve three roles in \model{}: 
it provides intermediate
rewards that prioritize promising regions of the search tree, supplies
natural-language feedback that conditions subsequent reasoning, and determines
which discovered trajectories are successful to enter the
training dataset.

\paragraph{Experience Memory.}

Independent search episodes often encounter recurring physical and semantic
failure modes.
To transfer experience across episodes, \model{} distills completed search
trajectories into compact semantic memories.
Each memory associates a scenario or interaction pattern with a reusable
strategy, failure explanation, or verification rule.

Before proposing a new tool invocation and parameterization, a selector VLM retrieves the top-$K$
memories most relevant to the current task and subgoal.
The retrieved experience is included in the agent context, allowing future
searches to reuse previously discovered strategies and avoid repeatedly
exploring known failure modes. See Appendix~\ref{app:memory_module} for details on the memory design.

\subsection{From Exploration to Robot Training Data}
\label{sec:dataset}

Successful trajectories discovered during exploration are automatically
converted into training examples.
Because the agent interacts with a simulator, each trajectory provides not
only state--action sequences but also synchronized multimodal observations
and privileged semantic and geometric supervision.

For each trajectory, we retain semantic task descriptions, object attributes,
spatial relations, execution traces, and verifier feedback.
We augment language supervision with VLM-generated paraphrases and grounded
object descriptions synthesized from asset metadata.
Privileged simulator state further provides object poses, contacts, support
relations, orientations, and temporal geometric relationships.
RGB, depth, segmentation masks, robot state, object state, contacts, and
language annotations are therefore naturally synchronized without additional
human labeling.

The resulting dataset is used to train deployable visuomotor policies,
distilling behaviors discovered through agentic exploration into a policy
that no longer requires search, privileged state, or access to the NIS
library at deployment.

\section{Experiments}
\label{sec:experiments}

We evaluate \model{} around four questions.
First, can agentic exploration over Neural Interaction Skills (NIS)
autonomously discover successful robot behaviors and generate experience at scale?
Second, does the resulting experience improve downstream visuomotor policy
generalization and transfer across simulated environments?
Third, which components of \model{}---learned closed-loop NIS, branching search,
hierarchical verification, and experience memory---are important for efficient
and reliable behavior discovery?
Finally, can policies trained on \model{}-generated simulation experience
transfer to real-world manipulation without any real-world teleoperation data?

\subsection{Agentic Exploration and Data Generation at Scale}

We first evaluate whether \model{} can autonomously discover successful
behaviors across diverse manipulation problems and generate robot experience
at scale.

For broad \emph{in-the-wild} data generation, we consider five task families
requiring increasingly rich physical interaction or reasoning complexity.
\emph{Pick} tasks require grasping and lifting an object, while
\emph{Place} additionally requires transporting it to a specified target.
\emph{Orientation-aware} tasks require reasoning about the desired final
object pose, such as straightening a lying object or inserting an elongated
object into a narrow container.
\emph{Articulation} tasks require opening or closing prismatic and revolute
parts such as drawers and cabinet doors.
Finally, \emph{Long-horizon} tasks compose multiple NIS and may require
discovering prerequisite interactions that are not explicitly stated in the
instruction.
Appendix~\ref{app:task_design} provides detailed task definitions, example
instructions, and interaction sequences.

In addition to in-the-wild exploration, we use \model{} to generate
demonstrations for benchmark-derived tasks from
LIBERO-PRO~\cite{zhou2025liberopro} and
SIMPLER-WidowX~\cite{SIMPLER} that can be expressed through the current
NIS library, as well as task-specific data for reconstructed real-world
environments.

Table~\ref{tab:search-data-stats} summarizes both the behavior-discovery
process and the resulting dataset.
Across task families, \model{} discovers successful trajectories in
$81.5\%$ of search runs overall.
Long-horizon tasks require substantially deeper exploration and remain
the most challenging, with $59.9\%$ search success and an average successful
trajectory depth of $4.29$.
Our data generation is highly
parallelizable, with throughput scaling linearly with the increase of compute. 
On 8 NVIDIA A6000 GPUs, \model{} produces
$14.95\times24\times8\approx2.9$K successful trajectories per day.
In total, \model{} generates 39.1K verified demonstrations across 14.1K
simulated scenes, comprising 11.0M frames or 153.1 robot hours.
Per-task search success rates for benchmark-derived tasks are reported in
App.~\ref{app:tree_search_stats}.

\newcommand{\statstbd}{{\color{red}--}}
\begin{table}[!h]
\centering
\small

\textbf{(a) Tree search statistics.}
\vspace{2pt}

\setlength{\tabcolsep}{3pt}
\begin{tabular}{lcccccc}
\toprule
Task & Breadth & Depth & Expansions & Success (\%) & Trajectories/GPU-hour & Frames/GPU-hour \\
\midrule
Pick              & 2 & 1.00 & 1.00 & 84.2 & 27.24 & 3.7K \\
Place             & 2 & 2.00 & 2.37 & 77.8 & 11.07 & 3.0K \\
Orientation-aware & 2 & 2.13 & 2.29 & 81.5 & 12.52 & 3.6K \\
Articulation      & 2 & 1.00 & 1.16 & 90.0 & 22.15 & 5.4K \\
Long-horizon      & 2 & 4.29 & 4.62 & 59.9 & 5.58 & 3.2K \\
\midrule
Overall           & 2 & 1.64 & 1.79 & 81.5 & 14.95 & 3.4K \\
\bottomrule
\end{tabular}

\vspace{6pt}
\textbf{(b) Data statistics.}
\vspace{2pt}

\setlength{\tabcolsep}{6pt}
\begin{tabular}{lcccc}
\toprule
Source & \#Scenes & \#Trajectories & \#Frames & Robot Hours \\
\midrule
In-the-wild              & 9.4K & 16.9K & 3.7M & 50.9 \\
LIBERO-PRO               & 4.3K & 19.5K & 6.8M & 94.7 \\
SIMPLER                  & 0.2K & 1.9K  & 0.4M & 5.6 \\
Real-world post-training & 0.2K & 0.8K  & 0.1M & 1.8 \\
\midrule
Total                    & 14.1K & 39.1K & 11.0M & 153.1 \\
\bottomrule
\end{tabular}
\caption{
\textbf{Agentic exploration and generated data statistics.}
(a) Search statistics and generation throughput across task families:
tree breadth, average successful-trajectory depth, expansions per successful
trajectory, search success rate, and successful trajectories and frames
generated per GPU-hour.
(b) Scale of the resulting verified dataset across in-the-wild exploration,
benchmark-derived tasks, and reconstructed real-world environments.
}
\label{tab:search-data-stats}
\end{table}

\subsection{Learning from \model{}-Generated Experience}

\begin{wrapfigure}[16]{R}{0.45\textwidth}
\centering
\includegraphics[width=0.45\textwidth]{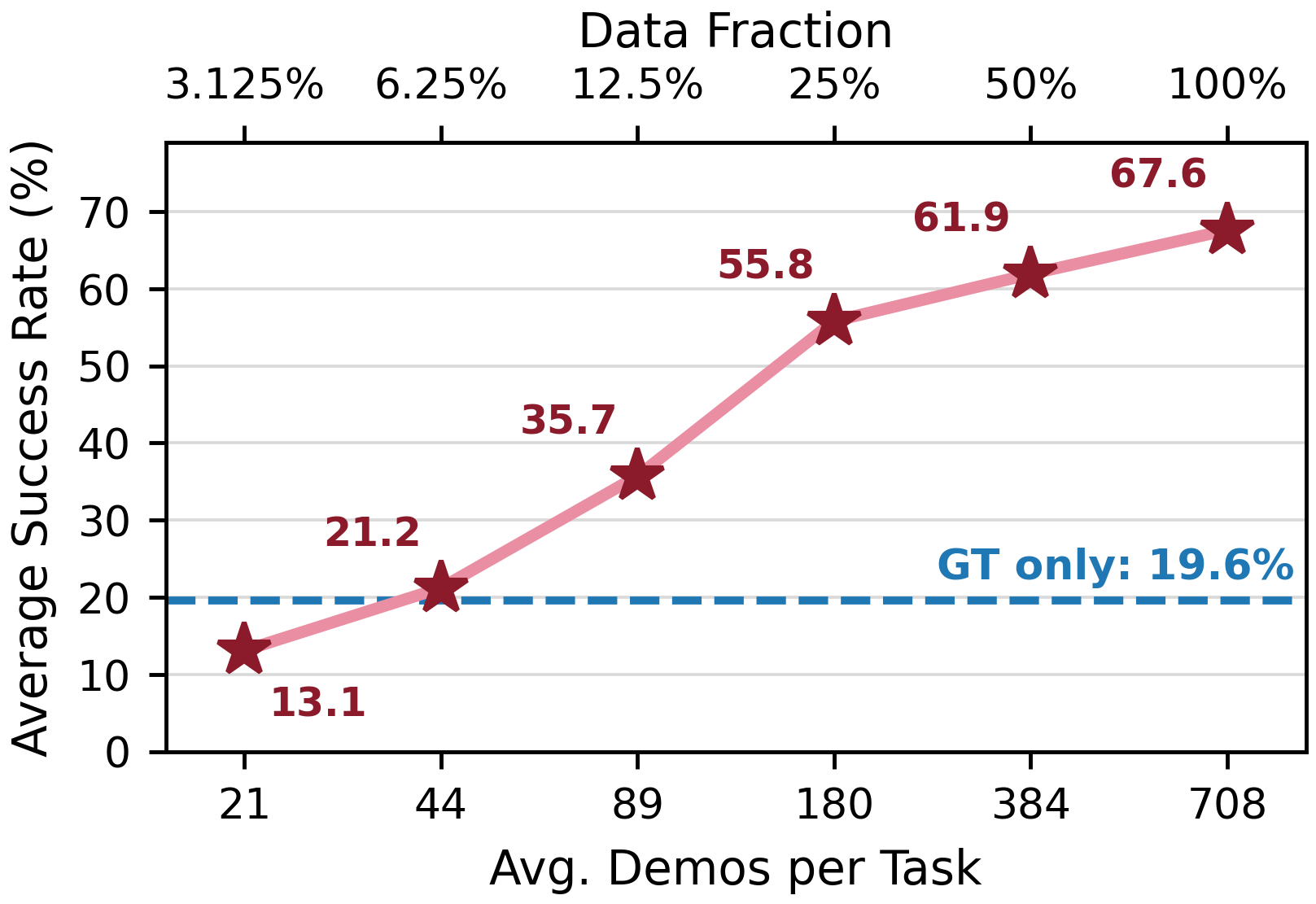}
\caption{OOD success rate on the LIBERO-PRO~\cite{zhou2025liberopro} Object suite improves with data scaling, averaged over the Position, Task, and Environment perturbations.}
\label{fig:libero-pro-scaling}
\end{wrapfigure}

\paragraph{Generated experience improves OOD generalization.}
We first test whether experience discovered by \model{} improves the
generalization of downstream VLAs.
We evaluate on LIBERO-PRO~\cite{zhou2025liberopro}, an augmented version
of LIBERO~\cite{liu2023libero} designed to test generalization under
language, object, position, task, and environment perturbations.

We reconstruct the LIBERO environments in IsaacLab, systematically vary
objects, layouts, visual appearance, and task specifications, and use
\model{} to autonomously generate demonstrations in the resulting
environments.
This produces a dataset approximately \textbf{$10\times$} larger than
the original one.
We then train Pi0.5~\cite{pi05} either on the original LIBERO data alone
or augmented with \model{}-generated trajectories, and evaluate on
$33$ pick-and-place tasks across the four LIBERO-PRO suites.
Details on environment generation, data collection, training, and
evaluation are provided in Appendix~\ref{app:sim_details}.

Table~\ref{tab:libero-pro} shows that adding \model{} experience yields
large gains under distribution shifts requiring new spatial grounding or
behavior.
The strongest improvements occur under \emph{Position} and \emph{Task}
perturbations: for example, on the Object suite, success increases from
$0.0\%$ to $77.0\%$ under Position shifts and from $10.0\%$ to $43.2\%$
under Task shifts.
Performance under \emph{Environment} perturbations also improves across
all four suites.

Following LIBERO-PRO~\cite{zhou2025liberopro}, we treat apparent robustness
to isolated \emph{Language} and \emph{Object} perturbations with caution,
as these dimensions can preserve exploitable correlations from the original
benchmark.
We therefore focus the main table on Position, Task, and Environment
perturbations and report Language/Object results together with all per-task
results in Appendix~\ref{app:sim_res}.
Figure~\ref{fig:libero-pro-scaling} further shows that OOD performance
improves steadily as more \model{}-generated experience is added.

\begin{table}[htbp]
\centering
\small
\setlength{\tabcolsep}{4pt}
\begin{tabular}{l ccc ccc}
\toprule
& \multicolumn{3}{c}{Object} & \multicolumn{3}{c}{Spatial} \\
\cmidrule(lr){2-4}\cmidrule(lr){5-7}
Method & Pos. & Task & Env. & Pos. & Task & Env. \\
\midrule
GT only          & 0.0 & 10.0 & 48.8 & 46.6 & 49.4 & 52.4 \\
\model{} Data\ + GT & \best{77.0}~\up{77.0} & \best{43.2}~\up{33.2} & \best{82.6}~\up{33.8} & \best{71.6}~\up{25.0} & \best{72.4}~\up{23.0} & \best{72.4}~\up{20.0} \\
\midrule
& \multicolumn{3}{c}{Goal} & \multicolumn{3}{c}{10} \\
\cmidrule(lr){2-4}\cmidrule(lr){5-7}
Method & Pos. & Task & Env. & Pos. & Task & Env. \\
\midrule
GT only          & 11.3 & 18.7 & 74.7 & 4.3 & 4.6 & 20.3 \\
\model{} Data\ + GT & \best{60.3}~\up{49.0} & \best{50.7}~\up{32.0} & \best{78.7}~\up{4.0} & \best{43.4}~\up{39.1} & \best{33.4}~\up{28.8} & \best{43.4}~\up{23.1} \\
\bottomrule
\end{tabular}
\caption{Success rate (\%) under the Position, Task, and Environment perturbations of all four LIBERO-PRO~\cite{zhou2025liberopro} suites, with and without our generated data. }
\label{tab:libero-pro}
\end{table}

\begin{table}[htbp]
\centering
\small
\setlength{\tabcolsep}{4pt}
\begin{tabular}{lccccc}
\toprule
Method & L+O & L+P & L+T & L+E & O+P \\
\midrule
GT only          & \best{97.8} & 0.0 & 10.0 & 39.8 & 0.0 \\
\model{} Data + GT & 89.4~\dn{8.4} & \best{78.4}~\up{78.4} & \best{40.8}~\up{30.8} & \best{77.1}~\up{37.3} & \best{82.6}~\up{82.6} \\
\midrule
Method & O+T & O+E & P+T & P+E & T+E \\
\midrule
GT only          & 10.0 & 40.2 & 0.0 & 19.4 & 9.8 \\
\model{} Data + GT & \best{34.0}~\up{24.0} & \best{57.3}~\up{17.1} & \best{73.0}~\up{73.0} & \best{70.6}~\up{51.2} & \best{50.6}~\up{40.8} \\
\bottomrule
\end{tabular}
\caption{Stress test on the Object suite of LIBERO-PRO~\cite{zhou2025liberopro} under \textbf{pairwise} augmentation combinations with or without our data. L = Language, O = Object, P = Position, T = Task, E = Environment.}
\label{tab:libero-pro-pairs}
\end{table}

We further stress-test compositional generalization by applying two
perturbations simultaneously (Table~\ref{tab:libero-pro-pairs}).
Adding \model{} experience improves success across all nine pairwise
combinations containing at least one of \emph{Position}, \emph{Task},
or \emph{Environment}, with gains as large as $82.6$ percentage points.
The only degradation occurs for the Language+Object combination,
consistent with the caveat above regarding these two LIBERO-PRO~\cite{zhou2025liberopro}
perturbation dimensions.

\paragraph{Generated experience enables zero- and few-shot cross-simulator transfer.}
We next ask whether behaviors discovered in IsaacLab provide useful training
experience for deployment in independently constructed simulation benchmarks.

On LIBERO-Object~\cite{liu2023libero}, we train Pi0.5~\cite{pi05}
exclusively on \model{} trajectories generated in IsaacLab.
Without any LIBERO demonstrations, the resulting policy achieves $66.0\%$
success.
Adding only $5$ in-domain demonstrations per task increases success to
$82.2\%$, compared with $95.0\%$ when training on all $50$ ground-truth
demonstrations per task (Fig.~\ref{fig:libero_fewshot}).

We further instantiate \model{} with a WidowX embodiment and generate
experience for four SIMPLER-WidowX~\cite{SIMPLER} tasks in IsaacLab.
Training Pi0.5 solely on this generated experience yields $77.1\%$
average success when deployed zero-shot in SIMPLER
(Fig.~\ref{fig:simpler_zeroshot}).
Together, these results show that \model{}-generated experience transfers
beyond the simulator in which it was collected.

\subsection{What Enables Effective Agentic Exploration?}
\label{sec:ablation_study}

We isolate four components that enable \model{} to efficiently discover
successful behaviors: robust physical execution through learned NIS,
branching exploration through tree search, outcome verification and
reflection, and experience memory.
\ifappendix
Details on task/asset selection and experimental setup are provided in
Appendix~\ref{app:ablation_study}.
\fi

\suppressfloats[t]
\begin{figure}[t]
\centering
\begin{minipage}[t]{0.30\textwidth}
\centering
\includegraphics[height=3.1cm]{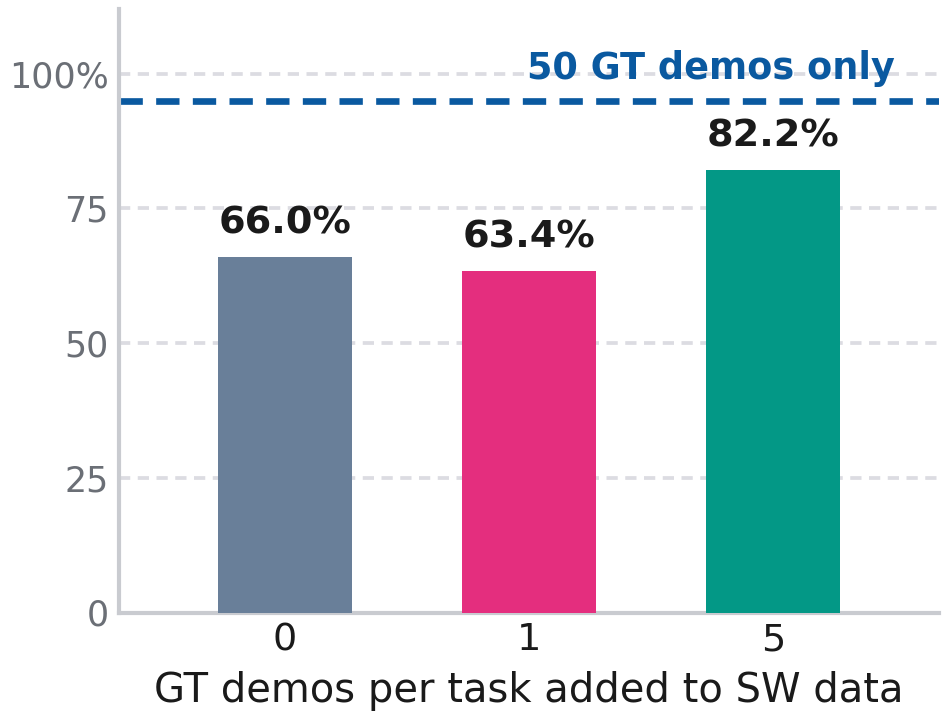}
\captionsetup{font=footnotesize}
\caption{Zero-/few-shot sim-to-sim transfer to LIBERO-Object.}
\label{fig:libero_fewshot}
\end{minipage}\hfill
\begin{minipage}[t]{0.30\textwidth}
\centering
\includegraphics[height=3.1cm]{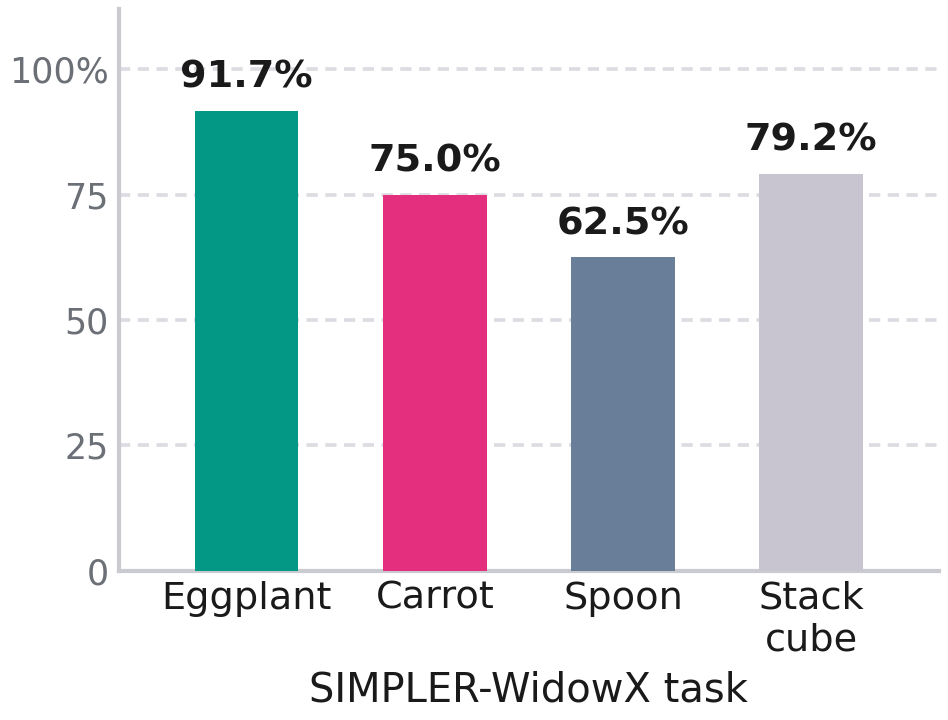}
\captionsetup{font=footnotesize}
\caption{Zero-shot transfer to SIMPLER-WidowX.}
\label{fig:simpler_zeroshot}
\end{minipage}\hfill
\begin{minipage}[t]{0.32\textwidth}
\centering
\includegraphics[height=3.1cm]{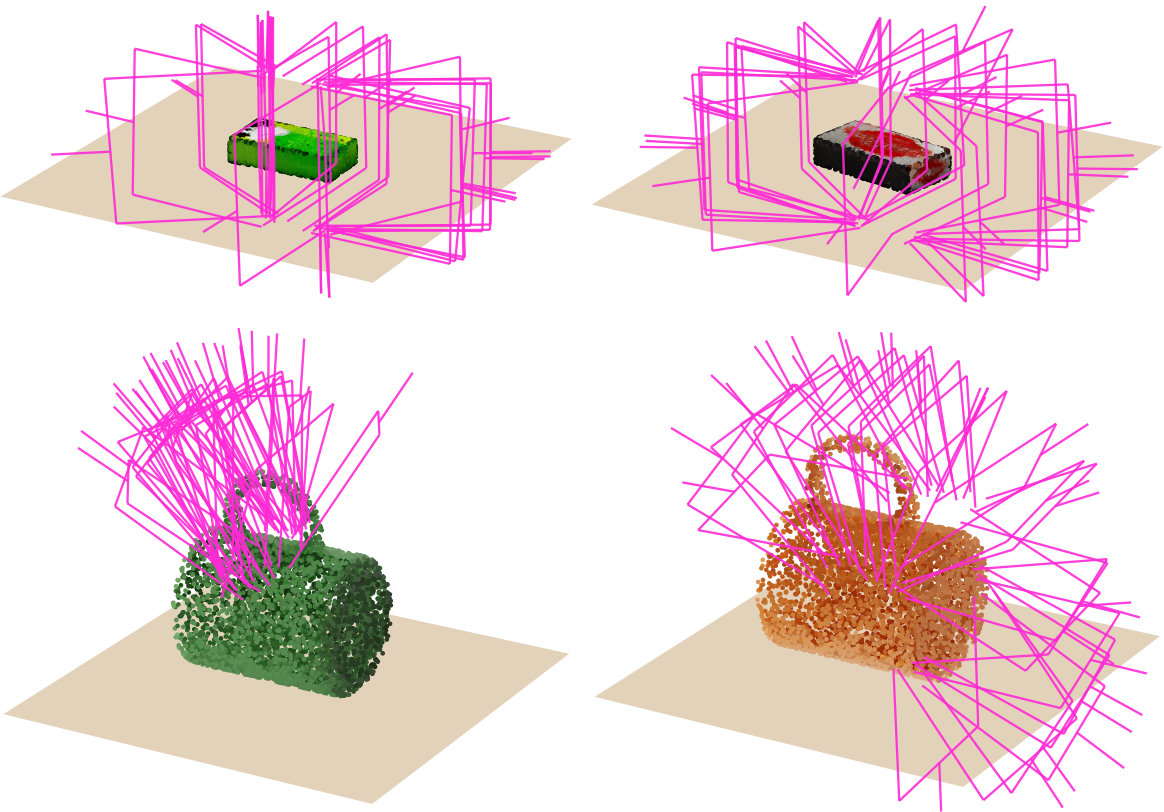}
\captionsetup{font=footnotesize}
\caption{Common failure modes of grasp proposers.}
\label{fig:grasp_proposer_failure}
\end{minipage}
\end{figure}

\paragraph{NIS provide robust contact-rich execution.}
Agentic exploration requires physical skills that execute reliably across
diverse interaction states.
We compare our learned \textsc{Pick} NIS against an engineered pipeline
combining GraspGen~\cite{murali2025graspgen} with a motion
planner~\cite{sundaralingam2023curobo}.
We evaluate on $170$ assets excluded from NIS training and spanning diverse
geometry and pose regimes (see App.~\ref{app:rl_skill_ablation}).

As shown in Table~\ref{tab:execution-search-ablation}(a), the learned NIS
achieves $97.4\%$ overall pick success compared with $65.1\%$ for the
scripted pipeline, with particularly large gains for 
flat objects and concave objects with thin rims.
Figure~\ref{fig:grasp_proposer_failure} illustrates common failure modes
of the engineered pipeline: proposed grasps can be unreachable or
incompatible with the surrounding geometry, for example penetrating the
table with side grasps or targeting structures exceeding the gripper aperture.
Learned closed-loop execution instead adapts throughout the contact-rich
interaction.

\paragraph{Branching exploration improves data-generation efficiency.}
We compare MCTS against linear skill sequencing, corresponding to a
degenerate search tree with breadth one, on five long-horizon tasks under
the same simulation budget (task details in App.~\ref{app:mcts_ablation}).
Both methods use the same agent and NIS library; they differ in whether
alternative interaction branches can be explored while preserving
successful prefixes.

As shown in Table~\ref{tab:execution-search-ablation}(b), branching search
reduces the average number of expansions per successful trajectory from
$6.74$ to $3.60$ and wall-clock time from $891$s to $645$s.
Explicitly maintaining alternative physical strategies therefore enables
more efficient discovery than committing to a single evolving skill sequence.

\begin{table}[htbp]
\centering
\small

\textbf{(a) Pick success rate: NIS vs.\ scripted planning.}
\vspace{2pt}

\setlength{\tabcolsep}{3pt}
\begin{tabular}{lccccc|c}
\toprule
Method & round & flat & elongated upright & concave & concave lying & Overall \\
\midrule
Scripted   & 87.0 & 55.6 & 93.2 & 6.0 & 2.0 & 65.1 \\
NIS (Ours)  & \best{99.0} & \best{94.0} & \best{99.6} & \best{97.0} & \best{98.0} & \best{97.4} \\
\bottomrule
\end{tabular}

\vspace{6pt}
\textbf{(b) Exploration efficiency: MCTS vs.\ linear skill sequencing.}
\vspace{2pt}

\setlength{\tabcolsep}{3pt}
\begin{tabular}{lccccc}
\toprule
Method & Breadth & Max Depth & Simulation Budget & Expansions / Success Traj. & Time / Success Traj. (s) \\
\midrule
Linear & 1 & 20 & 20 & 6.74 & 891 \\
MCTS   & 2 & 4 & 20 & \best{3.60} & \best{645} \\
\bottomrule
\end{tabular}
\caption{
\textbf{Physical execution and exploration ablations.}
(a) Pick-up success rate (\%) across geometry/pose profiles for our learned
\textsc{Pick} NIS versus an engineered grasp-proposal and motion-planning
pipeline.
(b) Search efficiency of branching MCTS versus linear skill sequencing on
five long-horizon tasks under the same simulation budget.
}
\label{tab:execution-search-ablation}
\end{table}

\paragraph{Hierarchical verification guides exploration and filters data.}
We ablate two distinct roles of the verifier in \model{}: providing intermediate
feedback that guides exploration and determining whether discovered
trajectories should enter the training dataset.

For search guidance, verifier feedback is particularly valuable under tight
exploration budgets.
At $K\leq4$, verifier-guided search succeeds on $58.3\%$ of trials,
compared with $8.3\%$ without intermediate verifier rewards and $6.7\%$
with standard rollout-based MCTS.
The advantage persists as the search budget increases, while verifier-guided
search also substantially reduces wall-clock time per successful trajectory
(Table~\ref{tab:verifier-ablation}(a)).

For dataset filtering, we compare verifier decisions against handcrafted
task-specific success conditions.
An image-only verifier reaches $78.8\%$ overall accuracy, whereas exposing
privileged simulator state raises accuracy to $99.5\%$
(Table~\ref{tab:verifier-ablation}(b)).
This highlights an important advantage of generating experience through
agentic interaction in simulation: privileged physical state can provide
reliable feedback during exploration and data generation, while the
downstream policy is trained for deployment from sensory observations.

Appendix~\ref{app:verifier_ablation} provides the exact configurations,
evaluation tasks, and handcrafted success criteria.

\begin{table}[htbp]
\centering
\small

\textbf{(a) Search guidance with intermediate rewards.}

\setlength{\tabcolsep}{4pt}
\begin{tabular}{l ccc ccc ccc}
\toprule
 & \multicolumn{3}{c}{w/ Verifier, w/o Rollout}
 & \multicolumn{3}{c}{w/o Verifier, w/o Rollout}
 & \multicolumn{3}{c}{w/o Verifier, w/ Rollout} \\
\cmidrule(lr){2-4} \cmidrule(lr){5-7} \cmidrule(lr){8-10}
 & $K{\le}4$ & $K{\le}5$ & $K{\le}6$
 & $K{\le}4$ & $K{\le}5$ & $K{\le}6$
 & $K{\le}4$ & $K{\le}5$ & $K{\le}6$ \\
\midrule
Avg. Success Rate (\%)
 & \best{58.3} & \best{76.7} & \best{95.0}
 & 8.3 & 31.7 & 50.0
 & 6.7 & 21.7 & 43.3 \\
\midrule
Time / Success Traj (s)
 & \multicolumn{3}{c}{\best{454}}
 & \multicolumn{3}{c}{943}
 & \multicolumn{3}{c}{2872} \\
\bottomrule
\end{tabular}

\vspace{6pt}
\textbf{(b) Final success detection accuracy.}
\vspace{2pt}

\setlength{\tabcolsep}{6pt}
\begin{tabular}{lccccc}
\toprule
Setting & pick up & place & orientation aware & straighten & Overall \\
\midrule
Image-only          & 85.7  & 77.0  & 76.0 & 76.5 & 78.8 \\
+ Privileged States & \best{100.0} & \best{100.0} & \best{99.0} & \best{99.0} & \best{99.5} \\
\bottomrule
\end{tabular}

\caption{
Top: verifier-guided MCTS uses verifier scores as intermediate rewards without rollouts; 
no-verifier MCTS performs uniform selection on frontier nodes; 
rollout-based search performs standard MCTS rollouts.
Bottom: verifier accuracy compared to handcrafted ground-truth labels across task types with and without privileged simulator states.
}
\label{tab:verifier-ablation}
\end{table}

\paragraph{Experience memory transfers discoveries across searches.}
Finally, we test whether strategies discovered during previous searches
can improve future exploration.
We evaluate four challenging tasks involving orientation adjustment,
spatial reasoning, or implicit prerequisite discovery.
As shown in Table~\ref{tab:memory-ablation}, experience memory increases
average search success from $37.5\%$ to $85.0\%$.

The effect is particularly striking when successful behavior requires an
implicit prerequisite.
For Toothbrush $\rightarrow$ Lying Cup, success increases from $0\%$
without memory to $80\%$ with memory: retrieved experience informs the
agent that the lying cup should first be placed upright before attempting
insertion.
Thus, \model{} can reuse abstractions learned from previous interaction
rather than rediscovering successful strategies independently in every
search.
\begin{table}[htbp]
\centering
\small
\setlength{\tabcolsep}{4pt}
\begin{tabular}{lccccc}
\toprule
Method
& Pudding $\rightarrow$ Side
& Bottle Upright 
& Toothbrush $\rightarrow$ Cup 
& Toothbrush $\rightarrow$ Lying Cup
& Avg. \\
\midrule
w/o Memory & 20.0 & 50.0 & 80.0 & 0.0  & 37.5 \\
w/ Memory  & \best{100.0} & \best{60.0} & \best{100.0} & \best{80.0} & \best{85.0} \\
\bottomrule
\end{tabular}
\caption{
MCTS success rate (\%) with and without memory on challenging tasks involving orientation adjustment, spatial reasoning,
or implicit prerequisite discovery.
}
\label{tab:memory-ablation}
\end{table}

\subsection{Transfer to Real-World Manipulation}
\label{sec:realworld}

Finally, we ask whether experience autonomously generated by \model{} in
simulation can improve real-world robot policies \textit{\textbf{without any real-world
teleoperation.}}

Our training pipeline has two stages.
During \emph{large-scale simulation training} (LST), \model{} explores
diverse in-the-wild scenes and generates broad manipulation experience.
During \emph{post-training} (PT), we reconstruct the target real-world
environment in simulation and use \model{} to generate additional
task-specific trajectories in the reconstructed scene.
Both stages therefore use simulation-generated demonstrations; no
real-world demonstrations are used for training.
We deploy the resulting Pi0.5 policies zero-shot on a real Franka Panda.
Details are provided in Appendix~\ref{app:real_res}.

We evaluate object, color, and spatial grounding in pick-and-place tasks
under the original real-world setup and two distribution shifts: swapped
object positions and a changed tablecloth.
Table~\ref{tab:realworld} compares policies trained using only
target-environment post-training data (PT) against policies additionally
trained on large-scale \model{} experience (LST+PT).

\newcommand{\srci}[3]{\hphantom{\textbf{00}}\llap{#1} {\scriptsize\mdseries\color{gray}\rlap{[#2,#3]}\hphantom{[00.0,00.0]}}}

\begin{table*}[htbp]
\centering
\setlength{\tabcolsep}{2pt}
\renewcommand{\arraystretch}{1.15}
\footnotesize

\resizebox{0.98\textwidth}{!}{%
\begin{tabular}{l l cc cc cc}
\toprule
& & \multicolumn{2}{c}{Original}
& \multicolumn{2}{c}{Swap Objects}
& \multicolumn{2}{c}{Change Background} \\
\cmidrule(lr){3-4}
\cmidrule(lr){5-6}
\cmidrule(lr){7-8}

& Instruction
& PT & LST+PT
& PT & LST+PT
& PT & LST+PT \\
\midrule

Object Recognition
& pick up the apple
& \srci{40}{26.4}{54.8}
& \best{\srci{82}{68.6}{91.4}}
& \srci{0}{0.0}{7.1}
& \best{\srci{86}{73.3}{94.2}}
& \srci{32}{19.5}{46.7}
& \best{\srci{50}{35.5}{64.5}} \\

& pick up the banana
& \best{\srci{82}{68.6}{91.4}}
& \srci{66}{51.2}{78.8}
& \srci{10}{3.3}{21.8}
& \best{\srci{80}{66.3}{90.0}}
& \best{\srci{78}{64.0}{88.5}}
& \srci{44}{30.0}{58.7} \\

& place the apple in the bowl
& \srci{4}{0.5}{13.7}
& \best{\srci{94}{83.5}{98.7}}
& \srci{0}{0.0}{7.1}
& \best{\srci{60}{45.2}{73.6}}
& \srci{0}{0.0}{7.1}
& \best{\srci{78}{64.0}{88.5}} \\

& place the lemon in the bowl
& \srci{0}{0.0}{7.1}
& \best{\srci{50}{35.5}{64.5}}
& \srci{0}{0.0}{7.1}
& \best{\srci{48}{33.7}{62.6}}
& \srci{0}{0.0}{7.1}
& \best{\srci{76}{61.8}{86.9}} \\

\cmidrule(lr){2-8}

Spatial Recognition
& pick up the fruit on the left
& \srci{36}{22.9}{50.8}
& \best{\srci{80}{66.3}{90.0}}
& \srci{0}{0.0}{7.1}
& \best{\srci{72}{57.5}{83.8}}
& \srci{16}{7.2}{29.1}
& \best{\srci{44}{30.0}{58.7}} \\

& pick up the fruit on the right
& \best{\srci{72}{57.5}{83.8}}
& \srci{58}{43.2}{71.8}
& \srci{0}{0.0}{7.1}
& \best{\srci{58}{43.2}{71.8}}
& \srci{18}{8.6}{31.4}
& \best{\srci{20}{10.0}{33.7}} \\

& place the fruit at the front in the bowl
& \srci{0}{0.0}{7.1}
& \best{\srci{26}{14.6}{40.3}}
& \srci{0}{0.0}{7.1}
& \best{\srci{6}{1.3}{16.5}}
& \srci{0}{0.0}{7.1}
& \best{\srci{26}{14.6}{40.3}} \\

& place the fruit at the back in the bowl
& \srci{0}{0.0}{7.1}
& \best{\srci{18}{8.6}{31.4}}
& \srci{0}{0.0}{7.1}
& \best{\srci{30}{17.9}{44.6}}
& \srci{0}{0.0}{7.1}
& \best{\srci{30}{17.9}{44.6}} \\

\cmidrule(lr){2-8}

Color Recognition
& pick up the red fruit
& \srci{50}{35.5}{64.5}
& \best{\srci{90}{78.2}{96.7}}
& \srci{0}{0.0}{7.1}
& \best{\srci{82}{68.6}{91.4}}
& \srci{34}{21.2}{48.8}
& \best{\srci{58}{43.2}{71.8}} \\

& pick up the yellow fruit
& \best{\srci{84}{70.9}{92.8}}
& \srci{52}{37.4}{66.3}
& \srci{6}{1.3}{16.5}
& \best{\srci{78}{64.0}{88.5}}
& \srci{16}{7.2}{29.1}
& \best{\srci{18}{8.6}{31.4}} \\

& place the red fruit in the bowl
& \srci{0}{0.0}{7.1}
& \best{\srci{76}{61.8}{86.9}}
& \srci{0}{0.0}{7.1}
& \best{\srci{50}{35.5}{64.5}}
& \srci{0}{0.0}{7.1}
& \best{\srci{50}{35.5}{64.5}} \\

& place the yellow fruit in the bowl
& \srci{2}{0.1}{10.6}
& \best{\srci{58}{43.2}{71.8}}
& \srci{0}{0.0}{7.1}
& \best{\srci{36}{22.9}{50.8}}
& \srci{0}{0.0}{7.1}
& \best{\srci{6}{1.3}{16.5}} \\

\midrule

\multicolumn{2}{l}{Average}
& 30.8
& \best{62.5}\,\up{31.7}
& 1.3
& \best{57.2}\,\up{55.8}
& 16.2
& \best{41.7}\,\up{25.5} \\

\bottomrule
\end{tabular}%
}

\caption{
Per-task real-world success rates (\%)
where entries report the pooled success rate across $50$ trials with the 95\% Clopper--Pearson exact confidence interval.
}
\label{tab:realworld}
\end{table*}

Large-scale \model{} experience substantially improves average real-world
success in all three settings: from $30.8\%$ to $62.5\%$ in the original
environment, from $1.3\%$ to $57.2\%$ after swapping object positions,
and from $16.2\%$ to $41.7\%$ after changing the background.
The particularly large gain under object swaps suggests that broad
simulation experience reduces reliance on the specific object
configurations encountered during target-environment post-training.
While individual tasks remain challenging---particularly spatially
specified placement---the overall results show that agent-generated
simulation experience can substantially improve real-world robustness
without additional real-world teleoperation.
Additional qualitative results are provided in
Appendix~\ref{app:qualitative_real}.

\FloatBarrier

\section{Limitations and Future Work}

\model{} currently explores over a small library of Neural Interaction Skills (NIS),
instantiated as learned closed-loop policies for \textsc{Pick}, \textsc{Place},
\textsc{Open}, and \textsc{Close}.
While these tools support a broad range of manipulation tasks, the space of behaviors
that the agent can discover is ultimately bounded by the physical capabilities exposed
through its tool library.
Extending \model{} to non-prehensile manipulation, tool use, bimanual coordination,
deformable objects, and richer embodiments such as dexterous hands will require learning
a broader and more compositional repertoire of interaction tools.
An exciting direction is to move beyond a fixed tool library and allow the agent to
continually acquire, refine, or compose new Neural Interaction Skills from experience.

The quality of autonomously generated experience also depends on both the fidelity of
the simulated environment and the reasoning capabilities of the agent.
Improving simulation fidelity, reasoning and verification, and extending agentic
exploration to incorporate real-world interaction are therefore important directions
for future work.

It is also worth investigating more expressive forms of distillation, such as
hierarchical policies that preserve the tool-use structure of \model{},
as well as continual data-generation loops in which downstream policy failures trigger targeted
agentic exploration and additional data collection.

\section{Conclusion}

We presented \model{}, a framework for scalable robot data generation through
agentic exploration over Neural Interaction Skills (NIS).
Rather than relying on teleoperation, scripted execution pipelines, or task-specific
program generation, \model{} equips a reasoning agent with learned closed-loop
physical capabilities and allows it to explore their composition and parameterization
through interaction with simulation.
Hierarchical verification, reflection, and semantic memory guide this exploration,
enabling the autonomous discovery of successful long-horizon manipulation behaviors
that are retained as robot training data.

Across LIBERO~\cite{liu2023libero}, LIBERO-PRO~\cite{zhou2025liberopro}, SIMPLER-WidowX~\cite{SIMPLER}, 
and real-world Franka manipulation,
the resulting data substantially improves downstream visuomotor policy generalization
and enables robust sim-to-sim and sim-to-real transfer.
More broadly, our results suggest a path toward robot learning systems in which
agents not only learn from demonstrations provided to them, but also actively
\emph{generate their own training experience} by reasoning, acting, verifying,
and exploring in interactive worlds.

\appendix

\makeatletter
\let\app@oldsection\section
\renewcommand{\section}{\FloatBarrier\app@oldsection}
\let\app@oldsubsection\subsection
\renewcommand{\subsection}{\FloatBarrier\app@oldsubsection}
\makeatother

\setcounter{table}{0}\renewcommand{\thetable}{A\arabic{table}}
\setcounter{figure}{0}\renewcommand{\thefigure}{A\arabic{figure}}
\setcounter{equation}{0}\renewcommand{\theequation}{A\arabic{equation}}
\clearpage
\addtocontents{toc}{\protect\setcounter{tocdepth}{2}}
\makeatletter\newcommand\appendixtoclist{\@starttoc{toc}}\makeatother
{\hypersetup{linkcolor=inkdark}%
 \titlecontents{section}[1.9em]{\addvspace{0.6em}\sffamily\bfseries}%
   {\contentslabel[{\color{tgradB}\thecontentslabel}]{1.9em}}{}{\hfill\contentspage}%
 \titlecontents{subsection}[4.4em]{\addvspace{0.25em}}%
   {\contentslabel{2.5em}}{}%
   {\,{\color{affilgray}\titlerule*[0.7pc]{.}}\contentspage}%
 \noindent\hspace*{\dimexpr(\textwidth-\cardwidth)/2\relax}\begin{tikzpicture}
 \node[fill=cardbg,rounded corners=9mm,inner sep=\cardinner,text width=\cardtext,align=left]{%
   \setlength{\parindent}{0pt}\setlength{\parskip}{0pt}%
   {\sffamily\bfseries\fontsize{15}{18}\selectfont\color{tgradB}Appendix Contents\par}%
   \vspace{5pt}\gradrule{\linewidth}\par\vspace{6pt}%
   \appendixtoclist\par};
 \end{tikzpicture}\par}
\clearpage

\section{Additional Tree Search Statistics}
\label{app:tree_search_stats}

\subsection{Per-skill Success Rates during Tree Search}

\begin{table}[htbp]
\centering
\small
\begin{tabular}{lcccccc}
\toprule
 & Pick & Place & Open prismatic & Open revolute & Close prismatic & Close revolute \\
\midrule
Success rate (\%) & 89.9 & 79.6 & 53.7 & 78.3 & 69.5 & 72.9 \\
\bottomrule
\end{tabular}
\caption{Success rates (\%) for each of the learned NIS computed over all tree search nodes that executed the skill.}
\label{tab:per_skill_success}
\end{table}

\subsection{Per-task Success Rates for LIBERO-PRO tasks}

We report the tree search success rate of every LIBERO-PRO~\cite{zhou2025liberopro} task for which we collected data, i.e., the fraction of search runs that yield at least one successful trajectory, pooled over all runs of the task. Task ID $i$ denotes the $i$-th original LIBERO~\cite{liu2023libero} task of each suite, and $i$-T$k$ its $k$-th LIBERO-PRO task augmentation. Tasks involving pushing or turning, and augmented variants without collected data, are omitted.

{\footnotesize
\captionsetup{font=normalsize,position=below}
\setlength{\LTcapwidth}{\textwidth}
\begin{longtable}{p{0.60\textwidth} c c c}
\toprule
Instruction & Task ID & Type & Succ. (\%) \\
\midrule
\endfirsthead
\toprule
Instruction & Task ID & Type & Succ. (\%) \\
\midrule
\endhead
\bottomrule
\endfoot
\bottomrule
\caption{Per-task tree search success rates (\%) on the \textbf{LIBERO-PRO-Object} tasks, covering original tasks and LIBERO-PRO task augmentations.}
\label{tab:libero_pro_tree_object} \\
\endlastfoot
pick up the alphabet soup and place it in the basket & 0 & Original & 77.0 \\
pick the cream cheese and place it in the basket & 0-T1 & Augmented & 74.6 \\
pick up the cream cheese and place it in the basket & 1 & Original & 63.2 \\
pick the alphabet soup and place it in the basket & 1-T1 & Augmented & 69.6 \\
pick up the salad dressing and place it in the basket & 2 & Original & 82.3 \\
pick the tomato sauce and place it in the basket & 2-T1 & Augmented & 84.6 \\
pick up the bbq sauce and place it in the basket & 3 & Original & 81.2 \\
pick the ketchup and place it in the basket & 3-T1 & Augmented & 82.9 \\
pick up the ketchup and place it in the basket & 4 & Original & 72.5 \\
pick the milk and place it in the basket & 4-T1 & Augmented & 81.5 \\
pick up the tomato sauce and place it in the basket & 5 & Original & 75.1 \\
pick the bbq sauce and place it in the basket & 5-T1 & Augmented & 82.9 \\
pick up the butter and place it in the basket & 6 & Original & 69.3 \\
pick the orange juice and place it in the basket & 6-T1 & Augmented & 85.1 \\
pick up the milk and place it in the basket & 7 & Original & 64.0 \\
pick the butter and place it in the basket & 7-T1 & Augmented & 82.9 \\
pick up the chocolate pudding and place it in the basket & 8 & Original & 72.2 \\
pick the salad dressing and place it in the basket & 8-T1 & Augmented & 69.1 \\
pick up the orange juice and place it in the basket & 9 & Original & 72.5 \\
pick the chocolate pudding and place it in the basket & 9-T1 & Augmented & 72.4 \\
\end{longtable}}

{\footnotesize
\captionsetup{font=normalsize,position=below}
\setlength{\LTcapwidth}{\textwidth}
\begin{longtable}{p{0.60\textwidth} c c c}
\toprule
Instruction & Task ID & Type & Succ. (\%) \\
\midrule
\endfirsthead
\toprule
Instruction & Task ID & Type & Succ. (\%) \\
\midrule
\endhead
\bottomrule
\endfoot
\bottomrule
\caption{Per-task tree search success rates (\%) on the \textbf{LIBERO-PRO-Spatial} tasks, covering original tasks and LIBERO-PRO task augmentations.}
\label{tab:libero_pro_tree_spatial} \\
\endlastfoot
pick up the black bowl between the plate and the ramekin and place it on the plate & 0 & Original & 94.3 \\
pick the akita black bowl not between the plate and the ramekin and place it on the plate & 0-T1 & Augmented & 96.2 \\
pick up the black bowl next to the ramekin and place it on the plate & 1 & Original & 97.1 \\
pick the akita black bowl next to the cookie box and place it on the plate & 1-T1 & Augmented & 72.4 \\
pick up the black bowl from table center and place it on the plate & 2 & Original & 90.9 \\
pick the akita black bowl next to the plate and place it on the plate & 2-T1 & Augmented & 96.2 \\
pick up the black bowl on the cookie box and place it on the plate & 3 & Original & 81.2 \\
pick the akita black bowl on the top of the cabinet and place it on the plate & 3-T1 & Augmented & 91.4 \\
pick up the black bowl in the top drawer of the wooden cabinet and place it on the plate & 4 & Original & 0.0 \\
pick the akita black bowl on the top of the wooden cabinet and place it on the plate & 4-T1 & Augmented & 89.2 \\
pick up the black bowl on the ramekin and place it on the plate & 5 & Original & 93.7 \\
pick the akita black bowl on the cookie box and place it on the plate & 5-T1 & Augmented & 99.3 \\
pick up the black bowl next to the cookie box and place it on the plate & 6 & Original & 68.9 \\
pick the akita black bowl on the stove and place it on the plate & 6-T1 & Augmented & 95.5 \\
pick up the black bowl on the stove and place it on the plate & 7 & Original & 89.4 \\
pick the akita black bowl on the top of the cabinet and place it on the plate & 7-T1 & Augmented & 97.4 \\
pick up the black bowl next to the plate and place it on the plate & 8 & Original & 93.0 \\
pick the akita black bowl next to the ramekin and place it on the plate & 8-T1 & Augmented & 98.0 \\
pick up the black bowl on the wooden cabinet and place it on the plate & 9 & Original & 87.2 \\
pick the akita black bowl on the stove and place it on the plate & 9-T1 & Augmented & 95.5 \\
\end{longtable}}

{\footnotesize
\captionsetup{font=normalsize,position=below}
\setlength{\LTcapwidth}{\textwidth}
\begin{longtable}{p{0.60\textwidth} c c c}
\toprule
Instruction & Task ID & Type & Succ. (\%) \\
\midrule
\endfirsthead
\toprule
Instruction & Task ID & Type & Succ. (\%) \\
\midrule
\endhead
\bottomrule
\endfoot
\bottomrule
\caption{Per-task tree search success rates (\%) on the \textbf{LIBERO-PRO-Goal} tasks, covering original tasks and LIBERO-PRO task augmentations.}
\label{tab:libero_pro_tree_goal} \\
\endlastfoot
open the middle drawer of the cabinet & 0 & Original & 28.9 \\
open the top drawer of the cabinet & 0-T2 & Augmented & 79.1 \\
put the bowl on the stove & 1 & Original & 73.0 \\
put the plate on the stove & 1-T1 & Augmented & 29.9 \\
put the wine bottle on top of the cabinet & 2 & Original & 62.0 \\
put the wine bottle in the bowl & 2-T1 & Augmented & 63.2 \\
open the top drawer and put the bowl inside & 3 & Original & 35.4 \\
open the top layer of the drawer and put the cream cheese inside & 3-T1 & Augmented & 54.1 \\
open the middle layer of the drawer and put the bowl inside & 3-T2 & Augmented & 14.3 \\
put the bowl on top of the cabinet & 4 & Original & 76.7 \\
put the plate on the top of the drawer & 4-T1 & Augmented & 26.7 \\
put the cream cheese in the bowl & 6 & Original & 72.3 \\
put the wine bottle in the bowl & 6-T1 & Augmented & 67.2 \\
put the bowl on the plate & 8 & Original & 92.6 \\
put the wine bottle on the plate & 8-T1 & Augmented & 66.2 \\
put the wine bottle on the rack & 9 & Original & 56.9 \\
put the wine bottle on the stove & 9-T2 & Augmented & 71.4 \\
\end{longtable}}

{\footnotesize
\captionsetup{font=normalsize,position=below}
\setlength{\LTcapwidth}{\textwidth}
\begin{longtable}{p{0.60\textwidth} c c c}
\toprule
Instruction & Task ID & Type & Succ. (\%) \\
\midrule
\endfirsthead
\toprule
Instruction & Task ID & Type & Succ. (\%) \\
\midrule
\endhead
\bottomrule
\endfoot
\bottomrule
\caption{Per-task tree search success rates (\%) on the \textbf{LIBERO-PRO-10} tasks, covering original tasks and LIBERO-PRO task augmentations. For task 4, LIBERO-PRO's T1 and T2 augmentations share the same goal and are listed once.}
\label{tab:libero_pro_tree_long} \\
\endlastfoot
put both the alphabet soup and the tomato sauce in the basket & 0 & Original & 74.7 \\
put both the cream cheese and the tomato sauce in the basket & 0-T1 & Augmented & 48.3 \\
put both the alphabet soup and the cream cheese in the basket & 0-T2 & Augmented & 52.4 \\
put both the cream cheese box and the butter in the basket & 1 & Original & 54.2 \\
put both the alphabet soup and the butter in the basket & 1-T1 & Augmented & 69.5 \\
put both the cream cheese box and the alphabet soup in the basket & 1-T2 & Augmented & 62.8 \\
put the black bowl in the bottom drawer of the cabinet and close it & 3 & Original & 30.6 \\
put the bottle in the bottom drawer of the cabinet and close it & 3-T1 & Augmented & 19.5 \\
put the white mug on the left plate and put the yellow and white mug on the right plate & 4 & Original & 87.8 \\
put the yellow and white mug on the left plate and put the white mug on the right plate & 4-T1 & Augmented & 80.6 \\
pick up the book and place it in the back compartment of the caddy & 5 & Original & 75.2 \\
pick up the book and place it in the front compartment of the caddy & 5-T2 & Augmented & 95.9 \\
put the white mug on the plate and put the chocolate pudding to the right of the plate & 6 & Original & 78.1 \\
put the red mug on the plate and put the chocolate pudding to the right of the plate & 6-T1 & Augmented & 78.3 \\
put the white mug on the plate and put the chocolate pudding to the left of the plate & 6-T2 & Augmented & 66.7 \\
put both the alphabet soup and the cream cheese box in the basket & 7 & Original & 85.0 \\
put both the ketchup and the cream cheese box in the basket & 7-T1 & Augmented & 89.8 \\
put both the alphabet soup and the ketchup in the basket & 7-T2 & Augmented & 93.2 \\
put both moka pots on the stove & 8 & Original & 48.8 \\
put the left moka pot on the stove & 8-T1 & Augmented & 73.4 \\
put the right moka pot on the stove & 8-T2 & Augmented & 73.2 \\
put the yellow and white mug in the microwave and close it & 9 & Original & 25.7 \\
put the white mug in the microwave and close it & 9-T1 & Augmented & 25.5 \\
\end{longtable}}

\subsection{Per-task Success Rates for SIMPLER-WidowX tasks}

\begin{table}[htbp]
\centering
\small
\begin{tabular}{lc}
\toprule
Instruction & Success rate (\%) \\
\midrule
put carrot on plate                       & 92.3 \\
put eggplant into yellow basket           & 93.1 \\
put the spoon on the towel                & 42.3 \\
stack the green block on the yellow block & 78.7 \\
\bottomrule
\end{tabular}
\caption{Per-task tree search success rates on the SIMPLER-WidowX~\cite{SIMPLER} tasks, measured as the fraction of search runs that yield at least one successful trajectory.}
\label{tab:simpler_tree_search_success}
\end{table}

\section{Implementation Details on Simulation Experiments}
\label{app:sim_details}

\subsection{Simulation Environment Generation for Benchmark-derived Tasks}

\paragraph{Reference scene reconstruction.}\mbox{}

For each benchmark task, we sample an initial state from the benchmark and render an image of it.
We then reconstruct the 3D geometry and layout of the objects from this image with the targeted exploration mode (App.~\ref{app:scenegen}), which serves as the reference scene for the task.
The background is reconstructed differently for the two benchmark families.
LIBERO~\cite{liu2023libero} and LIBERO-PRO~\cite{zhou2025liberopro} scenes consist of textured walls, tables, and floors, so we prompt GPT to generate texture maps for these surfaces from the rendered image.
For SIMPLER~\cite{SIMPLER} scenes, we extract the foreground from our rendering with the per-object segmentation masks provided by the simulator,
and overlay it onto a background image of the benchmark scene, which we obtain by prompting GPT to erase the foreground objects from the rendered benchmark image.

\paragraph{Scene augmentation.}\mbox{}

Starting from the reference scenes, we augment LIBERO~\cite{liu2023libero} and LIBERO-PRO~\cite{zhou2025liberopro} tasks along the following axes.
\emph{Object size}: each reconstructed benchmark object is rescaled anisotropically, with independent scale factors along its three axes.
\emph{Object position}: we apply local jitter around the reference layout, resample object poses globally over the table, and swap the positions of objects.
\emph{Object replacement}: for LIBERO-PRO~\cite{zhou2025liberopro}, we also render the scenes of its object augmentations and reconstruct the replaced objects with the targeted exploration mode.
\emph{Environment}: we render all LIBERO~\cite{liu2023libero} scenes, prompt GPT to erase their foregrounds, and generate textured walls, tables, and floors from the resulting images;
this LIBERO-derived background pool is shared across all tasks and further mixed with a large in-the-wild background pool,
in which the walls are images of living rooms, kitchens, and offices generated by GPT, the floors are textures generated by GPT, and the tables use textures downloaded from the Internet.
\emph{Task}: we broaden LIBERO~\cite{liu2023libero} tasks by replacing the source and target objects, or articulation parts, in their instructions.
Finally, we randomize lighting, camera poses, and robot initialization as described in App.~\ref{app:env_rand}.
For SIMPLER~\cite{SIMPLER} tasks, we only augment object sizes and positions.

\subsection{Data Collection for Benchmark-derived Tasks}

For benchmark-derived tasks, all augmentations described above are applied jointly when generating each scene,
rather than generating a separate set of scenes for each type of perturbation as in LIBERO-PRO~\cite{zhou2025liberopro}.
As a result, our training distribution covers a much wider range of object sizes, layouts, appearances, backgrounds, and task specifications,
while the individually perturbed LIBERO-PRO evaluation scenes remain out-of-distribution for our dataset.
The same data is also used in the sim-to-sim transfer experiments on LIBERO~\cite{liu2023libero}:
although it is not constructed to match the LIBERO evaluation distribution, the policy trained on it still achieves $66.0\%$ zero-shot success on LIBERO-Object.
In total, we collect about $10\times$ as many LIBERO-style trajectories as the original benchmark provides,
and about $475$ trajectories per task for the four SIMPLER-WidowX~\cite{SIMPLER} tasks.

\subsection{Policy Training}
\label{app:sim_training}

All \model{} training trajectories are collected in IsaacLab~\cite{mittal2025isaac} with a Franka Panda robot for LIBERO~\cite{liu2023libero} and LIBERO-PRO~\cite{zhou2025liberopro},
and a WidowX robot for the SIMPLER~\cite{SIMPLER} tasks.
The LIBERO ground-truth demonstrations are recorded as end-effector delta commands, so we first replay them in simulation to obtain joint-space actions.
Starting from each demonstration's initial state, we drive the arm with the same absolute joint position controller used at evaluation (App.~\ref{app:sim_deploy}), commanding the recorded joint positions of the next frame together with the original gripper command, and record the reached joint angles, the target finger positions, and re-rendered camera images.
We keep only the replays that still complete the task.

For policy distillation, robot arm actions are represented as $k$-dimensional joint positions \emph{relative} to the start of each action chunk, where $k$ is the DoF of the arm: 
for a chunk beginning at frame $t_0$, the action at frame $t$ is
\begin{equation}
    a_t = q_t - q_{t_0},
\end{equation}
where $q_t$ denotes the robot joint position at frame $t$. 
Gripper open/close commands are represented as continuous signals indicating the target finger positions, appended to the end of the action vector.

For each comparison reported, all policies use this same action representation and are trained with the same number of steps and batch size, specifically $256$, on $8$ NVIDIA H100 GPUs.

\subsection{Policy Deployment and Evaluation}
\label{app:sim_deploy}

\paragraph{Robot controllers.}\mbox{}

In LIBERO~\cite{liu2023libero} and LIBERO-PRO~\cite{zhou2025liberopro}, the default robosuite controller is \texttt{OSC\_POSE}, which takes end-effector delta poses and a binary gripper command at 20\,Hz.
We replace it with robosuite's \texttt{JOINT\_POSITION} controller to take in the absolute 7-DoF joint targets predicted by the policy directly.
For finger control, we convert the policy's continuous finger position outputs to LIBERO's binary gripper command 
by its sign relative to the current finger position.
We keep the benchmark's 20\,Hz control rate and replan every 5 steps.

In SIMPLER-WidowX~\cite{SIMPLER}, we similarly write the predicted 6-DoF joint targets and a continuous gripper target directly to the simulator's PD joint drives, 
keeping the robot's default stiffness and damping.
We apply a 20\,Hz control rate and an action chunk length of 10.

\paragraph{Evaluation Setups}\mbox{}

\emph{LIBERO-PRO.}

We evaluate on the $33$ pick-and-place tasks of the four LIBERO-PRO~\cite{zhou2025liberopro} suites: $10$ in Object, $10$ in Spatial, $6$ in Goal, and $7$ in LIBERO-10.
For each perturbation type, we generate the perturbed task definitions with the generation script released by LIBERO-PRO.
The script samples each task's perturbation from a predefined candidate pool, e.g., alternative object placements for Position, alternative goals for Task, and alternative scenes for Environment.
We then regenerate the initial states for the perturbed scenes and evaluate each task with $50$ trials, each starting from a different initial state.
We use the same step budget for all compared policies. 

To isolate the effect of our data, we train the GT-only baseline ourselves with the same action representation, training steps, and batch size as our model, and evaluate all policies with the same controller, step budget, and perturbed tasks.
Absolute success rates may thus differ from those reported in LIBERO-PRO, while the comparison between policies remains fair.

\emph{LIBERO and SIMPLER-WidowX.}

For the sim-to-sim transfer experiments, 
we follow the benchmarks' standard evaluation setups and change only the robot controller described above.
On LIBERO-Object~\cite{liu2023libero}, each of the $10$ tasks is evaluated on its $50$ predefined initial states.
On SIMPLER-WidowX~\cite{SIMPLER}, each of the $4$ tasks is evaluated on its $24$ predefined episode configurations, 
with the benchmark's default scene, camera, and real-image overlay.
We use a stricter success criterion than the benchmark:
besides the benchmark's own condition that the object is placed on the target,
we also require the gripper to have released the object at the same step,
since the original criterion also counts episodes where the robot places the object but never lets go.

We note that simulation results for the same checkpoint can vary across evaluation runs, 
due to stochastic action sampling and the sensitivity of contact-rich physics.
This variance is more pronounced in SIMPLER-WidowX~\cite{SIMPLER}, where each task is evaluated on only $24$ episodes and a single episode changes the success rate by about $4$ percentage points.
For each checkpoint, we run the evaluation multiple times and report the best run.

\section{Additional Results on Simulation Experiments for the Trained Policy}
\label{app:sim_res}

\subsection{Per-task VLA Success Rates on all $4$ LIBERO-PRO Suites}

We report per-task success rates (\%) of the Pi0.5 model on the pick-and-place tasks of all four LIBERO-PRO~\cite{zhou2025liberopro} suites in Tabs.~\ref{tab:libero_pro_per_task_object}--\ref{tab:libero_pro_per_task_long}.
Each suite is evaluated under the original setting and all five out-of-distribution (OOD) perturbations (Language, Object, Position, Task, and Environment).
Within every block, the left sub-column (\emph{GT}) is Pi0.5~\cite{pi05} trained on the original LIBERO ground-truth demonstrations only, 
and the right sub-column (\emph{SW+GT}) additionally uses our SkillWeaver-generated data to train with the same number of steps.

\begin{table}[htbp]
\centering
\resizebox{\textwidth}{!}{%
\begin{tabular}{l cc cc cc cc cc cc}
\toprule
& \multicolumn{2}{c}{Orig.} & \multicolumn{2}{c}{Lang.} & \multicolumn{2}{c}{Obj.} & \multicolumn{2}{c}{Pos.} & \multicolumn{2}{c}{Task} & \multicolumn{2}{c}{Env.} \\
\cmidrule(lr){2-3}\cmidrule(lr){4-5}\cmidrule(lr){6-7}\cmidrule(lr){8-9}\cmidrule(lr){10-11}\cmidrule(lr){12-13}
Task & GT & SW+GT & GT & SW+GT & GT & SW+GT & GT & SW+GT & GT & SW+GT & GT & SW+GT \\
\midrule
Alphabet soup $\rightarrow$ Basket     & 92.0 & 92.0 & 98.0 & 98.0 & 100.0 & 84.0 & 0.0 & 100.0 & 0.0 & 68.0 & 8.0 & 80.0 \\
Cream cheese $\rightarrow$ Basket      & 94.0 & 98.0 & 76.0 & 100.0 & 96.0 & 96.0 & 0.0 & 98.0 & 100.0 & 98.0 & 52.0 & 90.0 \\
Salad dressing $\rightarrow$ Basket    & 96.0 & 100.0 & 98.0 & 100.0 & 92.0 & 100.0 & 0.0 & 74.0 & 0.0 & 4.0 & 100.0 & 92.0 \\
BBQ sauce $\rightarrow$ Basket         & 100.0 & 94.0 & 94.0 & 92.0 & 100.0 & 94.0 & 0.0 & 100.0 & 0.0 & 0.0 & 100.0 & 98.0 \\
Ketchup $\rightarrow$ Basket           & 92.0 & 100.0 & 100.0 & 100.0 & 100.0 & 94.0 & 0.0 & 6.0 & 0.0 & 42.0 & 34.0 & 94.0 \\
Tomato sauce $\rightarrow$ Basket      & 100.0 & 98.0 & 100.0 & 88.0 & 98.0 & 80.0 & 0.0 & 88.0 & 0.0 & 0.0 & 90.0 & 88.0 \\
Butter $\rightarrow$ Basket            & 94.0 & 100.0 & 98.0 & 98.0 & 96.0 & 100.0 & 0.0 & 98.0 & 0.0 & 0.0 & 14.0 & 60.0 \\
Milk $\rightarrow$ Basket              & 92.0 & 98.0 & 100.0 & 100.0 & 100.0 & 88.0 & 0.0 & 94.0 & 0.0 & 74.0 & 0.0 & 36.0 \\
Chocolate pudding $\rightarrow$ Basket & 92.0 & 78.0 & 88.0 & 94.0 & 94.0 & 84.0 & 0.0 & 30.0 & 0.0 & 84.0 & 0.0 & 92.0 \\
Orange juice $\rightarrow$ Basket      & 98.0 & 100.0 & 100.0 & 100.0 & 98.0 & 100.0 & 0.0 & 82.0 & 0.0 & 62.0 & 90.0 & 96.0 \\
\midrule
Average           & 95.0 & 95.8 & 95.2 & 97.0 & 97.4 & 92.0 & 0.0 & 77.0 & 10.0 & 43.2 & 48.8 & 82.6 \\
\bottomrule
\end{tabular}}
\caption{Per-task success rates (\%) on the \textbf{LIBERO-PRO-Object} suite. \emph{GT}: GT only; \emph{SW+GT}: SkillWeaver Data + GT.}
\label{tab:libero_pro_per_task_object}
\end{table}

\begin{table}[htbp]
\centering
\resizebox{\textwidth}{!}{%
\begin{tabular}{l cc cc cc cc cc cc}
\toprule
& \multicolumn{2}{c}{Orig.} & \multicolumn{2}{c}{Lang.} & \multicolumn{2}{c}{Obj.} & \multicolumn{2}{c}{Pos.} & \multicolumn{2}{c}{Task} & \multicolumn{2}{c}{Env.} \\
\cmidrule(lr){2-3}\cmidrule(lr){4-5}\cmidrule(lr){6-7}\cmidrule(lr){8-9}\cmidrule(lr){10-11}\cmidrule(lr){12-13}
Black bowl location & GT & SW+GT & GT & SW+GT & GT & SW+GT & GT & SW+GT & GT & SW+GT & GT & SW+GT \\
\midrule
Between plate \& ramekin   & 100.0 & 100.0 & 92.0 & 94.0 & 96.0 & 96.0 & 90.0 & 96.0 & 0.0 & 2.0 & 96.0 & 96.0 \\
Next to ramekin            & 100.0 & 100.0 & 100.0 & 98.0 & 100.0 & 98.0 & 6.0 & 88.0 & 100.0 & 98.0 & 4.0 & 96.0 \\
From table center          & 100.0 & 100.0 & 100.0 & 100.0 & 100.0 & 100.0 & 100.0 & 100.0 & 0.0 & 48.0 & 98.0 & 100.0 \\
On cookie box              & 98.0 & 100.0 & 100.0 & 100.0 & 100.0 & 100.0 & 0.0 & 12.0 & 2.0 & 96.0 & 96.0 & 94.0 \\
In top drawer of cabinet   & 84.0 & 90.0 & 86.0 & 94.0 & 82.0 & 84.0 & 74.0 & 82.0 & 0.0 & 14.0 & 14.0 & 56.0 \\
On ramekin                 & 98.0 & 96.0 & 100.0 & 98.0 & 86.0 & 90.0 & 96.0 & 98.0 & 100.0 & 100.0 & 0.0 & 6.0 \\
Next to cookie box         & 100.0 & 98.0 & 100.0 & 100.0 & 100.0 & 88.0 & 0.0 & 0.0 & 100.0 & 100.0 & 96.0 & 98.0 \\
On stove                   & 98.0 & 100.0 & 98.0 & 100.0 & 100.0 & 100.0 & 0.0 & 98.0 & 66.0 & 96.0 & 20.0 & 42.0 \\
Next to plate              & 98.0 & 100.0 & 88.0 & 90.0 & 94.0 & 94.0 & 100.0 & 100.0 & 26.0 & 72.0 & 96.0 & 94.0 \\
On wooden cabinet          & 88.0 & 92.0 & 94.0 & 88.0 & 98.0 & 90.0 & 0.0 & 42.0 & 100.0 & 98.0 & 4.0 & 42.0 \\
\midrule
Average                    & 96.4 & 97.6 & 95.8 & 96.2 & 95.6 & 94.0 & 46.6 & 71.6 & 49.4 & 72.4 & 52.4 & 72.4 \\
\bottomrule
\end{tabular}}
\caption{Per-task success rates (\%) on the \textbf{LIBERO-PRO-Spatial} suite. \emph{GT}: GT only; \emph{SW+GT}: SkillWeaver Data + GT.}
\label{tab:libero_pro_per_task_spatial}
\end{table}

\begin{table}[htbp]
\centering
\resizebox{\textwidth}{!}{%
\begin{tabular}{l cc cc cc cc cc cc}
\toprule
& \multicolumn{2}{c}{Orig.} & \multicolumn{2}{c}{Lang.} & \multicolumn{2}{c}{Obj.} & \multicolumn{2}{c}{Pos.} & \multicolumn{2}{c}{Task} & \multicolumn{2}{c}{Env.} \\
\cmidrule(lr){2-3}\cmidrule(lr){4-5}\cmidrule(lr){6-7}\cmidrule(lr){8-9}\cmidrule(lr){10-11}\cmidrule(lr){12-13}
Task & GT & SW+GT & GT & SW+GT & GT & SW+GT & GT & SW+GT & GT & SW+GT & GT & SW+GT \\
\midrule
Bowl $\rightarrow$ stove              & 100.0 & 100.0 & 100.0 & 100.0 & 94.0 & 56.0 & 62.0 & 60.0 & 0.0 & 38.0 & 100.0 & 100.0 \\
Wine bottle $\rightarrow$ top of cab. & 92.0 & 82.0 & 94.0 & 94.0 & 96.0 & 84.0 & 2.0 & 18.0 & 30.0 & 60.0 & 60.0 & 90.0 \\
Bowl $\rightarrow$ top of cabinet     & 100.0 & 98.0 & 100.0 & 96.0 & 38.0 & 30.0 & 0.0 & 86.0 & 0.0 & 26.0 & 86.0 & 100.0 \\
Cream cheese $\rightarrow$ bowl       & 96.0 & 98.0 & 96.0 & 94.0 & 46.0 & 70.0 & 0.0 & 82.0 & 28.0 & 68.0 & 100.0 & 96.0 \\
Bowl $\rightarrow$ plate              & 100.0 & 100.0 & 100.0 & 100.0 & 78.0 & 32.0 & 4.0 & 72.0 & 24.0 & 34.0 & 100.0 & 84.0 \\
Wine bottle $\rightarrow$ rack        & 90.0 & 38.0 & 90.0 & 42.0 & 36.0 & 42.0 & 0.0 & 44.0 & 30.0 & 78.0 & 2.0 & 2.0 \\
\midrule
Average                               & 96.3 & 86.0 & 96.7 & 87.7 & 64.7 & 52.3 & 11.3 & 60.3 & 18.7 & 50.7 & 74.7 & 78.7 \\
\bottomrule
\end{tabular}}
\caption{Per-task success rates (\%) on the \textbf{LIBERO-PRO-Goal} suite. \emph{GT}: GT only; \emph{SW+GT}: SkillWeaver Data + GT.}
\label{tab:libero_pro_per_task_goal}
\end{table}

\begin{table}[htbp]
\centering
\resizebox{\textwidth}{!}{%
\begin{tabular}{l cc cc cc cc cc cc}
\toprule
& \multicolumn{2}{c}{Orig.} & \multicolumn{2}{c}{Lang.} & \multicolumn{2}{c}{Obj.} & \multicolumn{2}{c}{Pos.} & \multicolumn{2}{c}{Task} & \multicolumn{2}{c}{Env.} \\
\cmidrule(lr){2-3}\cmidrule(lr){4-5}\cmidrule(lr){6-7}\cmidrule(lr){8-9}\cmidrule(lr){10-11}\cmidrule(lr){12-13}
Task & GT & SW+GT & GT & SW+GT & GT & SW+GT & GT & SW+GT & GT & SW+GT & GT & SW+GT \\
\midrule
Alphabet soup \& tomato sauce $\rightarrow$ basket      & 84.0 & 58.0 & 86.0 & 74.0 & 94.0 & 56.0 & 6.0 & 70.0 & 2.0 & 86.0 & 32.0 & 36.0 \\
Cream cheese \& butter $\rightarrow$ basket             & 88.0 & 98.0 & 94.0 & 100.0 & 82.0 & 96.0 & 10.0 & 88.0 & 8.0 & 62.0 & 4.0 & 82.0 \\
White \& yellow-white mugs $\rightarrow$ plates         & 86.0 & 30.0 & 92.0 & 56.0 & 98.0 & 46.0 & 0.0 & 44.0 & 0.0 & 10.0 & 4.0 & 22.0 \\
Book $\rightarrow$ caddy compartment                    & 84.0 & 56.0 & 96.0 & 56.0 & 56.0 & 10.0 & 6.0 & 48.0 & 0.0 & 0.0 & 84.0 & 68.0 \\
White mug $\rightarrow$ plate, choc.\ pudding $\rightarrow$ right & 94.0 & 72.0 & 94.0 & 74.0 & 98.0 & 40.0 & 8.0 & 14.0 & 0.0 & 10.0 & 0.0 & 28.0 \\
Alphabet soup \& cream cheese $\rightarrow$ basket      & 36.0 & 70.0 & 50.0 & 84.0 & 60.0 & 68.0 & 0.0 & 40.0 & 0.0 & 58.0 & 18.0 & 68.0 \\
Both moka pots $\rightarrow$ stove                      & 48.0 & 12.0 & 28.0 & 4.0 & 50.0 & 16.0 & 0.0 & 0.0 & 22.0 & 8.0 & 0.0 & 0.0 \\
\midrule
Average                                                 & 74.3 & 56.6 & 77.1 & 64.0 & 76.9 & 47.4 & 4.3 & 43.4 & 4.6 & 33.4 & 20.3 & 43.4 \\
\bottomrule
\end{tabular}}
\caption{Per-task success rates (\%) on the \textbf{LIBERO-PRO-10} suite. \emph{GT}: GT only; \emph{SW+GT}: SkillWeaver Data + GT.}
\label{tab:libero_pro_per_task_long}
\end{table}

\section{Implementation Details on Real-world Experiments}
\label{app:real_res}

\subsection{Policy Training and Evaluation}

Training trajectories are collected in IsaacLab~\cite{mittal2025isaac} with a Franka Panda robot. 
On the real Franka we use joint impedance control via \texttt{frankapy}, also at 10\,Hz.
The policy is trained in the same setting as in simulation (App.~\ref{app:sim_training}), replanning every 10 steps during inference.

\subsection{Large-Scale Simulation Pre-training}

\paragraph{Environment and Task Generation.} 
We create $11K$ scenes with the \emph{broad exploration} mode for task types illustrated in Sec.~\ref{app:task_design}.
Scene distributions over task types are shown in Tab.~\ref{tab:pretrain_scene_dist}.

\begin{table}[htbp]
\centering
\footnotesize
\setlength{\tabcolsep}{6pt}
\begin{tabular}{lcc}
\toprule
Task Type & \# Scenes \\
\midrule
Pick                       & 5000 \\
Place                      & 3500 \\
Orientation-Aware        &  2000 \\
Long-Horizon                   &  500 \\
\midrule
Total                      & 11000 \\
\bottomrule
\end{tabular}
\caption{Distribution of pre-training scenes generated with the \emph{broad exploration} mode across task types.}
\label{tab:pretrain_scene_dist}
\end{table}

\paragraph{Language Augmentation and Annotation.}
We efficiently augment each trajectory's language along four axes, using either Gemini~\cite{gemini2025} or privileged information easily accessible from simulation.

\emph{(i) Task expression}: we generate multiple semantics-preserving paraphrases (e.g., Place the oil bottle into the bucket $\rightarrow$ Deposit the oil bottle so it lands in the bucket).

\emph{(ii) Object attributes}: we substitute referenced objects with fine-grained visual descriptions from asset metadata (e.g. the red apple, the round basket).

\emph{(iii) Spatial relationships}: from per-rollout object poses, we annotate discriminative references among same-category objects along the front--back and left--right axes (e.g., ``the cup on the left'')
based on initial 3D coordinates easily captured in simulation. 

\emph{(iv) Size relations}: within a category, we annotate relative size when object volumes differ enough (e.g., ``the small bowl'') based on asset meshes.

All candidates form an instruction pool for diverse sampling during training, which enables generalization to various language expressions and robust grounding of object references during inference.

\subsection{Post-training for Target Environments}

Given a single image of a target environment, we reconstruct the scene in IsaacLab~\cite{mittal2025isaac} using the \emph{targeted exploration} mode.
We align the camera parameters implicitly by matching the rendered image to the reference, and apply mild randomization to
object poses, background, lighting, and head/wrist camera poses, bridging the sim-to-real gap while preserving the scene layout prior.
We then collect $100$ trajectories per task and apply the same policy distillation pipeline to adapt the pre-trained policy to the target environment and tasks.

Note that scenes in large-scale simulation pre-training and target environment post-training are
built from the broad exploration mode and the targeted exploration mode, respectively, as illustrated in Sec.~\ref{app:scenegen}.

\section{Implementation Details on Ablation Study}
\label{app:ablation_study}

\subsection{Hand-Designed Tools vs. Neural Interaction Skills}
\label{app:rl_skill_ablation}

In this part, we provide more details on the hand-designed tools, which include a grasp proposer ~\cite{murali2025graspgen} plus a motion planner ~\cite{sundaralingam2023curobo},
and our NIS used in the ablation study in Sec.~\ref{sec:ablation_study}.

\paragraph{Asset selection.} We sample $170$ assets from the SceneGen
asset database across five geometry/pose profiles---\emph{round}, \emph{flat},
\emph{elongated upright}, \emph{concave}, and \emph{concave lying}---that together
stress the pick-up difficulty for a parallel-jaw gripper. Within
each profile we sample $10$ assets per category and $1-5$ categories
per profile (Tab.~\ref{tab:asset_selection}). 
Per-trial pose initialization
is profile-specific: any physics-settled orientation drawn from
the asset's pose bank can be used for round and flat; \emph{upright} (canonical
long axis aligned with world~$+z$, filtered from the same bank) for elongated
upright and concave, so that tall bottles and concave containers stand on the table;
and \emph{lying} for concave lying, where the containers lie on their side. 
All $170$ assets are \textbf{out of the training set} of the NIS.

\begin{table}[htbp]
\centering
\footnotesize
\setlength{\tabcolsep}{4pt}
\begin{tabular}{l l p{6.5cm} c}
\toprule
Profile & Geometry / Pose & Categories (Scenario) & \#Assets \\
\midrule
round              & squat, low-aspect convex     & apothecary jar, moisturizer jar, pill bottle, storage canister       & 40 \\
flat               & thin, flat         & butter box, cream cheese box, chocolate pudding box, cookie box, tea box & 50 \\
elongated upright  & long axis vertical           & shampoo, conditioner, lotion, cologne, perfume bottles               & 50 \\
concave            & open container, standing up   & cup, green bowl                                                      & 20 \\
concave lying      & open container, on its side  & cup                                           & 10 \\
\midrule
Total              &                              &                                                                                         & 170 \\
\bottomrule
\end{tabular}
\caption{Asset selection for different geometry/pose profiles.}
\label{tab:asset_selection}
\end{table}

\paragraph{Hand-designed pick-up pipeline.} For each trial we run a fixed
three-stage pipeline. \emph{(i) Grasp proposal:} we render the object's
point cloud in the world frame and query GraspGen~\cite{murali2025graspgen},
which returns up to $\sim\!150$ candidate $6$-DoF panda-hand grasps with
confidence scores. \emph{(ii) Motion planning:} we sort candidates by score
descending and iterate; for each candidate we call cuRobo~\cite{sundaralingam2023curobo}
to plan a collision-free trajectory from the robot's home configuration to the
grasp pose (with the table and the object registered as obstacles), and accept
the \emph{first} candidate whose IK is reachable. If every candidate fails we
mark the trial as a no-IK failure. \emph{(iii) Execution:} we play back the
planned joint trajectory with the gripper open, close the gripper for $30$
control steps to engage the object, then plan and execute a second cuRobo
trajectory that lifts the panda-hand by $25$~cm along the world~$+z$ axis with
the gripper held closed. Success is determined by the same
criterion as the \textsc{Pick} NIS: the object must rise by at least
$15$~cm above its initialization height while a fingertip contact is maintained
for $3$ consecutive control steps.

\subsection{MCTS vs. No MCTS}
\label{app:mcts_ablation}

In this part, we present details of the $5$ long-horizon tasks used in the MCTS vs. linear planning ablation of Sec.~\ref{sec:ablation_study}
in Tab.~\ref{tab:long_horizon_tasks}.

\begin{table}[htbp]
\centering
\footnotesize
\setlength{\tabcolsep}{4pt}
\renewcommand{\arraystretch}{1.2}
\begin{tabular}{c p{4.5cm} p{7.0cm} c}
\toprule
\# & Task Description & Ideal Subgoal Sequence & Minimum Depth \\
\midrule
1 &
Place the cream cheese and the butter into the basket. &
\textit{(i)} pick \texttt{cream\_cheese\_box} $\to$ \textit{(ii)} place into \texttt{basket} $\to$ \textit{(iii)} pick \texttt{butter\_box} $\to$ \textit{(iv)} place into \texttt{basket}. & 4 \\
\addlinespace
2 &
Place the white mug on the left plate and the yellow-and-white mug on the right plate. &
\textit{(i)} pick \texttt{white\_mug} $\to$ \textit{(ii)} place on \texttt{left\_plate} $\to$ \textit{(iii)} pick \texttt{yellow\_and\_white\_mug} $\to$ \textit{(iv)} place on \texttt{right\_plate}. & 4 \\
\addlinespace
3 &
Straighten the lying mug and place a fork inside it. &
\textit{(i)} pick lying \texttt{mug} $\to$ \textit{(ii)} place upright on \texttt{table} $\to$ \textit{(iii)} pick \texttt{fork} $\to$ \textit{(iv)} place into \texttt{mug}. & 4 \\
\addlinespace
4 &
Straighten the lying cup and place a toothbrush inside it. &
\textit{(i)} pick lying \texttt{cup} $\to$ \textit{(ii)} place upright on \texttt{table} $\to$ \textit{(iii)} pick \texttt{toothbrush} $\to$ \textit{(iv)} place into \texttt{cup}. & 4 \\
\addlinespace
5 &
Straighten the lying utensil holder and place a butter knife inside it. &
\textit{(i)} pick lying \texttt{utensil\_holder} $\to$ \textit{(ii)} place upright on \texttt{table} $\to$ \textit{(iii)} pick \texttt{butter\_knife} $\to$ \textit{(iv)} place into \texttt{utensil\_holder}. & 4 \\
\bottomrule
\end{tabular}
\caption{The $5$ long-horizon tasks used in the MCTS vs. linear planning ablation. All tasks require at least $4$ skill steps to complete.}
\label{tab:long_horizon_tasks}
\end{table}

\subsection{Verifier vs. No Verifier}
\label{app:verifier_ablation}

\paragraph{Elaboration on evaluated settings.}\mbox{}

\emph{Search guidance.} Under matched maximum-depth budgets
$K\!\in\!\{4,5,6\}$, we compare three configurations:
\textbf{(i) w/~Verifier, w/o~Rollout} --- the verifier scores every newly
expanded node from the visual observations and the structured scene
description, and that score is back-propagated along the path as the node's
reward; UCB selection is performed over all frontier nodes and no skill
rollouts are taken;
\textbf{(ii) w/o~Verifier, w/o~Rollout} --- no intermediate reward is provided,
and frontier selection degenerates to uniform-random;
\textbf{(iii) w/o~Verifier, w/~Rollout} --- the standard MCTS recipe: each new
node is value-estimated by executing the downstream skill sequence until a
terminal state, and the terminal value is back-propagated. All three
configurations share the same VLM agent and NIS library, searched $10$ times per~$K$, and we report
mean success rate together with wall-clock time per successful trajectory.

\emph{Success detection.}
We evaluate two conditions:
\textbf{(i) Image-only} --- the prompt contains only the task instruction and
rendered \texttt{front} and \texttt{top} views of the initial and final scenes, four $512\!\times\!512$ images in total;
\textbf{(ii) +~Privileged States} --- on top of the same four images we
append privileged simulation information, including per-object 3D center coordinates, 3D axis-aligned bounding boxes, and the gripper's current
contact list.

\paragraph{Task selection.}\mbox{}

\emph{Search guidance.} We use the $6$ long-horizon tasks listed in
Tab.~\ref{tab:verifier_search_tasks}. All require at least $4$ skill steps
and combine different objects, so that pick, place, and
straighten subgoals must interleave. Per configuration, each task is searched
$10$ times.

\begin{table}[htbp]
\centering
\footnotesize
\setlength{\tabcolsep}{4pt}
\renewcommand{\arraystretch}{1.2}
\begin{tabular}{c p{4.7cm} p{6.6cm} c}
\toprule
\# & Task & Ideal Subgoal Sequence & Min.\ Depth \\
\midrule
1 & Fork $\to$ Lying Mug &
\textit{(i)} pick lying \texttt{mug} $\to$ \textit{(ii)} place upright on \texttt{table} $\to$ \textit{(iii)} pick \texttt{fork} $\to$ \textit{(iv)} place into \texttt{mug} & 4 \\
\addlinespace
2 & Toothbrush $\to$ Lying Cup &
\textit{(i)} pick lying \texttt{cup} $\to$ \textit{(ii)} place upright on \texttt{table} $\to$ \textit{(iii)} pick \texttt{toothbrush} $\to$ \textit{(iv)} place into \texttt{cup} & 4 \\
\addlinespace
3 & Knife $\to$ Lying Utensil Holder &
\textit{(i)} pick lying \texttt{utensil\_holder} $\to$ \textit{(ii)} place upright on \texttt{table} $\to$ \textit{(iii)} pick \texttt{butter\_knife} $\to$ \textit{(iv)} place into \texttt{utensil\_holder} & 4 \\
\addlinespace
4 & Two Boxes $\to$ Basket &
\textit{(i)} pick \texttt{cream\_cheese\_box} $\to$ place into \texttt{basket} $\to$ \textit{(ii)} pick \texttt{butter\_box} $\to$ place into \texttt{basket} & 4 \\
\addlinespace
5 & Two Mugs $\to$ Two Plates &
\textit{(i)} pick \texttt{white\_mug} $\to$ place on \texttt{left\_plate} $\to$ \textit{(ii)} pick \texttt{yellow\_and\_white\_mug} $\to$ place on \texttt{right\_plate} & 4 \\
\addlinespace
6 & Pudding $\to$ Side; Mug $\to$ Plate &
\textit{(i)} pick \texttt{chocolate\_pudding\_box} $\to$ place on side $\to$ \textit{(ii)} pick \texttt{mug} $\to$ place on \texttt{plate} & 4 \\
\bottomrule
\end{tabular}
\caption{The $6$ long-horizon tasks used in the verifier search-guidance ablation. Each task is searched $10$ times per budget $K\!\in\!\{4,5,6\}$.}
\label{tab:verifier_search_tasks}
\end{table}

\emph{Success Detection.} For pick, place, orientation-aware place (e.g. pen into pen holder), 
and straighten objects up, 
we sample $50$ tasks each for success detection accuracy evaluation, performing one search per task. 

\paragraph{Hand-crafted GT Success Labels.} To avoid using the verifier's own judgment as the ground truth, 
we hand-label success for each of the evaluated trajectories by matching their final states against hand-designed rules tailored to their task type.

\begin{itemize}
  \item Pick up rule: objects must be lifted by at least $15$\,cm above their initial height, and a fingertip contact must be maintained between the gripper and the object.
  \item Place rule: The center of the source object must be within the target container's bounding box.
  \item Orientation-related rule: in orientation-related scenarios, such as placing elongated objects into thin containers or straightening lying objects, 
  we also require the object's local $z$-axis to be aligned with the world $+z$ axis within a cosine-similarity threshold.
\end{itemize}

\subsection{Memory vs. No Memory}
\label{app:memory_ablation}

We ablate the memory module on $4$ challenging tasks listed in Tab.~\ref{tab:memory_tasks}, 
all of them chosen to stress \emph{orientation reasoning} and \emph{spatial reasoning}.

For each task we compare MCTS \textbf{w/~Memory} against \textbf{w/o~Memory}
under a fixed maximum search depth $K\!=\!6$ and $10$ independent searches
per configuration. Both runs share the same VLM agent, verifier, NIS
library, and per-search compute budget.

\begin{table}[htbp]
\centering
\footnotesize
\setlength{\tabcolsep}{4pt}
\renewcommand{\arraystretch}{1.2}
\begin{tabular}{c p{3.0cm} p{4.8cm} p{4.6cm} c}
\toprule
\# & Task & Ideal Subgoal Sequence & Why Challenging & Min.\ Depth \\
\midrule
1 & \texttt{Pudding $\to$ Side} (LIBERO-$10$ task~$6$) &
\textit{(i)} pick \texttt{chocolate\_pudding} $\to$ \textit{(ii)} place on the designated side of the reference object &
the spatial relation ``side'' must be resolved \emph{relative} to the reference object's front; ambiguous from the agent's egocentric view alone & 2 \\
\addlinespace
2 & \texttt{Bottle Upright} &
\textit{(i)} pick lying \texttt{bottle} $\to$ \textit{(ii)} place upright on \texttt{table} &
requires selecting the \texttt{place\_with\_orientation} skill and a narrow upright drop pose; \texttt{place} alone is not sufficient & 2 \\
\addlinespace
3 & \texttt{Toothbrush $\to$ Cup} &
\textit{(i)} pick \texttt{toothbrush} $\to$ \textit{(ii)} place into upright \texttt{cup} &
thin tall object into a narrow opening; only a near-vertical insertion succeeds geometrically & 2 \\
\addlinespace
4 & \texttt{Toothbrush $\to$ Lying Cup} &
\textit{(i)} pick lying \texttt{cup} $\to$ \textit{(ii)} place upright on \texttt{table} $\to$ \textit{(iii)} pick \texttt{toothbrush} $\to$ \textit{(iv)} place into \texttt{cup} &
requires the \emph{non-obvious prerequisite} of standing the cup up before placing; an agent without memory keeps proposing direct ``place into lying cup'' and exhausts the budget & 4 \\
\bottomrule
\end{tabular}
\caption{Tasks used in the memory ablation. All searches use $K\!=\!6$ and $10$ independent runs per configuration. The first three are depth-$2$ tasks whose failure is local (orientation / skill selection); the fourth is depth-$4$ and requires discovering a prerequisite step.}
\label{tab:memory_tasks}
\end{table}

\section{Automated Simulation Environment Pipeline}
\label{app:scenegen}

\subsection{Asset Generation}

\paragraph{Generating Simulation-ready Assets from Scratch.}
For each of two indoor scenarios, \emph{Kitchen} and \emph{Bathroom}, we prompt Gemini-2.5-Flash~\cite{Gemini}
to propose object categories serving as \emph{manipulated objects} or \emph{containers},
synthesize multiple diverse instance images per category with Nano-Banana~\cite{gemini2025},
and reconstruct each into a 3D visual mesh with SAM3D-Objects~\cite{sam3d2025}.
We then decompose every visual mesh into convex collision components with CoACD~\cite{wei2022coacd},
which preserves the functionality of concave objects (\emph{e.g.}, mugs, bowls) for accurate contact simulation.

We restrict the categories to objects whose geometry and physical size are suitable for 
manipulation by a single-arm robot equipped with a parallel-finger gripper. 
Specifically, manipulated objects are required to have at least one graspable dimension 
below 8\,cm, or to contain thin graspable structures such as rims, handles, or elongated 
parts that can be reliably grasped by parallel fingers. The asset library also contains assets reconstructed from images in the LIBERO-PRO benchmark~\cite{zhou2025liberopro}.

The asset generation pipeline and examples of both visual and collision meshes are illustrated in Fig.~\ref{fig:asset_gen_pipeline}. 
We also present statistics of the resulting asset library in Tab.~\ref{tab:asset_library_stats} and Gemini prompts in Sec.~\ref{vlm_prompt_asset}.

\begin{longtable}{p{0.16\linewidth} >{\footnotesize\itshape\raggedright\arraybackslash}p{0.50\linewidth} r r}
\caption{Asset library statistics. We list representative categories for each scenario and the number of reconstructed assets per category.}
\label{tab:asset_library_stats} \\
\toprule
Scenario & {\normalfont\normalsize Example Categories (\# assets)} & Total Categories & Total Assets \\
\midrule
\endfirsthead

\toprule
Scenario & {\normalfont\normalsize Example Categories (\# assets)} & Total Categories & Total Assets \\
\midrule
\endhead

Kitchen &
apple (15), avocado (15), banana (15), 
beer bottle (100), bell pepper (15), bottle opener (15), 
bowl (100), butter dish (15), butter knife (65), 
cereal box (100), cheese grater (15), coffee pot (27), 
cutting board (12), digital scale (15), 
drinking glass (100), food can (100), 
gravy boat (15), basket (50), bucket (50), hot sauce bottle (50), jam jar (100), 
juice carton (100), milk carton (100), ...
& 84 & 4306 \\

Bathroom &
aftershave bottle (50), apothecary jar (50), 
beard oil bottle (50), cologne bottle (100), 
conditioner bottle (100), contact lens case (50),
eye drops bottle (50), face wash bottle (100), foundation bottle (50), 
hair brush (100), hair dryer (100), ...
& 44 & 3300 \\

LIBERO & 
alphabet soup can (160), white mug (100), 
bbq sauce bottle (140), book (20), butter (120), 
chocolate pudding (130), cookie box (50), ...
& 31 & 2586 \\

\midrule
\textbf{Total} & -- & \textbf{159} & \textbf{10192} \\
\bottomrule
\end{longtable}

\begin{center}
\includegraphics[width=\linewidth]{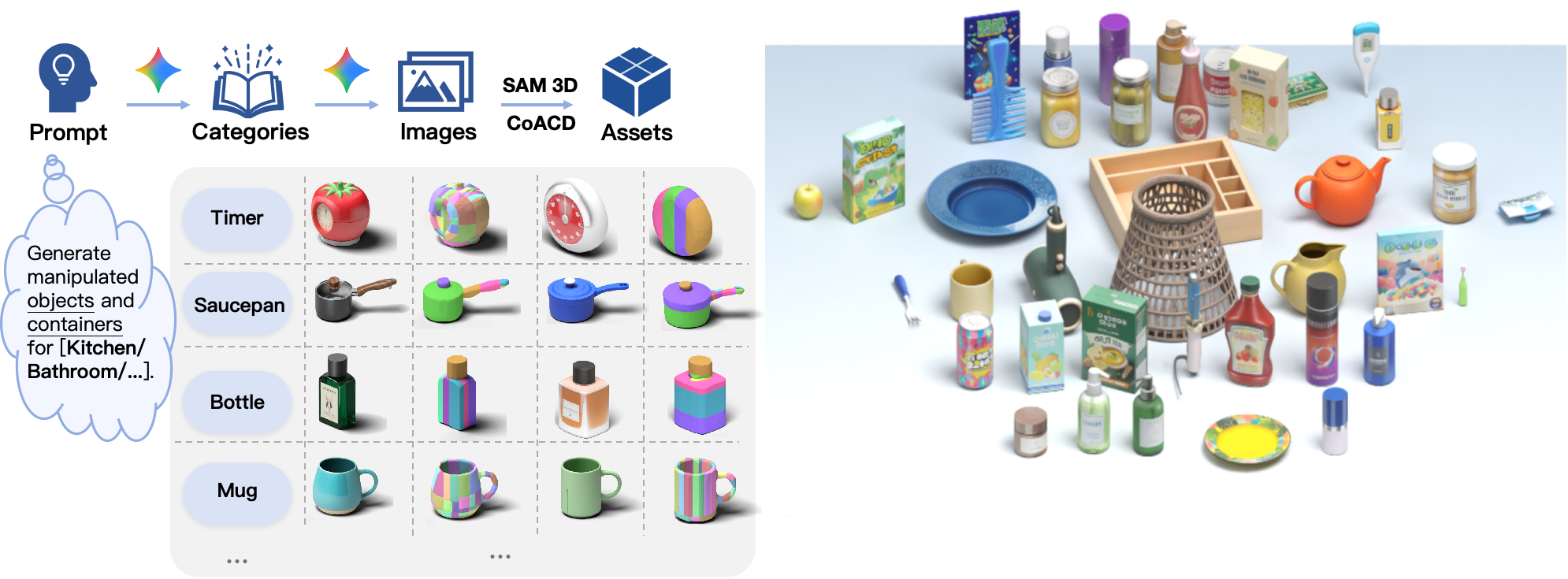}
\captionof{figure}{Left: Our pipeline for generating directly simulatable assets from scratch.
Right: A subset of our asset library.}
\label{fig:asset_gen_pipeline}
\end{center}

\paragraph{Asset Annotation.}
We prompt Gemini~\cite{Gemini} to annotate each reconstructed asset with the semantic, geometric, and physical metadata required for scene composition, physics simulation, and skill learning.
The annotations are produced in a staged pipeline in which later prompts consume earlier outputs, and are stored in each asset's metadata:
\begin{itemize}
\item \textbf{Fine-grained visual attributes.} From the object image, the VLM produces short shape/color/texture phrases (\emph{e.g.}, a \textit{rectangular milk carton}, \textit{white milk carton}, and \textit{cow-graphed milk carton}).
\item \textbf{Metric size.} The VLM estimates a plausible real-world range for the object's longest axis while reasoning explicitly about graspability by the small Panda gripper; we sample a size within this range to rescale the mesh and inject scale diversity (\emph{e.g.}, an \textit{apple} at ${\sim}0.065$\,m versus a \textit{wine bottle} at ${\sim}0.18$\,m).
\item \textbf{Physical properties.} Conditioned on the estimated size, the VLM assigns a mass and friction coefficient that are written directly into the simulation asset (\emph{e.g.}, a teapot at $0.5$\,kg with friction $1.0$).
\item \textbf{Canonical axis semantics.} We render three orthographic views (front/top/side) and ask the VLM to describe the geometric meaning of each positive axis, 
enabling semantic orientation alignment across instances of a category (\emph{e.g.}, for a teapot $+X$ points \textit{from the handle through the body to the spout} and $+Z$ \textit{from the base to the lid knob}).
\item \textbf{Stable Orientation.} Reasoning over the axis descriptions, the VLM determines stable ways to orient the object on a table (\emph{e.g.}, a \textit{bowl} should be placed upright to hold objects, while a \textit{wine bottle} can either stand upright or lie on its side).
Based on this prior, we sample a valid orientation library for each asset.
\end{itemize}

\subsection{Scene Generation}

\paragraph{Broad Exploration Mode.}

To efficiently generate diverse scenes at scale, we compose tabletop layouts procedurally. 
For each scene, we randomly extract a set of assets from the annotated asset library, 
spanning multiple object categories. 
Each sampled object is assigned a random location on the table and an orientation drawn from its valid orientation library, 
to be placed in a physically stable pose. 
To guarantee a physically valid arrangement, 
we run oriented bounding-box (OBB) collision detection between all objects via a separating-axis test, 
and resample the location and orientation of any colliding object until a collision-free layout is obtained. 
This procedure is highly efficient and \emph{generates tens of thousands of distinct scenes within half an hour.}

\paragraph{Targeted Exploration Mode.}

To enable data collection in a specific environment of interest, we
reconstruct manipulatable objects with SAM3D-Objects~\cite{sam3d2025} directly from a single image of that environment.
A scale-aware pose estimation model~\cite{geng2025oneview} is then applied to further align the absolute scale and 6-DoF pose of each reconstructed object.
We use ~\cite{xu2025pixelperfectdepth} to acquire depth maps required for pose recovery. We empirically find that in challenging scenarios, 
the position estimation is generally more accurate than orientation, and therefore discard the estimated orientation and instead 
sample random orientations from the valid library obtained in asset annotation, while keeping the estimated position fixed.

\subsection{Environment Randomization}
\label{app:env_rand}

To further diversify the simulation environments and ensure the downstream policies' robustness to distribution shift, 
we introduce a structured environment randomization pipeline that systematically perturbs the scene by 
decomposing the environment into seven independently controllable components: scene layout, object assets, background, lighting, camera pose, robot initialization, and task specification. 
Each factor is randomized while preserving task semantics, enabling controlled diversity without breaking physical or functional consistency.

{\small
\setlength{\tabcolsep}{6pt}
\renewcommand{\arraystretch}{1.2}
\captionsetup{font=normalsize,position=below}
\setlength{\LTcapwidth}{\textwidth}
\begin{longtable}{@{}l p{0.40\linewidth} p{0.34\linewidth}@{}}
\toprule
\textbf{Factor} & \textbf{What is randomized} & \textbf{Range / distribution} \\
\midrule
\endfirsthead
\toprule
\textbf{Factor} & \textbf{What is randomized} & \textbf{Range / distribution} \\
\midrule
\endhead
\bottomrule
\endfoot
\bottomrule
\caption{Factorized augmentation dimensions in the environment generation pipeline. Each factor is
randomized per asset or scene. $\mathcal{U}$ denotes a uniform and $\mathcal{T}(a,c,b)$ a triangular
distribution with support $[a,b]$ and mode $c$.}
\label{tab:augmentation_summary} \\
\endlastfoot
Scene Layout &
6-DoF object poses are drawn from a precomputed collision-free layout bank, together with the table and floor heights. &
Robot frame position $x\in[-0.3, 0.27],y\in[-0.38, 0.38]$ \newline
Orientation: category dependent, sampled from a valid orientation library \newline
Table height $\sim\mathcal{U}(-0.05, 0.0)$\,m \newline
Floor height $\sim\mathcal{U}(-1.0, -0.7)$\,m \\
\addlinespace
Asset Attributes &
For each category a different reconstructed instance is sampled, varying geometry, texture, and size along with VLM-annotated physical properties. &
Per-category instance sampling \newline
Size Random Scale $\sim\mathcal{U}(0.8, 1.5)$ \newline
Mass: $[0.025, 1.5]kg$ \newline
Friction: $[0.8, 1.1]$ \\
\addlinespace
Background &
Floor, wall, and table assets are swapped per scene from large pools split into train/test. &
Floor Variants: $5$ \newline
Wall Variants: $3122$ \newline
Table Variants: $89$ \\
\addlinespace
Lighting &
A dome light varies global illumination in strength, direction, and textures sourced from public CC0 HDRI libraries. &
Intensity $\sim\mathcal{U}(1500, 4000)$\,nits \newline
Yaw $\sim\mathcal{U}(-\pi, \pi)$ \newline
HDRI Texture Variants: $20$ \newline
Color tint per Channel\ $\sim\mathcal{U}(0.6, 1.4)$ \\
\addlinespace
Camera Pose &
The head camera is sampled on a forward-facing spherical sector around the table center, keeping the workspace in frame. &
Radius $\sim\mathcal{T}(0.8, 1.0, 1.4)$\,m \newline
Azimuth $\sim\mathcal{T}(-20, 0, 20)$ \newline
Elevation $\sim\mathcal{T}(15, 30, 50)^\circ$ \\
\addlinespace
Robot Initialization &
The initial arm configuration is drawn from pools of valid IK solutions obtained by perturbing the end-effector pose of a default arm configuration. &
EE position noise std $\le 5$\,cm \newline
Gripper direction deviation $\le 30^\circ$ \\
\addlinespace
Task Instruction &
A VLM paraphrases each task instruction into diverse natural-language expressions. &
- \\
\end{longtable}}

\section{Neural Interaction Skill Learning}
\label{app:skill_details}

In this section, we provide additional details on the design and training of the NIS with PPO~\cite{schulman2017proximalppo}.

\paragraph{Implementation Details of PPO Policies.}
Each NIS is a closed-loop PPO policy trained with rl\_games.
\textit{Pick} adopts an $80$-D observation, while \textit{Place} appends the goal pose and its residual for a $94$-D observation.
\textit{Open} and \textit{Close} reuse the robot and keypoint terms of \textit{Pick}, 
but replace the object position with the center of the handle or of the movable part.
They further append articulation-related observations, making up an $87$-D observation for prismatic drawers and $93$-D for revolute doors.

We also train \textit{Pick} and \textit{Place} policies on the WidowX robot to solve the SIMPLER-WidowX~\cite{SIMPLER} tasks, with the joint-related observation terms adjusted to its $6$-DoF arm.

All skills output a $7$-D action, namely a $6$-D end-effector delta pose and a $1$-D gripper command, executed at $20$\,Hz, 
and share the PPO hyperparameters listed in Tab.~\ref{tab:rl_io_ppo}.

\begin{table}[htbp]
\centering
\begin{minipage}[t]{0.50\linewidth}
\vspace{0pt}
\centering
\footnotesize
\begin{tabular}{l r}
\toprule
Observation component & Dim \\
\midrule
Joint position (normalized)      & 9 \\
Joint velocity                   & 9 \\
Joint position target            & 9 \\
Hand state (pose\,+\,twist)      & 13 \\
Finger / grasp-site positions    & 9 \\
Object / part center position    & 3 \\
Keypoint--surface distance       & 6 \\
Keypoint closest-point direction & 18 \\
Fingertip contact flags          & 4 \\
\midrule
\textbf{Pick}     & \textbf{80} \\
Goal pose \emph{(Place)}          & 7 \\
Goal pose residual \emph{(Place)} & 7 \\
\textbf{Place}                   & \textbf{94} \\
\midrule
Part joint pos./goal/err./vel. \emph{(Open/Close)} & 4 \\
Motion direction \emph{(Drawer)}                  & 3 \\
Hinge axis\,/\,radial\,/\,tangent \emph{(Door)}   & 9 \\
\textbf{Open/Close (drawer / door)}               & \textbf{87 / 93} \\
\midrule
\multicolumn{2}{l}{\emph{Action} (all skills): \textbf{7}} \\
\quad EE delta pose / gripper    & 6 / 1 \\
\bottomrule
\end{tabular}
\end{minipage}%
\hfill
\begin{minipage}[t]{0.46\linewidth}
\vspace{0pt}
\centering
\footnotesize
\begin{tabular}{l r}
\toprule
PPO hyperparameter & Value \\
\midrule
Parallel envs            & 4096 \\
Horizon length           & 8 \\
Minibatch size           & 1024 \\
Mini-epochs              & 5 \\
Learning rate            & $5{\times}10^{-4}$ \\
LR schedule (KL target)  & adaptive ($0.016$) \\
Discount $\gamma$        & 0.99 \\
GAE $\lambda$            & 0.95 \\
Clip $\epsilon$          & 0.2 \\
Entropy coef.            & $10^{-3}$ \\
Critic coef.             & 4 \\
Gradient norm            & 1.0 \\
MLP units                & 512-512-256-128 \\
Activation               & swish \\
Norm.\ (obs/val/adv)     & yes \\
Max epochs               & 20000 \\
\bottomrule
\end{tabular}
\end{minipage}
\caption{NIS policy I/O and PPO hyperparameters. Rows marked \emph{(Place)} are appended only for the \textit{Place} policy ($+14$);
rows marked \emph{(Open/Close)}, \emph{(Drawer)}, and \emph{(Door)} are appended for the articulation policies ($+7$ for prismatic drawers and $+13$ for revolute doors).
All other observation terms are shared by all skills, where \textit{Open} and \textit{Close} use the handle or movable-part center as the object position.}
\label{tab:rl_io_ppo}
\end{table}

\paragraph{Tailored Skill Designs for Different Object Geometry, Pose, and Articulation Regimes.}

As introduced in Sec.~\ref{sec:rl_train}, we split each skill into geometry-, pose-, and articulation-specific sub-policies, since no single interaction strategy is optimal across objects: a round or horizontally lying object is best grasped from above, an upright elongated object from the side to avoid tipping, an open container around its rim, and a tool by its handle.
Articulated parts further differ in joint type and interaction mode: a drawer slides along its joint axis while a door swings about its hinge,
and opening requires pulling a firmly grasped handle, whereas closing is more robustly achieved by pushing the part with any gripper surface.

Tab.~\ref{tab:skill_policy_design} details the resulting \textsc{Pick}, \textsc{Place}, \textsc{Open}, and \textsc{Close} sub-policies, where each row gives a sub-policy, the object/pose regime it handles, and its tailored object-centric observation and reward design.

{\setlength{\tabcolsep}{5pt}
\renewcommand{\arraystretch}{1.15}
\captionsetup{position=below}
\setlength{\LTcapwidth}{\textwidth}
\begin{longtable}{l l p{4.3cm} p{4.7cm}}
\toprule
Skill & Sub-policy & Object / Pose Regime & Policy Design \\
\midrule
\endfirsthead
\toprule
Skill & Sub-policy & Object / Pose Regime & Policy Design \\
\midrule
\endhead
\bottomrule
\endfoot
\bottomrule
\caption{Specialized NIS sub-policies categorized by object geometry, pose, and articulation type, together with their corresponding policy designs.}
\label{tab:skill_policy_design} \\
\endlastfoot
\textit{Pick}
& General
& Round or horizontally elongated objects that can be grasped from above \textit{(e.g., apple, sponge, short can)}
& Top-grasp policy with rewards for contact and lifting \\
& Upright
& Vertically elongated objects standing upright \textit{(e.g., wine bottle, milk carton)}
& Side-grasp policy with tilt penalty and upright-orientation success constraint \\
& Concave
& Open containers grasped around the rim \textit{(e.g., bowl, mug, utensil holder)}
& Object-centric observation exposes near-rim points \\
& Concave Lying
& Open containers resting on their side, grasped by approaching the opening from the side \textit{(e.g., toppled mug or cup)}
& Object-centric observation exposes the side-facing opening/rim points \\
& Handle
& Objects with a graspable handle \textit{(e.g., moka pot)}
& Object-centric observation exposes VLM-segmented handle points \\
\midrule
\textit{Place}
& General
& Objects placed at a target position without strict orientation constraints \textit{(e.g., apple on table, onion on plate)}
& Reward encourages reducing goal-position residual and stable release; orientation is unconstrained \\
& Upright
& Objects whose goal pose requires an upright orientation \textit{(e.g., place wine bottle upright)}
& Additional upright-orientation reward and success constraint \\
\midrule
\textit{Open}
& Drawer
& Prismatic parts opened by pulling the handle along the joint axis \textit{(e.g., cabinet drawer)}
& Starts from a gripper pose near a sampled handle grasp; opening reward requires a stable pinch, and losing the grasp terminates the episode \\
& Door
& Revolute parts opened by pulling the handle along the hinge arc \textit{(e.g., microwave door, cabinet door)}
& Starts from a gripper pose near a sampled handle grasp; hinge-aware observation and a start-state curriculum annealed from near-open to closed \\
\midrule
\textit{Close}
& Drawer
& Prismatic parts closed by pushing the front surface \textit{(e.g., cabinet drawer)}
& Starts from a gripper pose near a sampled push pose on the front; rewards contact by any gripper surface, pays closing rewards only while engaged in the pressing pose, and terminates on detachment \\
& Door
& Revolute parts closed by pushing the door face \textit{(e.g., microwave door, cabinet door)}
& Same push design as the drawer, with hinge-aware observation \\
\end{longtable}}

\newcommand{\relu}[1]{\left(#1\right)^{+}}
\newcommand{\ind}[1]{\mathbb{1}\!\left[#1\right]}
\newcommand{\rnorm}[1]{\left\lVert #1 \right\rVert}

\paragraph{Reward Formulations.}\mbox{}

We present the core reward terms for all skills in Tab.~\ref{tab:reward_terms}, where $w_{(\cdot)}$ are scalar weights and $(\cdot)^+\!=\max(\cdot,0)$.
Progress terms are measured against running extrema ($m_{t-1},M_{t-1}$) from previous steps, so each term pays only for new progress.
$\bm{p}_o,\bm{p}_e$ are the object-center and grasp-site positions, $\ell$ is the object's height gain, $n_c$ is the fingertip-contact count, and $\bar d$ is the mean fingertip--object distance.
For articulated parts, $q$ and $q^{*}$ are the normalized joint position and its goal, $\pi$ indicates a two-sided pinch on the handle, $\hat{\bm d}$ is the target motion direction, and $d_g$ is the distance from the gripper surface to the part.
Remaining symbols denote indicator flags or auxiliary geometric quantities, as named by the term labels.

{\footnotesize
\renewcommand{\arraystretch}{1.4}
\captionsetup{font=normalsize,position=below}
\setlength{\LTcapwidth}{\textwidth}
\begin{longtable}{@{} >{\bfseries}l | l @{}}
\toprule
Skill & Reward terms \\
\midrule
\endfirsthead
\toprule
Skill & Reward terms \\
\midrule
\endhead
\bottomrule
\endfoot
\bottomrule
\caption{Core reward terms per skill. \textit{Pick (non-upright)} covers the \textit{General}, \textit{Concave}, \textit{Concave-lying}, and \textit{Handle} sub-policies, 
which share the same reward and differ only in their object-centric observation. For \textit{Open}, doors replace the force-direction term with the normalized torque about the hinge axis; 
\textit{Close} uses identical rewards for drawers and doors.}
\label{tab:reward_terms} \\
\endlastfoot
Pick (non-upright) &
$\begin{aligned}[t]
R_{oe}   &= w_{oe}\,\relu{m_{t-1}-\rnorm{\bm{p}_o-\bm{p}_e}}                  && (\texttt{approach})\\
R_{c}    &= w_{c}\,\mathrm{clip}(n_c,0,2)\cdot g                              && (\texttt{contact})\\
R_{\ell} &= w_{\ell}\,\relu{\ell-M_{t-1}}\,\ind{n_c>0}\,\mathrm{xy\_ok}        && (\texttt{lift progress})\\
R_{\mathrm{lifted}} &= w_{1}\ind{\mathrm{raw}_{xy}\wedge\neg f}
                      + w_{2}\ind{\mathrm{pose}_{xy}\wedge u}\\
         &\quad + w_{3}\ind{\mathrm{pose}_{xy}\wedge\neg f\wedge u}            && (\texttt{lift milestones})
\end{aligned}$ \\
\midrule
Pick (upright) &
$\begin{aligned}[t]
&\text{Same terms as Pick, with the contact gate } g=\ind{s\ge s_{\mathrm{thr}}}\\
&\text{enforcing a side approach } (s=1-|\hat{\bm n}\cdot\hat{\bm z}|),\\
&\text{plus a wrist-tilt penalty and an upright success constraint.}
\end{aligned}$ \\
\midrule
Place &
$\begin{aligned}[t]
R_{\mathrm{pos}}  &= w_{p}\,\relu{m^{p}_{t-1}-\rnorm{\Delta\bm{p}^{xy}}}       && (\texttt{goal position})\\
R_{\mathrm{dist}} &= w_{d}\,\mathrm{clip}(\bar d,0,0.25)                       && (\texttt{hold/grasp})
\end{aligned}$ \\
\midrule
Place (upright) &
$\begin{aligned}[t]
&\text{Same terms as Place, plus an upright \emph{tilt} reward}\\
R_{\mathrm{tilt}} &= w_{r}\,\relu{m^{r}_{t-1}-\delta_{\mathrm{tilt}}},
   \quad \delta_{\mathrm{tilt}}=\angle(\hat{\bm z}_o,\hat{\bm z})\\
&\text{constraining only the object up-vector to within a small}\\
&\text{tilt tolerance of vertical (no full-orientation match).}
\end{aligned}$ \\
\midrule
Open (drawer / door) &
$\begin{aligned}[t]
R_{c} &= w_{c}\,\mathrm{clip}(n_c,0,2)\,(1+\pi) + w_{\pi}\,\pi && (\texttt{contact/pinch})\\
R_{\mathrm{dist}} &= w_{d}\,\mathrm{clip}(\bar d,0,0.25) && (\texttt{reach handle})\\
R_{\mathrm{open}} &= w_{o}\,\relu{q-M_{t-1}}\,\ind{\pi\wedge\mathrm{dir\_ok}} && (\texttt{open progress})\\
R_{\mathrm{opened}} &= w_{1}\ind{q\ge q^{*}\wedge\neg f} + w_{2}\ind{q\ge q^{*}} && (\texttt{open milestones})\\
R_{\mathrm{align}} &= w_{F}\cos\angle(\bm F_{o},\hat{\bm d}) && (\texttt{force direction})\\
&\quad + e^{-\alpha d_{\min}}\big(w_{n}\,\hat{\bm n}\!\cdot\!(-\hat{\bm d}) + w_{\perp}(1-|\hat{\bm f}\!\cdot\!\hat{\bm h}|)\big) && (\texttt{grasp pose})
\end{aligned}$ \\
\midrule
Close (drawer / door) &
$\begin{aligned}[t]
R_{\mathrm{close}} &= w_{cl}\,\relu{m_{t-1}-q}\,\gamma && (\texttt{close progress})\\
R_{\mathrm{closed}} &= \big(w_{1}\ind{q\le q^{*}\wedge\neg f} + w_{2}\ind{q\le q^{*}}\big)\,\gamma && (\texttt{close milestones})\\
R_{\mathrm{touch}} &= w_{t}\,\ind{d_g<\epsilon_t} && (\texttt{touch})\\
R_{\mathrm{app}} &= -w_{a}\,\mathrm{clip}(d_g,0,d_{\max}) && (\texttt{press on part})\\
R_{\mathrm{hold}} &= w_{R}\cos\theta_R + w_{\omega}\rnorm{\bm a_{\mathrm{rot}}} && (\texttt{pose hold})\\
\gamma &= \ind{d_g<d_{\mathrm{gate}}}\,\ind{\cos\theta_R\ge c_{\mathrm{gate}}} && (\texttt{engagement gate})
\end{aligned}$ \\
\end{longtable}}

\section{Implementation Details on MCTS-based Data Collection}
\label{app:mcts_details}

\subsection{Task Designs}
\label{app:task_design}

\model{} generates in-the-wild demonstrations across $5$ task families:
\begin{itemize}
    \item \emph{\textbf{Pick}}: grasp an object and lift it above a height threshold.
    \item \emph{\textbf{Place}}: pick up an object and place it at a specified location, e.g., on or inside a target object, without orientation requirements.
    \item \emph{\textbf{Orientation-aware}}: bring an object to a specific final orientation inferred by the agent, e.g., straightening up a lying bottle, pouring from a cup, or inserting a thin object into a narrow holder with aligned long axes.
    \item \emph{\textbf{Articulation}}: open or close a prismatic or revolute part, such as a drawer or a cabinet or microwave door.
    \item \emph{\textbf{Long-horizon}}: chain multiple steps from the above families.
    A task becomes long-horizon either because the instruction explicitly involves multiple subgoals (e.g., ``put the apple in the microwave and close the door''),
    or because it {\color{tgradB}carries implicit requirements that the agent must infer}:
    e.g., for ``put the toothpaste tube into the mug'' with the mug lying on its side, the agent must first straighten up the mug, although the instruction never mentions it.
\end{itemize}
Tab.~\ref{tab:task_examples} lists example language instructions and the corresponding skill chains for each family.

\DeclareRobustCommand{\orienttag}{\,{\setlength{\fboxsep}{1pt}\colorbox{tgradA!15}{\color{tgradA}\scriptsize\textsf{orient}}}}
{\footnotesize
\renewcommand{\arraystretch}{1.2}
\captionsetup{font=normalsize,position=below}
\setlength{\LTcapwidth}{\textwidth}
\par\vspace{\LTpre}\setlength{\LTpre}{0pt}
\begin{longtable}{l p{0.33\textwidth} p{0.45\textwidth}}
\toprule
Family & Example instruction & Skill chain \\
\midrule
\endfirsthead
\toprule
Family & Example instruction & Skill chain \\
\midrule
\endhead
\bottomrule
\endfoot
\bottomrule
\caption{Example language instructions and the corresponding skill chains for each task family. Skills executed by NIS are in \textbf{bold}, and steps that use \textbf{orientation alignment} are marked with \orienttag.}
\label{tab:task_examples} \\
\endlastfoot
Pick & Pick up the peanut butter jar. & \textit{(i)} \textbf{pick} \texttt{peanut\_butter\_jar} \\
 & Pick up the cheese grater. & \textit{(i)} \textbf{pick} \texttt{cheese\_grater} \\
 & Pick up the utensil holder. & \textit{(i)} \textbf{pick} \texttt{utensil\_holder} \\
\midrule
Place & Put the dental floss case inside the food storage container. & \textit{(i)} \textbf{pick} \texttt{dental\_floss\_case} $\to$ \textit{(ii)} \textbf{place} into \texttt{food\_storage\_container} \\
 & Set down the orange inside the saucer. & \textit{(i)} \textbf{pick} \texttt{orange} $\to$ \textit{(ii)} \textbf{place} into \texttt{saucer} \\
 & Place the butter dish on the digital scale. & \textit{(i)} \textbf{pick} \texttt{butter\_dish} $\to$ \textit{(ii)} \textbf{place} on \texttt{digital\_scale} \\
\midrule
Orientation-aware & Make the lying food can stand upright. & \textit{(i)} \textbf{pick} \texttt{food\_can} $\to$ \textit{(ii)} \textbf{place} upright on \texttt{table}\orienttag \\
 & Set the hot sauce bottle upright. & \textit{(i)} \textbf{pick} \texttt{hot\_sauce\_bottle} $\to$ \textit{(ii)} \textbf{place} upright on \texttt{table}\orienttag \\
 & Position the butter knife inside the utensil holder. & \textit{(i)} \textbf{pick} \texttt{butter\_knife} $\to$ \textit{(ii)} place into \texttt{utensil\_holder} with aligned long axes\orienttag \\
 & Pour from the moka pot into the mug. & \textit{(i)} \textbf{pick} \texttt{moka\_pot} $\to$ \textit{(ii)} move above \texttt{mug} and tilt to pour\orienttag \\
\midrule
Articulation & Open the middle drawer of the wooden cabinet. & \textit{(i)} \textbf{open drawer} of \texttt{wooden\_cabinet} \\
 & Push in the bottom drawer of the white cabinet. & \textit{(i)} \textbf{close drawer} of \texttt{white\_cabinet} \\
 & Swing the microwave door open. & \textit{(i)} \textbf{open door} of \texttt{microwave} \\
 & Close the microwave door. & \textit{(i)} \textbf{close door} of \texttt{microwave} \\
\midrule
Long-horizon & Put the toothpaste tube into the mug.\newline{\color{tgradB}(mug initially lying; straightening must be inferred)} & \textit{(i)} \textbf{pick} \texttt{mug} $\to$ \textit{(ii)} \textbf{place} upright on \texttt{table}\orienttag $\to$ \textit{(iii)} \textbf{pick} \texttt{toothpaste\_tube} $\to$ \textit{(iv)} place into \texttt{mug} with aligned long axes\orienttag \\
 & Pour from the green cup into the cup.\newline{\color{tgradB}(cup initially lying; straightening must be inferred)} & \textit{(i)} \textbf{pick} \texttt{cup} $\to$ \textit{(ii)} \textbf{place} upright on \texttt{table}\orienttag $\to$ \textit{(iii)} \textbf{pick} \texttt{green\_cup} $\to$ \textit{(iv)} move above \texttt{cup} while holding orientation $\to$ \textit{(v)} tilt to pour\orienttag \\
 & Open the top drawer of the cabinet and put the black bowl inside. & \textit{(i)} \textbf{open drawer} of \texttt{wooden\_cabinet} $\to$ \textit{(ii)} \textbf{pick} \texttt{black\_bowl} $\to$ \textit{(iii)} \textbf{place} into \texttt{drawer} \\
 & Put the black bowl in the bottom drawer of the cabinet and close it. & \textit{(i)} \textbf{pick} \texttt{black\_bowl} $\to$ \textit{(ii)} place into \texttt{drawer} $\to$ \textit{(iii)} \textbf{close drawer} of \texttt{white\_cabinet} \\
 & Put the apple in the microwave and close the door. & \textit{(i)} \textbf{pick} \texttt{apple} $\to$ \textit{(ii)} place into \texttt{microwave} with aligned gripper\orienttag $\to$ \textit{(iii)} \textbf{close door} of \texttt{microwave} \\
\end{longtable}}

\subsection{MCTS Hyperparameters}

Tab.~\ref{tab:mcts_hparams} lists the MCTS search and VLM sampling hyperparameters shared by all tasks.
The search budget varies with the number of skill steps a task family requires and is listed in Tab.~\ref{tab:mcts_budget}.

\begin{table}[htbp]
\centering
\footnotesize
\setlength{\tabcolsep}{6pt}
\begin{tabular}{ll p{0.46\linewidth}}
\toprule
Hyperparameter & Value & Description \\
\midrule
\multicolumn{3}{l}{\emph{MCTS search}} \\
\texttt{c\_puct}                    & $0.5$                      & UCB exploration constant \\
\texttt{num\_children\_per\_expand} & $2$                        & Branching factor per expansion \\
\texttt{n\_simulations}             & Tab.~\ref{tab:mcts_budget} & Number of MCTS iterations \\
\texttt{max\_depth}                 & Tab.~\ref{tab:mcts_budget} & Maximum search depth (skill steps) \\
\midrule
\multicolumn{3}{l}{\emph{VLM sampling}} \\
\texttt{temperature}                & $1.0$                      & Sampling temperature \\
\texttt{top\_p}                     & $0.95$                     & Nucleus sampling threshold \\
\texttt{top\_k}                     & $10$                       & Top-$k$ sampling \\
\texttt{max\_new\_tokens}           & $8192$                     & Max tokens generated per prompt \\
\bottomrule
\end{tabular}
\caption{MCTS search and VLM sampling hyperparameters shared by all tasks.}
\label{tab:mcts_hparams}
\end{table}

\begin{table}[htbp]
\centering
\footnotesize
\setlength{\tabcolsep}{6pt}
\begin{tabular}{l p{0.44\linewidth} cc}
\toprule
Task family & Representative tasks & \texttt{n\_simulations} & \texttt{max\_depth} \\
\midrule
Pick                           & Pick up an object                                                                                 & $1$       & $1$ \\
Place / Insert / Straighten-up & Place into or onto a target, insert into a narrow holder, straighten up a lying object          & $3$       & $2$ \\
Pour                           & Pour from a cup into a bowl                                                                       & $3$--$5$  & $2$--$3$ \\
Articulation                   & Open or close a drawer or door                                                                    & $3$       & $1$ \\
Articulation + Place           & Put an object into a drawer or microwave and close it; open a drawer and place an object inside & $4$--$6$  & $3$ \\
Long-horizon                   & Straighten up a lying container before inserting or pouring into it; place multiple objects     & $6$--$7$  & $4$--$6$ \\
Benchmark pick-and-place       & LIBERO-Object, LIBERO-Spatial, LIBERO-Goal, and SIMPLER-WidowX tasks                             & $3$       & $2$ \\
\bottomrule
\end{tabular}
\caption{MCTS search budget per task family. Ranges cover tasks within the same family that require different numbers of skill steps.}
\label{tab:mcts_budget}
\end{table}

\subsection{Data Modality and Representations}

Each trajectory is recorded step by step in IsaacLab~\cite{mittal2025isaac}. At every timestep we store:

\paragraph{Observations.} 
Five RGB camera streams---four head cameras with varying viewpoints and one wrist-mounted camera---providing both global context and a close-up, hand-centric view.

\paragraph{State.} The proprioceptive state at the current timestep, i.e., the $7$ arm joint angles and the gripper finger position.

\paragraph{Action.} The resulting robot configuration \emph{after} executing the step, again represented as the $7$ joint angles and the finger position. 
Depending on the phase, the action is produced either by a motion planner for collision-free mid-air transfer, or by an NIS for contact-rich manipulation; both are recorded in the same joint-space format 
so the downstream policy sees a unified action representation.

\paragraph{Auxiliary information.} 
For downstream applications, we can additionally log accurate 3D-related information extracted from simulation, such as per-step depth maps or the $6$-DoF pose of each object.

\subsection{Goal Orientation Prediction through Axis Alignment}
\label{app:axis_align}

Directly regressing 3D rotations with a VLM is unstable and often geometrically inconsistent,
so we represent goal orientations through semantic axis alignment.
Each asset is annotated in its canonical frame with the semantic meaning of its positive $x$, $y$, and $z$ axes (App.~\ref{app:scenegen}),
and these descriptions are provided to the VLM agent together with the object list.

For an orientation-aware skill, the agent predicts an alignment tuple $(a_h, a_t, s)$ with $a_h, a_t \in \{x, y, z\}$ and $s \in \{+, -\}$,
requiring that, at the end of the skill, axis $a_h$ of the held object points in the same ($+$) or opposite ($-$) direction as axis $a_t$ of the target.
For example, straightening up a lying bottle on a table uses $(z, z, +)$, aligning the bottle's up axis with the table's up axis.
When the target is a movable part, such as an opened door, $a_t$ is expressed in the part's current link frame, which moves with the joint.

Given the current world orientations $R_h$ and $R_t$ of the held object and the target, the goal orientation of the held object is computed analytically as
\begin{equation}
    \mathbf{u} = R_h\,\mathbf{e}_{a_h}, \qquad
    \mathbf{v} = s\,R_t\,\mathbf{e}_{a_t}, \qquad
    R_h^{\star} = \mathrm{Rot}(\mathbf{u} \!\rightarrow\! \mathbf{v})\,R_h,
\end{equation}
where $\mathbf{e}_{a}$ is the unit vector along axis $a$ and $\mathrm{Rot}(\mathbf{u} \!\rightarrow\! \mathbf{v})$ is the minimal rotation mapping $\mathbf{u}$ onto $\mathbf{v}$, obtained in closed form from Rodrigues' rotation formula.
The motion planner then moves the held object to the goal orientation $R_h^{\star}$.

We also allow the agent to instead align an axis of the gripper frame (e.g., its approach axis) with an axis of the target 
for placements where the approach direction of the gripper matters, such as inserting an object into a side-opening container like a microwave.
We apply the same computation to the end-effector orientation $R_{ee}$ to obtain $R_{ee}^{\star}$,
and convert it into the held-object goal through the rigid grasp, $R_h^{\star} = R_{ee}^{\star} R_{ee}^{-1} R_h$.

Fig.~\ref{fig:align_examples} in App.~\ref{app:qualitative_align} visualizes predicted alignments on successful trajectories.

\subsection{Memory Module}
\label{app:memory_module}

The memory module accumulates reusable experience across search episodes in two libraries.
Most of the time, we distill experience into \emph{General rules} 
that capture insights that transfer across tasks, objects, and scenes.
However, when the task is too difficult to solve with general guidance alone, 
we convert past experience into \emph{task-specific strategies} that give a step-by-step recipe for one task.

After a search episode, we give a VLM the extracted trajectory, including the initial object layout, the available skill primitives,
and each step's subgoal, reasoning, action parameters, and resulting image, together with only the final outcome,
and ask it to distill $1$--$2$ general rules or one task strategy.
For a failed trajectory, we prompt the VLM to trace the failure back to the earliest decision that made it inevitable and prescribe an alternative.

At planning time, a selector VLM retrieves the top-$k$ relevant entries for the current instruction and subgoal,
which are injected into the agent's subgoal-planning and action-grounding prompts and the verifier prompts.
(see prompts for memory distillation and retrieval in App.~\ref{sec:memory_prompt})

Tab.~\ref{tab:memory_examples} shows representative entries, each a (key, memory) pair where the key is a short retrieval handle and the memory is the full guidance in text.

\begin{table}[htbp]
\centering
\small
\renewcommand{\arraystretch}{1.2}
\begin{tabular}{@{}p{2.7cm}p{12.5cm}@{}}
\toprule
Key & Memory \\
\midrule
\multicolumn{2}{@{}l}{\emph{General rules}} \\
\addlinespace
place beside target &
Put a source object to a certain side of the target means to put the source onto the vacant spot on the supporting surface beside the target, NOT on the target. \\
\addlinespace
align elongated object for insertion &
When placing a thin, elongated object (spoon, pen, toothbrush) into a narrow-mouth container (cup, mug, vase), the actor's phase-2 grounder MUST align the object's long axis with the container's vertical axis. The judge MUST verify the object's final horizontal bounding box is contained within the container's opening. \\
\addlinespace
prepare container before insertion &
When the destination container (cup, mug, bowl) is lying on its side, the actor's phase-1 planner MUST first create subgoals to pick it up and place it upright. BEFORE attempting to insert another object, the judge MUST verify the container's pose flag is ``upright''. \\
\addlinespace
straighten up objects &
When straightening up an object, you should align the upright axis of the object with the world's vertical axis, and make sure the object is placed back to its original position on the supporting surface. \\
\addlinespace
lay tall objects down in shallow drawers &
When placing a tall object like a bottle into a shallow drawer, if its upright height exceeds the available clearance, the actor's grounder MUST orient it to lie on its side. Its long axis MUST be aligned horizontally, and the judge must verify its new vertical dimension (i.e., its diameter) fits. \\
\addlinespace
place into side-opening container &
When placing an object into a container with a side opening like a microwave or oven, the actor's planner MUST select a skill that controls the gripper's approach direction from the side. Standard top-down \texttt{place} skills are unsuitable and MUST be avoided as they will collide with the container's top. \\
\addlinespace
partition surface for multi object placement &
When an instruction requires placing multiple objects onto a single surface like a stove or tray, the planner MUST partition the surface area first. For two objects, it should derive placements near the 1/3 and 2/3 marks along the x and y axes, not just slightly off-center. \\
\midrule
\multicolumn{2}{@{}l}{\emph{Task-specific strategies}} \\
\addlinespace
strategy: place mug into microwave and close door &
1.~\texttt{pick} the \texttt{yellow\_and\_white\_mug}.
2.~\texttt{place\_with\_gripper\_orientation\_drop} into the \texttt{microwave}, aligning the gripper's W-axis opposite to the microwave's opening axis (its U-axis).
Target a point on the microwave's U-W center plane, positioned 1/4 of the way deep from the opening.
Judge: verify the mug is fully inside, clear of the door's path.
3.~\texttt{close\_door} on the \texttt{microwave}. Judge: verify the door is fully shut. \\
\bottomrule
\end{tabular}
\caption{Example memory entries distilled from search trajectories: general rules (top) and task-specific strategies (bottom).}
\label{tab:memory_examples}
\end{table}

\section{VLM Prompts}

We list the verbatim prompt templates used throughout our pipeline.
Curly-brace tokens such as \texttt{\{instruction\}} are runtime placeholders, and \texttt{[IMG]} marks where rendered images are inserted into the multimodal prompt.

\subsection{Asset Generation and Annotation}
\label{vlm_prompt_asset}

We first prompt Gemini to propose object categories for each scenario, then annotate each generated asset with its visual description, function, size range, physical properties, canonical axis directions, and resting pose.

\begin{promptbox}[Asset category proposal]
You are an expert asset librarian for robotics simulation.
Scenario: {scenario}

Generate two lists of VALID asset categories for this scenario:
1) Manipulanda List: objects suitable for a gripper to pick or move.
2) Container List: objects that can hold or contain other objects.

Return exactly {manipulanda_count} manipulanda items and {container_count} container items.
Guidelines:
- Use generic category names (no brands).
- Use singular nouns or short noun phrases.
- Keep items realistic for the scenario.
- Avoid duplicates or near-duplicates across both lists.

- AVOID SOFT items that are hard to simulate (e.g., cloth, liquids).
- Assign a level (1, 2, or 3) to each category based on how common
  and how diverse (i.e. usually displays a wide range of shapes or appearance) the category is (more common/diverse => higher level).

Level meaning:
- Level 1: very common and very diverse category
- Level 2: common or moderately diverse category
- Level 3: niche or limited diversity category

Output JSON only with this schema:
{
  "manipulanda": [{"category": "item1", "level": 1}, ...],
  "containers": [{"category": "item1", "level": 2}, ...]
}
\end{promptbox}

\begin{promptbox}[Fine-grained visual description]
You are given an image of a single object.
Object category: {category}
You should use 1-2 words to describe each of the following three attributes of the object:
shape, color, texture (e.g. patterns/markings/branding info/logos/...).
Then return JSON only with this schema:
{"fine_grained_visual":["..."]}
["..."] is a list of concise noun phrases combining these attributes.
Make sure you include all combinations of these attributes in the list.

Example output list: ["thin soda can", "green soda can", "sprite can", "thin, green soda can", "green, sprite can", "thin, green, sprite can"]
\end{promptbox}

\begin{promptbox}[Size-range estimation]
You are given an image of a single object.
Object category: {category}
The function label of this object is: {functions}
Please estimate a valid size range (meters) for the longest axis with the following criteria:
 - Use common sense.
 - If manipulanda is true, make sure the asset can be picked up and held steadily with a Franka-Panda arm. (Assets CANNOT be TOO BIG.) But if the asset serves as static container/surface, this can be overlooked.
Return JSON only with this schema:
{"reasoning": "reasoning for the size estimate especially about graspability by the Panda gripper", "long_axis_min_m":0.05,"long_axis_max_m":0.15}
\end{promptbox}

\begin{promptbox}[Physical properties (mass and friction)]
You are given an image of a single object.
Object category: {category}
The estimated longest axis (m) of this object is: {long_axis_length}
Assign plausible mass (kg) and friction coefficient (0.8-1.2).
Return JSON only with this schema:
{"mass_kg":0.2,"friction":1.0}
\end{promptbox}

\begin{promptbox}[Canonical axis annotation]
You are an expert of spatial reasoning.
Your task is to describe the geometric meaning of the positive direction of the three primary axes (X, Y, Z) of the given object.

***** Input *****
1. Object name: {object_name}

2. Front view image: [IMG]
The front view is looking at the object from the front along -X. In the front view image,
+X axis points outwards to the front of this image.
+Y axis points to the right side of this image.
+Z axis points upwards in this image.

3. Top view image: [IMG]
The top view is looking at the object from the top along -Z. In the top view,
+X axis points downwards in this image.
+Y axis points to the right side of this image.
+Z axis points outwards to the front of this image.

4. Side view image: [IMG]
The side view is looking at the object from the side along +Y. In the side view,
+X axis points to the right side of this image.
+Y axis points inwards into the image.
+Z axis points upwards in this image.

***** Output *****
Provide the annotation in the following JSON format:
{
    "Reasoning": "<brief reasoning about how to determine the axis directions based on the object shape and features>",
    "X_axis": "<description of the X axis direction (from which part to which part)>",
    "Y_axis": "<description of the Y axis direction (from which part to which part)>",
    "Z_axis": "<description of the Z axis direction (from which part to which part)>"
}
\end{promptbox}

\begin{promptbox}[Upright resting-pose annotation]
What is the 'upright pose' of a {category} to stand UPRIGHT and STEADILY on the table?
The {category}'s axes in its canonical frame are annotated as follows:

- x axis: {x_axis}
- y axis: {y_axis}
- z axis: {z_axis}

Which axis should be aligned with the z+ axis of the world frame? (choose from +x, -x, +y, -y, +z, -z)

Provide the annotation in the following JSON format:
{
    "Reasoning": "<reason about how to determine the axis alignment>",
    "Alignment": "<axis, direction>(e.g. (x, +)/(z, -)/(y, -)/...)"
}
\end{promptbox}

\subsection{VLM Agent in MCTS}
\label{sec: vlm_planner_prompt}

At each node, the agent first selects the next subgoal and skill primitive, and then grounds the chosen skill into concrete parameters (target asset and 3D location) with a per-skill prompt.

\begin{promptbox}[Subgoal planning]
# Overall Instruction
You are an expert roboticist tasked with solving a long-horizon manipulation task: {instruction} by breaking it into subgoals.
This task involves multiple turns of sequential subgoal planning, choosing the next skill primitive to execute until completion.


# Input

## Scene Information
### Multi-view Visual Observations
You have access to visual observations of the scene from the following viewpoints:
{view_list}
Here is the current visual observation of the scene rendered from all viewpoints available:
{visual_observations}
### Object Layout
The scene is composed of multiple objects with the following properties:
{object_information}
### Contact Information
{contact_information}
### History Information
{history_information}
### Skill Primitives
You have access to the following skill primitives to manipulate the scene:
{skill_primitives}


# Output
Your output must include EXACTLY one Thinking Phase and one Answer Phase.

## Thinking Phase
### Scene state analysis
Describe the current state of the scene shown in different views with a concise paragraph. Pay attention to critical information in each view.
{reflection_instruction}
### Subgoal Planning
Choose the most promising next subgoal from the list of skill primitives provided above. Justify your choice in a concise paragraph.
### View Selection
Choose the single view that best supports the chosen subgoal and asset.
Prefer the view that
(1) makes the relevant object or part easiest to identify
(2) most clearly exposes the contact surface and keypoints(e.g. container bottom for place, handles for articulated object manipulation, ...).
You MUST pick the view from the list provided above only. Output ONLY the integer index from the view list.

Please enclose the reasoning for this phase in a set of <thinking></thinking> tags.
An example thinking output:
<thinking>
reasoning content
</thinking>

## Answer Phase
Based on your Thinking Phase analysis, output the relevant information for the next subgoal:
- Subgoal: use one sentence to describe the next subgoal.
- Skill: the integer index of skill primitive chosen from the given list.
- View: the integer index of the chosen view from the provided list.
An example answer output:
<answer>
Subgoal: Pick up the mug from the table.
Skill: 1
View: 2
</answer>
\end{promptbox}

The skill-grounding prompts share a common structure; we show the grasp and orientation-aware place skill prompts as examples.

\begin{promptbox}[Grasp grounding]
{coordinate_system}You are an assistant tasked with predicting critical parameters for one step(a subgoal) in a long-horizon robotic manipulation task.
The overall task is '{instruction}', and the current subgoal is '{subgoal}'.

Your task is to determine the target asset to grasp in the scene based on the current state of the scene.
You MUST pick the target asset from the provided object list only. Output ONLY the integer index from the list.

Asset list:
{object_information}

Please provide your answer in the following format:
<thinking>
reasoning content.
</thinking>
<answer>
Asset: the id of the object to grasp (integer index from the object list).
</answer>

Example answer:
<thinking>
The target asset is the red apple on the table as the subgoal requires picking up the apple.
</thinking>
<answer>
Asset: 3
</answer>
\end{promptbox}

\begin{promptbox}[Orientation-aware Place grounding]
{coordinate_system}You are an assistant tasked with predicting critical parameters for one step(a subgoal) in a long-horizon robotic manipulation task.
The overall task is '{instruction}', and the current subgoal is '{subgoal}'.

Your task is to:
1. Determine the target object serving as place destination.
2. Determine the goal position in 3D space for the center of the held object, which is usually in a local region relative to the target object.
3. Align the held object orientation with respect to the target object.

For asset identification, you MUST pick the asset from the provided object list only. Output ONLY the integer index from the list.

For orientation alignment, we select one primary axis (x, y, or z) in the canonical object frame for both the held asset and the target asset, and align the selected axes in the same direction.
Thus, you need to specify which axis to use for both objects and the alignment direction (same, represented with '+', or opposite, represented with '-').
e.g. (x, z, +) means aligning the x axis of the held object with the z axis of the target object in the same direction;
(x, y, -) means aligning the x axis of the held object with the y axis of the target object in the opposite direction at the END STATE of this task.
The meaning of the axes in the canonical object frame is described in the asset list below.

Held Asset:
{contact_information}

Asset list:
{object_information}

Please provide your answer in the following format:
<thinking>
## Reasoning about asset identification
## Reasoning about point grounding
## Carefully reasoning about orientation alignment
</thinking>
<answer>
Asset: the id of the object as place destination (integer index from the object list).
Point: (x, y, z) in meters in the world frame.
Alignment: (Held_Axis, Target_Axis, Direction)
</answer>

Example answer:
<thinking>
The target asset is the mug as the subgoal requires putting the carrot into the mug.
To ensure the carrot fits into the mug, I will align the z axis of the carrot with the z axis of the mug in the same direction.
</thinking>
<answer>
Asset: 4
Point: (0.5, 0.5, 0.1)
Alignment: (z, z, +)
</answer>
\end{promptbox}

\subsection{VLM Verifier in MCTS}
\label{sec: vlm_verifier_prompt}

The verifier operates at two granularities: it checks whether the current subgoal was completed (and whether the action was safe), and it scores overall task progress, whose reflection text is reused as a planning hint. Safety-free variants of both prompts are also available.

\begin{promptbox}[Subgoal completion and safety check]
You are an expert roboticist. The robot is performing the task of '{instruction}', and the current subgoal is {subgoal}.
Your task is to determine whether the subgoal is completed safely by analyzing the provided visual observations, object layout, and contact information.

{coordinate_system}***** Input *****
This is the initial state of the scene.
(1) Visual Observations:
{init_visual_observations}
(2) Object Layout:
{init_object_information}
(3) Contact Information:
{init_contact_information}
From the initial state, the robot has made some progress, resulting in the following current state of the scene.
(1) Visual Observations:
{current_visual_observations}
(2) Object Layout:
{current_object_information}
(3) Contact Information:
{current_contact_information}

***** Output *****
## Thinking
1. Propose the criteria for subgoal completion and safety based on the subgoal.
2. Analyze the Initial State:
(1) Examine the initial visual observations
(2) Examine the initial object layout
(3) Examine the initial gripper-object contact information
to understand the starting conditions of the scene.
3. Analyze the Current State:
(1) Examine the current visual observations
(2) Examine the current object layout
(3) Examine the current gripper-object contact information
to assess the changes that have occurred since the initial state.
4. Compare States: Identify the differences between the initial and current states by careful comparison, focusing on changes relevant to the subgoal. If the relevant object is missing from view(e.g. occluded by other objects or some container), you should infer its position based on other information.
5. Apply Criteria: Please match the scene transitions with the proposed criteria above one by one before making decisions.
6. Conclusion - Completion: Decide whether this subgoal is completed.
7. Conclusion - Safety: Decide whether the robot's action is safe(i.e. no other objects being knocked over or SEVERELY damaged).

## Summary
Summarize the image content with one sentence.

## Subgoal Completion and Safety
Whether the subgoal is completed and the robot's action is safe.

Make sure you use '## Thinking', '## Summary', and '## Subgoal Completion and Safety' as section headers in your response.
\end{promptbox}

\begin{promptbox}[Overall task-progress scoring]
You are an expert roboticist. The robot is performing the task of '{instruction}'.
Your task is to evaluate the robot's progress towards completing the overall task by analyzing the provided initial/current scene states and robot action trajectory.

{coordinate_system}***** Input *****
This is the initial state of the scene.
(1) Visual Observations:
{init_visual_observations}
(2) Object Layout:
{init_object_information}
(3) Contact Information:
{init_contact_information}

From the initial state, the robot has made progress represented by the following trajectory:
{history_information}

Such progress results in the current state of the scene.
(1) Visual Observations:
{current_visual_observations}
(2) Object Layout:
{current_object_information}
(3) Contact Information:
{current_contact_information}

***** Output *****
Based on the above information, please do the following:
## Reflection
Evaluate the overall progress of the robot toward completing the task by analyzing the trajectory and comparing initial and current scene state.
Are the subgoals reasonable? Is the robot making meaningful progress towards the final goal?

## Progress Score
Predict a task completion percentage between 0 and 100.
    - In the initial state, the task completion percentage is 0.
    - If all necessary steps are completed, the task completion percentage is 100.
    - The score should increase when a **meaningful** subgoal is completed.
    and should decrease if an action **undoes or degrades** a previously completed meaningful subgoal.
    - The score should also decrease if the robot's action causes SEVERE damage to other objects in the environment(Minor disturbance is acceptable and can be ignored).
    - **OVERRIDE**: If the overall task is fully completed at the current state, output 100 even if some unsafe actions occurred along the trajectory. Safety penalties only reduce the score when the task is INCOMPLETE.
    In this part, just provide one score and one concise explaining sentence.

Make sure you use '## Reflection' and '## Progress Score' as section headers in your response.
Please use EXACTLY ONE number between 0 and 100 for the Progress Score.
\end{promptbox}

\subsection{Memory Module}

\label{sec:memory_prompt}

We use two prompts to distill general rules and task-specific strategies from search trajectories, and a third prompt to select memory keys at planning time.

\begin{promptbox}[General rule distillation]
You are extracting reusable robot-manipulation rules from a single MCTS-style
search trajectory. The trajectory shows a robot trying to satisfy a
natural-language instruction. You are given the initial object layout (3D
centers and axis-aligned bounding boxes, in meters), the robot's available
skill primitives and their descriptions, and for each step:

  - the subgoal the actor proposed
  - the actor's reasoning (phase 1: subgoal planning; phase 2: action grounding)
  - the action parameters that were executed
  - the resulting scene image

You are NOT shown any verification verdicts -- only whether the trajectory as a
whole ended in SUCCESS or FAILURE. Treat the actor's reasoning as claims, not
facts: a confident justification may itself be the mistake that caused the
final result. Using the AVAILABLE SKILL PRIMITIVES descriptions, re-examine
for each step whether the chosen skill was the right one.

IF THE TRAJECTORY RESULT IS FAILURE, follow this protocol before writing rules:
  1. Determine what the failure was from the step images and the instruction,
     then trace backwards to the EARLIEST decision that made it inevitable.
     That is often NOT the step where the failure became visible, but an
     earlier step whose downstream consequence was never checked.
  2. Check the actor's claims against the INITIAL OBJECT LAYOUT geometry.
     A claim that contradicts these numbers is the prime root-cause suspect.
  3. The rule you write MUST PREVENT that root-cause decision. NEVER restate a
     decision from this failed trajectory as a prescription -- if a choice was
     followed by downstream failure, prescribe the alternative instead.

Your job: extract 1-2 highly SPECIFIC, GENERALIZABLE rules that capture the
NON-OBVIOUS insight a naive policy would have missed. Quality over quantity.

Each rule should compress into 2-3 short sentences (~40-60 words) and cover:

  - TRIGGER CONDITION ("When the destination container is lying on its side...",
    "When placing a thin elongated object into a narrow-mouth container...").
    Use concrete object categories (e.g. "cup, mug, vase, bottle") not abstract
    nouns ("container").
  - PRESCRIPTIVE ORDERING with MUST / SHOULD / BEFORE / NOT. Name WHO acts
    (actor's phase-1 planner, actor's phase-2 grounder, or judge).
  - VERIFICATION CRITERION the judge can actually check from observations:
    axis alignment, bounding-box containment, pose flag (upright / lying),
    contact, position delta. Not "looks correct".
  - Optionally one short clause for an exclusion ("propped on rim doesn't
    count") OR a fallback ("drop directly from above if the gripper can't
    fit") -- pick at most one of the two, only if it's load-bearing.

OUTPUT FORMAT -- return ONLY a single JSON object. Each top-level key is the
rule's short title (3-6 lowercase words separated by spaces, no quotes).
Each value is:

  {
    "value":  "<2-3 sentences, ~40-60 words. Terse, prescriptive prose.>",
    "tags":   [<subset of "actor_phase_1", "actor_phase_2", "judge">]
  }

Tag meanings:
  - actor_phase_1: subgoal planning (deciding WHAT to do next and WHICH
                   skill primitive to use)
  - actor_phase_2: action grounding (deciding HOW: which asset/point, which
                   orientation alignment)
  - judge:         verifying subgoal completion and safety

STYLE EXAMPLE (illustrative only -- invent a DIFFERENT rule on a DIFFERENT
topic for the actual trajectory below; do not copy wording or topic).
Note the length and density -- aim for this:

{
  "stow knife into block slot": {
    "value": "When the destination is a slotted knife block, the actor MUST first align the held tool's long axis with the slot's long axis BEFORE the place action, targeting the slot center (not the block's top face). The judge MUST verify both that the tool's long axis is within ~10 deg of the slot's axis and that its lower bounding-box z is below the slot opening's z. A tool resting on the block's top face does NOT count.",
    "tags": ["actor_phase_2", "judge"]
  }
}

RULES FOR YOUR OUTPUT:
  - Output 1-2 rules. NEVER more than 2.
  - Stay within ~40-60 words per rule. If you go over 70 words, you have
    too many clauses -- cut the weakest one.
  - Each rule MUST name a specific trigger condition, not a generic principle.
  - Each rule MUST include a verification criterion expressed in concrete
    geometric / pose terms (axis alignment, bounding-box containment, pose
    flag, contact).
  - Do NOT mention specific asset names from this scene (no instance ids like
    'spoon_0057' or 'mug_0073'). Use category names ("cup, mug, vase").
  - Do NOT produce generic safety rules ("avoid disturbing nearby objects",
    "don't collide"). The safety judge already handles these.
  - Do NOT produce trivia rules ("use the appropriate view", "plan multiple
    subgoals", "the robot has a gripper").
  - Failure mode -> describe what should have been done differently (often a
    BEFORE-ordering or a judge-rejection rule).
  - Success -> describe the non-obvious invariant that made it work and that
    a naive policy would have skipped.
  - No markdown, no commentary, no code fences. Return JUST the JSON object.
\end{promptbox}

\begin{promptbox}[Task-specific strategy distillation]
You are distilling ONE concrete, task-specific execution strategy from a single
MCTS-style search trajectory of a robot following a natural-language instruction.
You are given the initial object layout (3D centers and axis-aligned bounding
boxes, in meters), the robot's available skill primitives and their descriptions,
and for each step: the subgoal, the actor's reasoning (phase 1: subgoal planning
incl. skill choice; phase 2: action grounding), the executed action parameters,
and the resulting scene image.

You are NOT shown any verification verdicts -- only whether the trajectory as a
whole ended in SUCCESS or FAILURE. Treat the actor's reasoning as claims, not
facts. Using the AVAILABLE SKILL PRIMITIVES descriptions, re-examine for each
step whether the chosen skill and parameters were right. Steps may list "other
branches attempted from the same parent state": entries with invalid=True and a
term_reason are parameter choices that FAILED -- your recipe MUST NOT reuse
those parameter values; contrast them with what worked to bracket the feasible
range. If the trajectory FAILED, your strategy must PRESCRIBE what would have
succeeded (fix the root causes evidenced in the trajectory), not restate what
was tried.

Output ONE strategy entry for THIS task (not a general rule):
  - Unlike general rules, you SHOULD name the concrete objects and containers of
    this task (e.g. "the mug", "the microwave") and the exact skill primitive
    names from the AVAILABLE SKILL PRIMITIVES list.
  - Give a step-by-step recipe covering the full task: which skill at each step,
    the alignment expressed in the TARGET OBJECT'S OWN frame axes, and the
    target point for each placement.
  - HARD CONSTRAINT: NO absolute world coordinates. Scene layout is randomized
    between episodes, so every position MUST be a formula relative to quantities
    readable at run time: AABB faces/centers, the cavity opening direction, the
    held object's own dimensions, clearance margins in cm.
  - State what the judge should verify at the critical steps.

OUTPUT FORMAT -- return ONLY a single JSON object with exactly ONE top-level key:
the task signature as a short lowercase phrase starting with "strategy:" (e.g.
"strategy: place mug into microwave and close door"). Its value is:

  {
    "value":  "<the recipe, <=120 words, terse imperative prose, numbered steps>",
    "tags":   ["actor_phase_1", "actor_phase_2", "judge"]
  }

No markdown, no commentary, no code fences. Return JUST the JSON object.
\end{promptbox}

\begin{promptbox}[Memory key selection at planning time]
You are a memory key selector for a robot manipulation policy.

Given the current task instruction and subgoal, select the most relevant memory keys from the provided key list.

Instruction:
{instruction}

Subgoal:
{subgoal}

Object poses (only whitelisted objects show a pose):
{pose_info}

Available memory keys:
{keys}

Rules:
- Select the {k} most relevant keys from the available memory keys to the current instruction/subgoal.
- Do not invent a new key.
- Output one key per line, no extra text, no numbering, no quotes.
\end{promptbox}

\raggedbottom
\section{Qualitative Results}

\subsection{Qualitative Results for Collected Trajectories}
\label{app:qualitative_traj}

We visualize representative trajectories collected by our pipeline for each task type.
Each strip shows six frames uniformly sampled from a single trajectory, including its first and last frames, with the corresponding language instruction below it.
Figs.~\ref{fig:traj_pick}--\ref{fig:traj_long_horizon} show in-the-wild tasks,
while Figs.~\ref{fig:traj_libero} and~\ref{fig:traj_simpler} show tasks derived from LIBERO~\cite{liu2023libero} and SIMPLER-WidowX~\cite{SIMPLER}.

\noindent\parbox{\linewidth}{\centering
    \includegraphics[width=\linewidth]{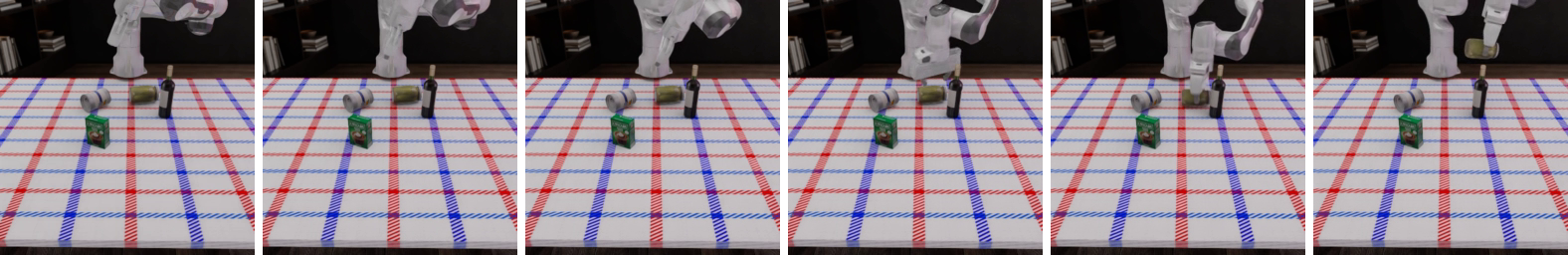}\\[1pt]
    {\footnotesize ``Grasp the pickle jar.''}
}\par\vspace{6pt}
\noindent\parbox{\linewidth}{\centering
    \includegraphics[width=\linewidth]{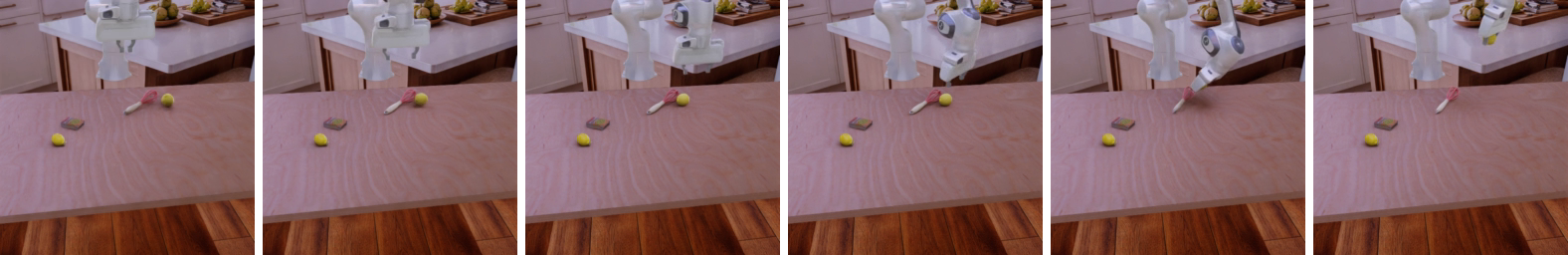}\\[1pt]
    {\footnotesize ``Take hold of the lemon.''}
}\par\vspace{6pt}
\noindent\parbox{\linewidth}{\centering
    \includegraphics[width=\linewidth]{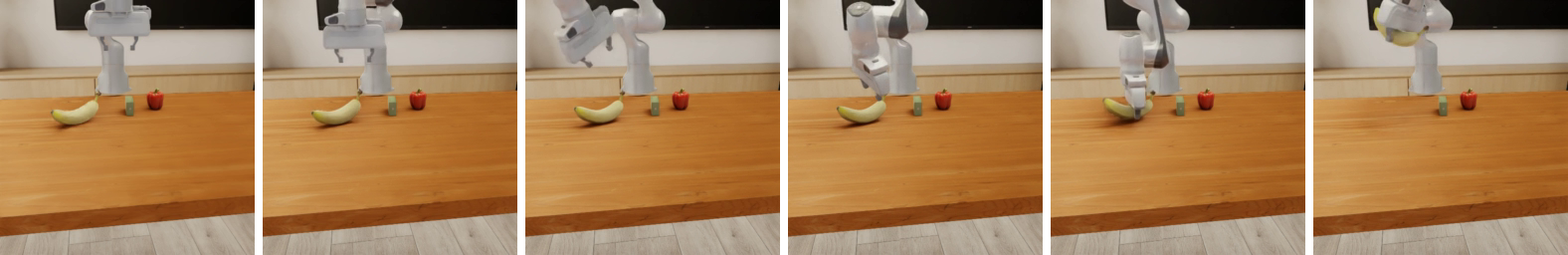}\\[1pt]
    {\footnotesize ``Pick up the banana.''}
}\par\vspace{6pt}
\noindent\parbox{\linewidth}{\centering
    \includegraphics[width=\linewidth]{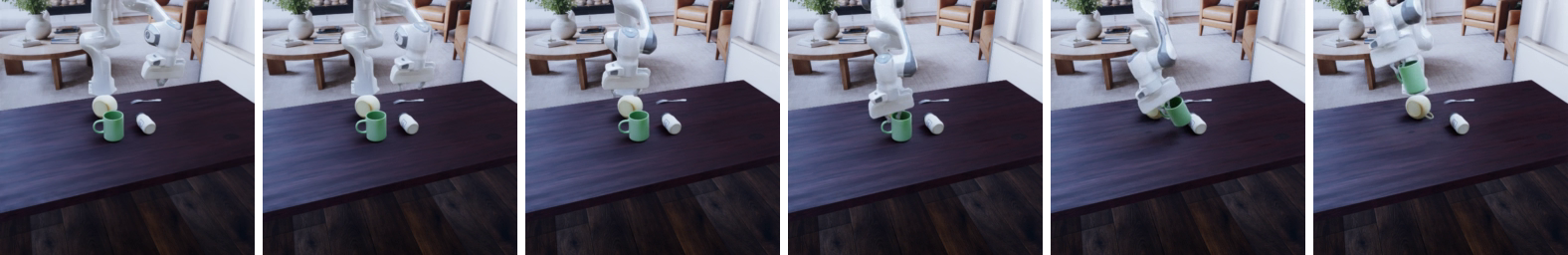}\\[1pt]
    {\footnotesize ``Lift the green cup.''}
}\par\vspace{6pt}
\nopagebreak
\captionof{figure}{Collected trajectories for \textbf{Pick} tasks.}\label{fig:traj_pick}\par
\vspace{10pt}

\noindent\parbox{\linewidth}{\centering
    \includegraphics[width=\linewidth]{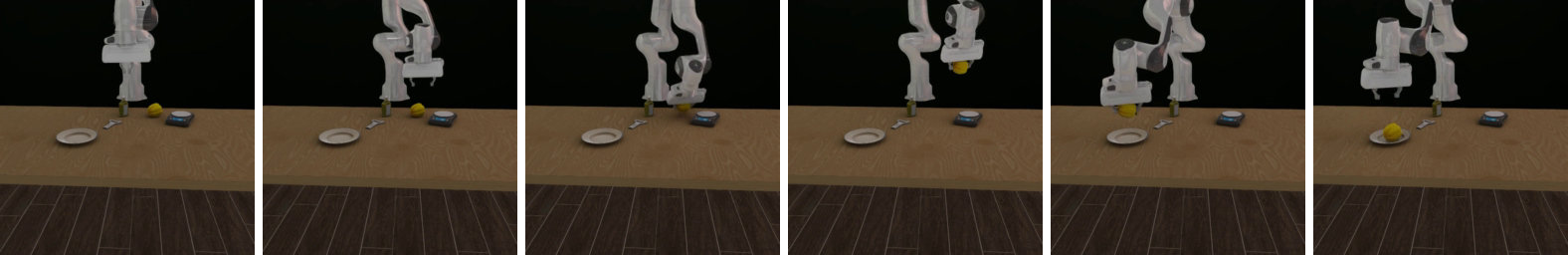}\\[1pt]
    {\footnotesize ``Put the bell pepper in the plate.''}
}\par\vspace{6pt}
\noindent\parbox{\linewidth}{\centering
    \includegraphics[width=\linewidth]{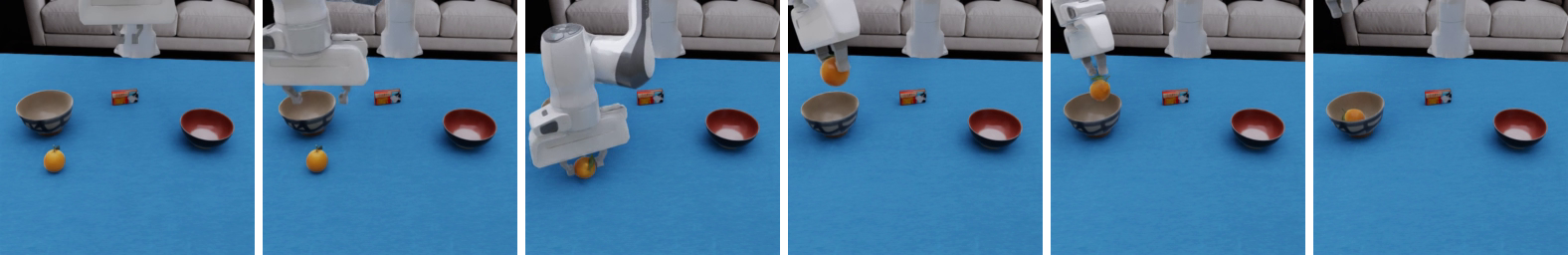}\\[1pt]
    {\footnotesize ``Place the orange into the bowl.''}
}\par\vspace{6pt}
\noindent\parbox{\linewidth}{\centering
    \includegraphics[width=\linewidth]{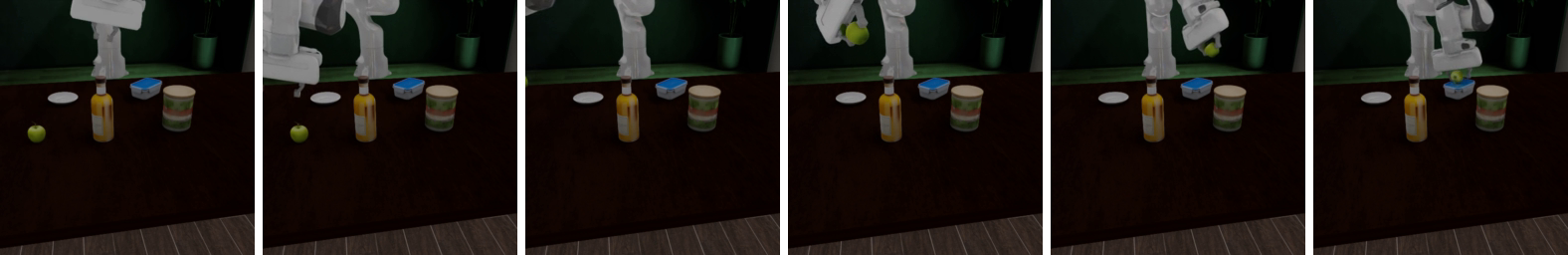}\\[1pt]
    {\footnotesize ``Deposit the apple in the food storage container.''}
}\par\vspace{6pt}
\noindent\parbox{\linewidth}{\centering
    \includegraphics[width=\linewidth]{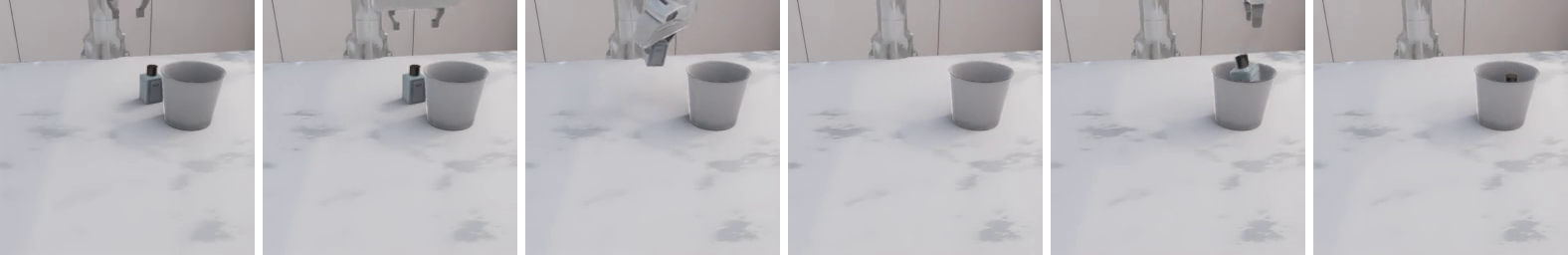}\\[1pt]
    {\footnotesize ``Settle the aftershave bottle in the bucket.''}
}\par\vspace{6pt}
\nopagebreak
\captionof{figure}{Collected trajectories for \textbf{Place} tasks.}\label{fig:traj_place}\par
\vspace{10pt}

\noindent\parbox{\linewidth}{\centering
    \includegraphics[width=\linewidth]{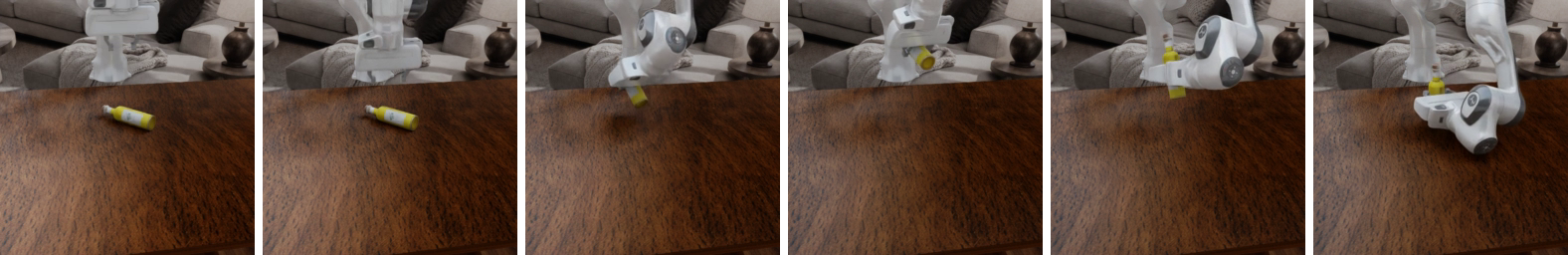}\\[1pt]
    {\footnotesize ``Straighten up the lying oil bottle.''}
}\par\vspace{6pt}
\noindent\parbox{\linewidth}{\centering
    \includegraphics[width=\linewidth]{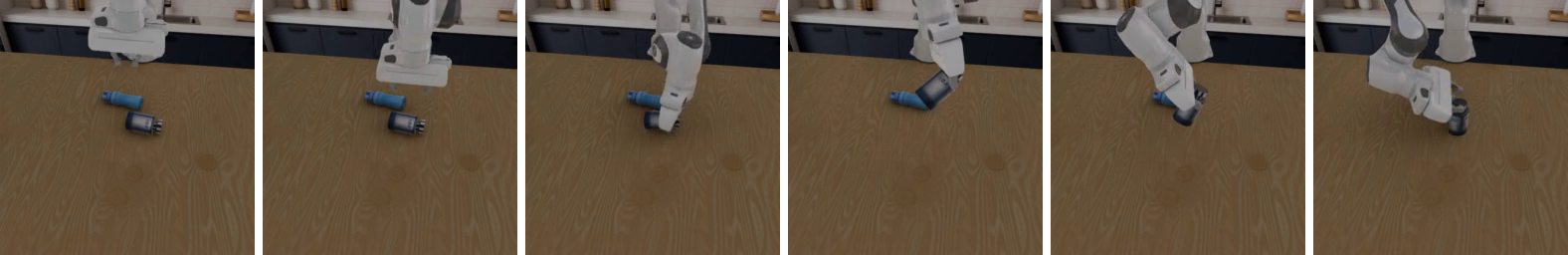}\\[1pt]
    {\footnotesize ``Upright the lying perfume bottle.''}
}\par\vspace{6pt}
\noindent\parbox{\linewidth}{\centering
    \includegraphics[width=\linewidth]{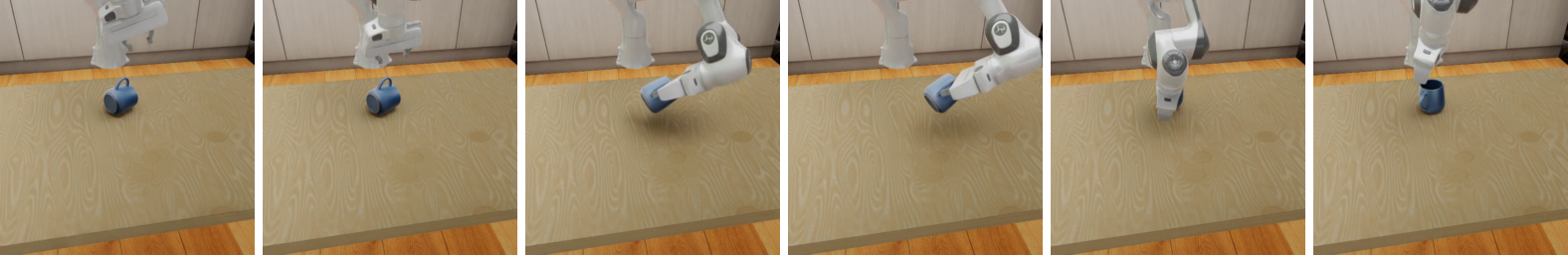}\\[1pt]
    {\footnotesize ``Make the lying mug stand straight up.''}
}\par\vspace{6pt}
\noindent\parbox{\linewidth}{\centering
    \includegraphics[width=\linewidth]{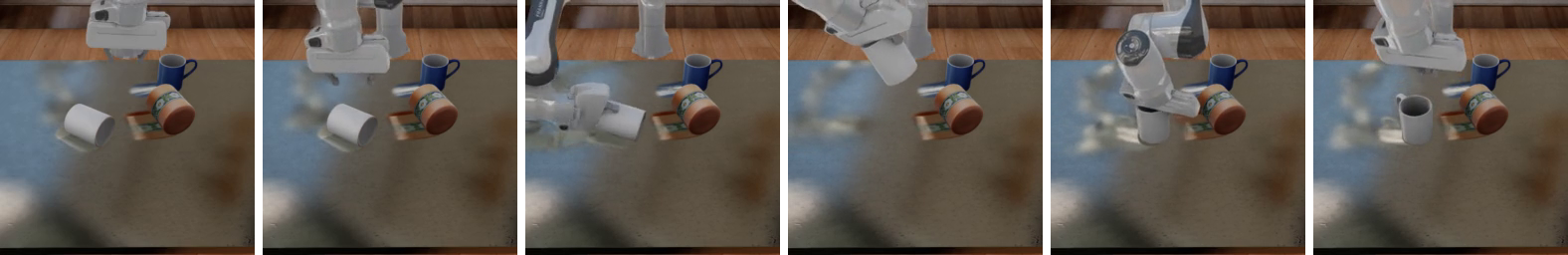}\\[1pt]
    {\footnotesize ``Set the cup upright.''}
}\par\vspace{6pt}
\nopagebreak
\captionof{figure}{Collected trajectories for \textbf{Straighten-up} tasks.}\label{fig:traj_straighten}\par
\vspace{10pt}

\noindent\parbox{\linewidth}{\centering
    \includegraphics[width=\linewidth]{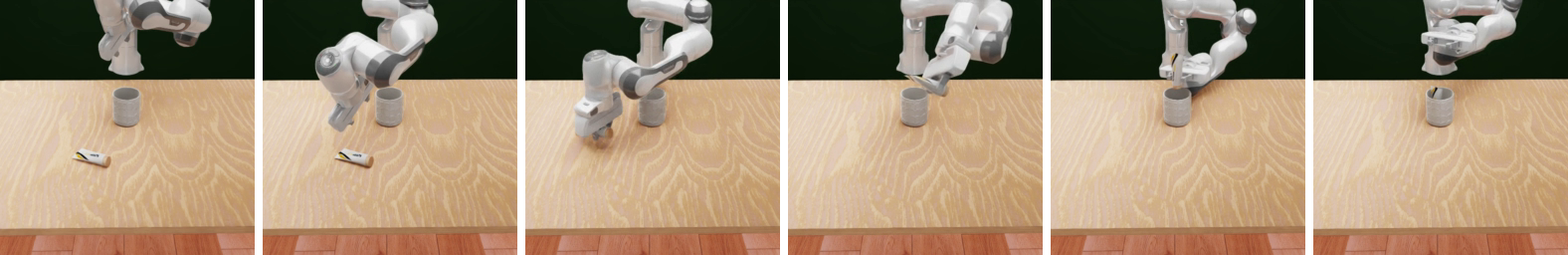}\\[1pt]
    {\footnotesize ``Place the toothpaste tube into the utensil holder.''}
}\par\vspace{6pt}
\noindent\parbox{\linewidth}{\centering
    \includegraphics[width=\linewidth]{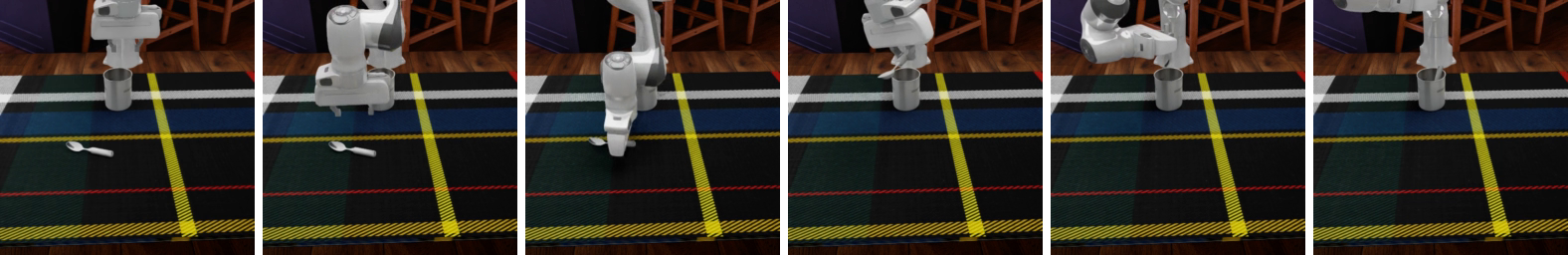}\\[1pt]
    {\footnotesize ``Rest the spoon inside the utensil holder.''}
}\par\vspace{6pt}
\noindent\parbox{\linewidth}{\centering
    \includegraphics[width=\linewidth]{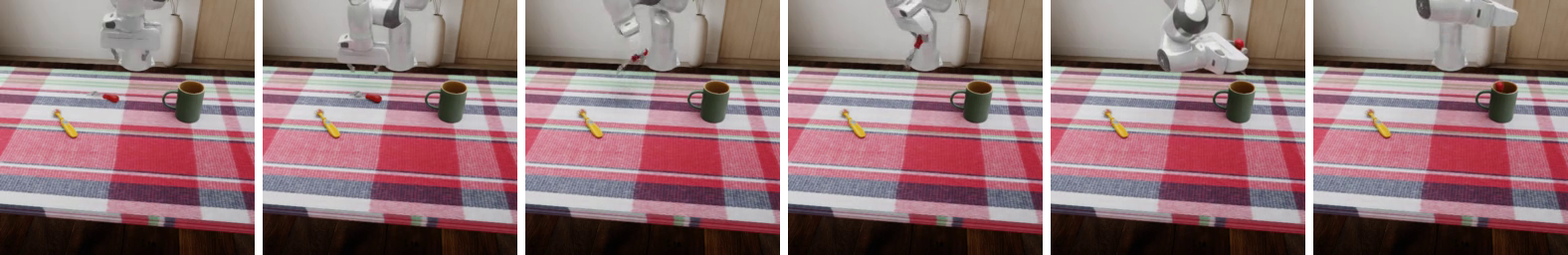}\\[1pt]
    {\footnotesize ``Deposit the fork into the cup.''}
}\par\vspace{6pt}
\noindent\parbox{\linewidth}{\centering
    \includegraphics[width=\linewidth]{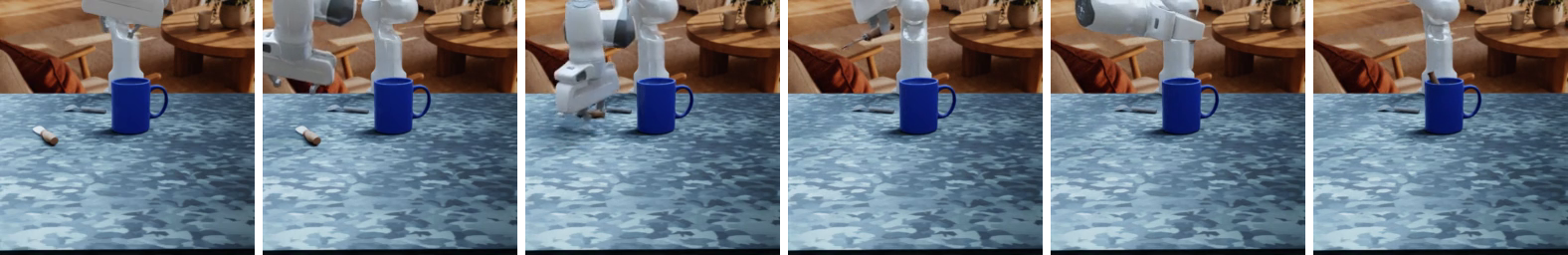}\\[1pt]
    {\footnotesize ``Pick up the butter knife and place it into the mug.''}
}\par\vspace{6pt}
\nopagebreak
\captionof{figure}{Collected trajectories for \textbf{Insert} tasks.}\label{fig:traj_insert}\par
\vspace{10pt}

\noindent\parbox{\linewidth}{\centering
    \includegraphics[width=\linewidth]{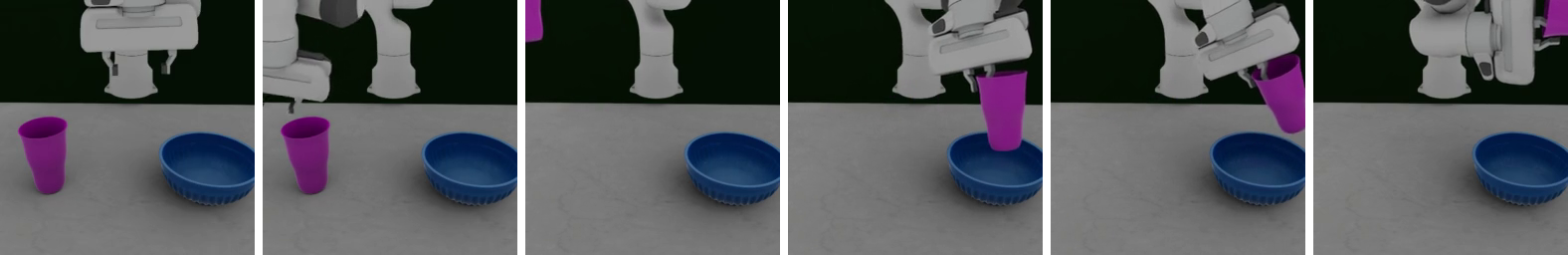}\\[1pt]
    {\footnotesize ``Pour from the cup into the bowl.''}
}\par\vspace{6pt}
\noindent\parbox{\linewidth}{\centering
    \includegraphics[width=\linewidth]{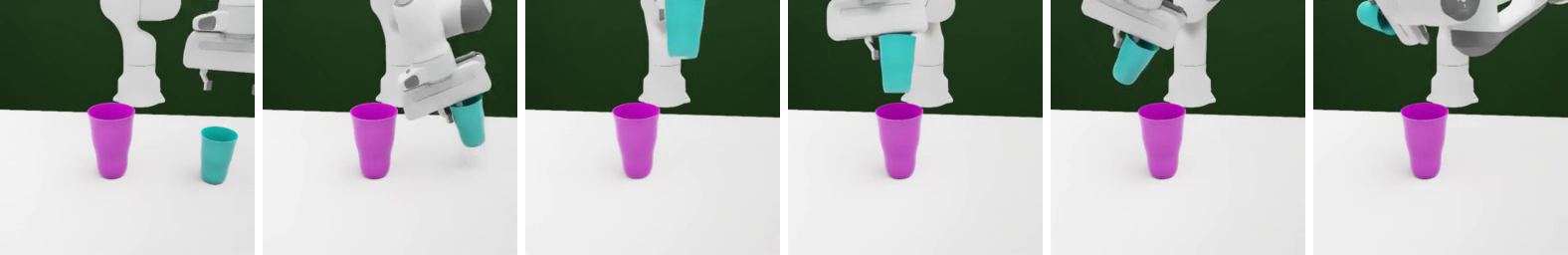}\\[1pt]
    {\footnotesize ``Pour from the green cup into the pink cup.''}
}\par\vspace{6pt}
\nopagebreak
\captionof{figure}{Collected trajectories for \textbf{Pour} tasks.}\label{fig:traj_pour}\par
\vspace{10pt}

\noindent\parbox{\linewidth}{\centering
    \includegraphics[width=\linewidth]{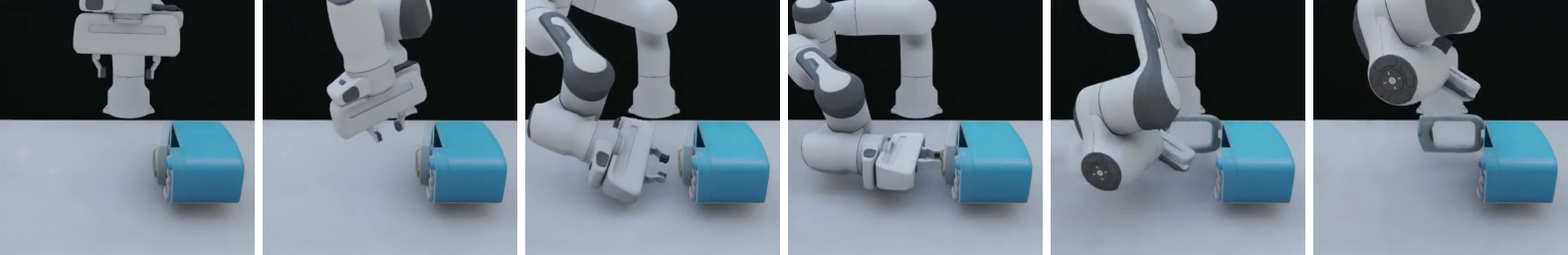}\\[1pt]
    {\footnotesize ``Open the microwave door.''}
}\par\vspace{6pt}
\noindent\parbox{\linewidth}{\centering
    \includegraphics[width=\linewidth]{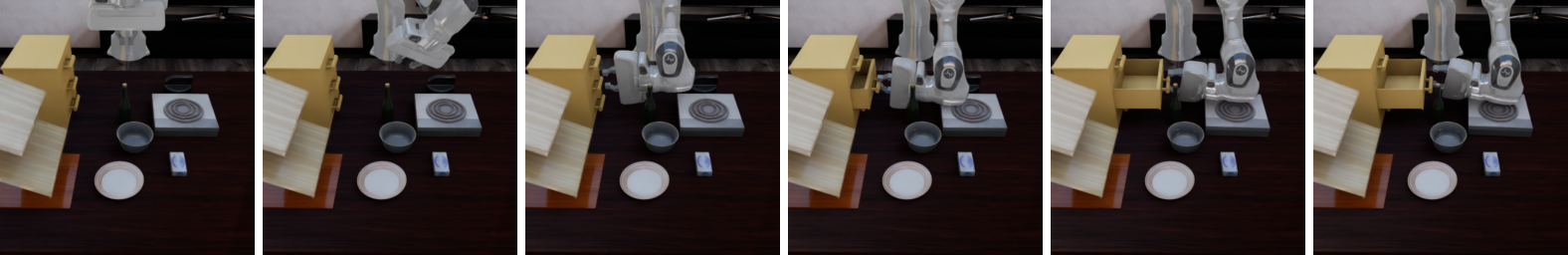}\\[1pt]
    {\footnotesize ``Open the middle drawer of the cabinet.''}
}\par\vspace{6pt}
\nopagebreak
\captionof{figure}{Collected trajectories for \textbf{Open} tasks.}\label{fig:traj_open}\par
\vspace{10pt}

\noindent\parbox{\linewidth}{\centering
    \includegraphics[width=\linewidth]{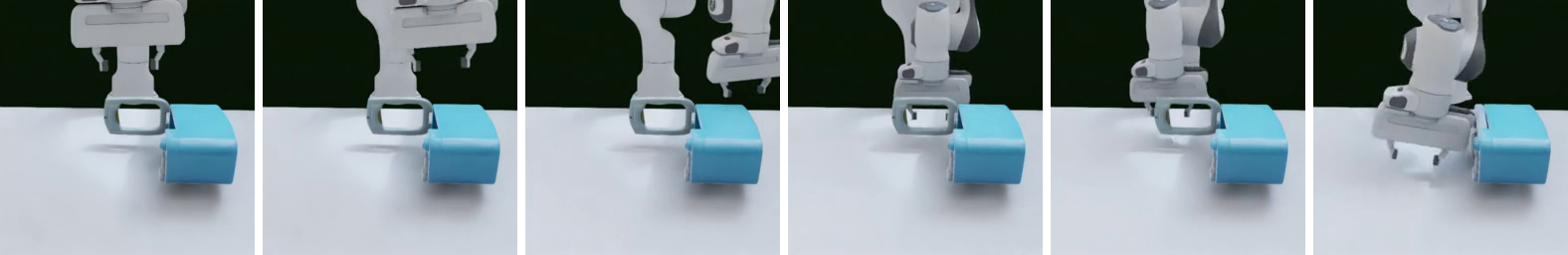}\\[1pt]
    {\footnotesize ``Close the microwave door.''}
}\par\vspace{6pt}
\noindent\parbox{\linewidth}{\centering
    \includegraphics[width=\linewidth]{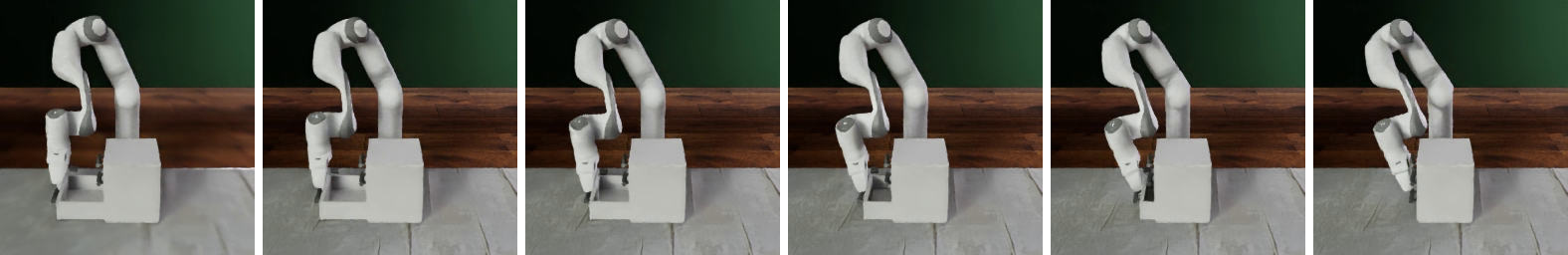}\\[1pt]
    {\footnotesize ``Close the bottom drawer.''}
}\par\vspace{6pt}
\nopagebreak
\captionof{figure}{Collected trajectories for \textbf{Close} tasks.}\label{fig:traj_close}\par
\vspace{10pt}

\noindent\parbox{\linewidth}{\centering
    \includegraphics[width=\linewidth]{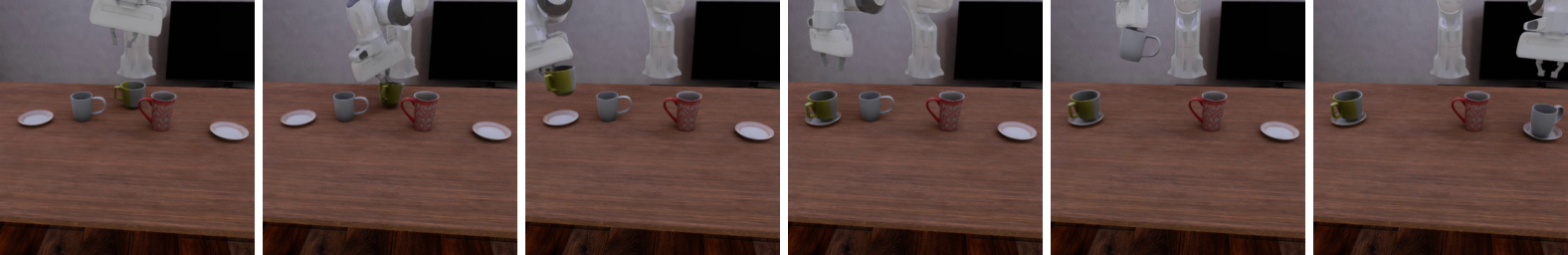}\\[1pt]
    {\footnotesize ``put the yellow-white mug on the left plate and the white mug on the right plate''}
}\par\vspace{6pt}
\noindent\parbox{\linewidth}{\centering
    \includegraphics[width=\linewidth]{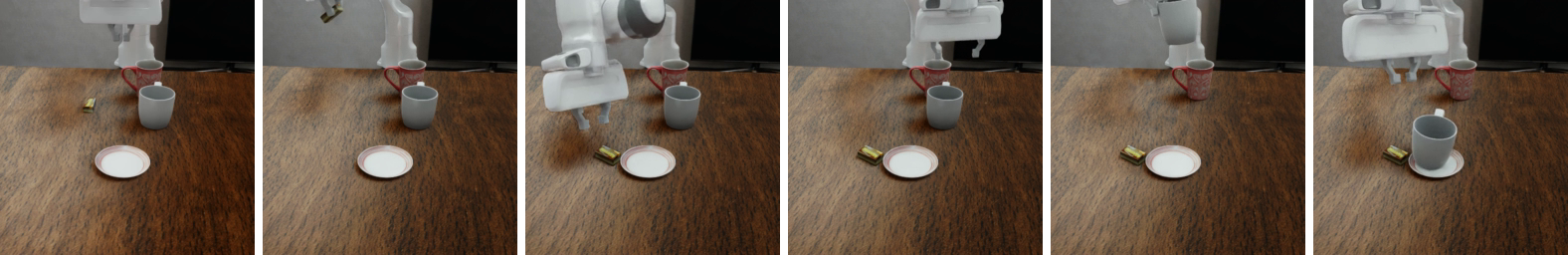}\\[1pt]
    {\footnotesize ``put the chocolate to the left of the plate and white mug on the plate''}
}\par\vspace{6pt}
\noindent\parbox{\linewidth}{\centering
    \includegraphics[width=\linewidth]{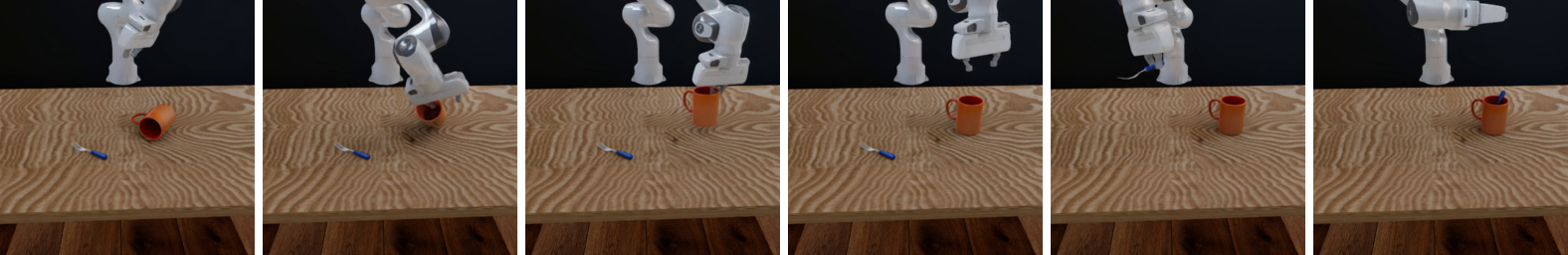}\\[1pt]
    {\footnotesize ``put the fork inside the mug''}
}\par\vspace{6pt}
\noindent\parbox{\linewidth}{\centering
    \includegraphics[width=\linewidth]{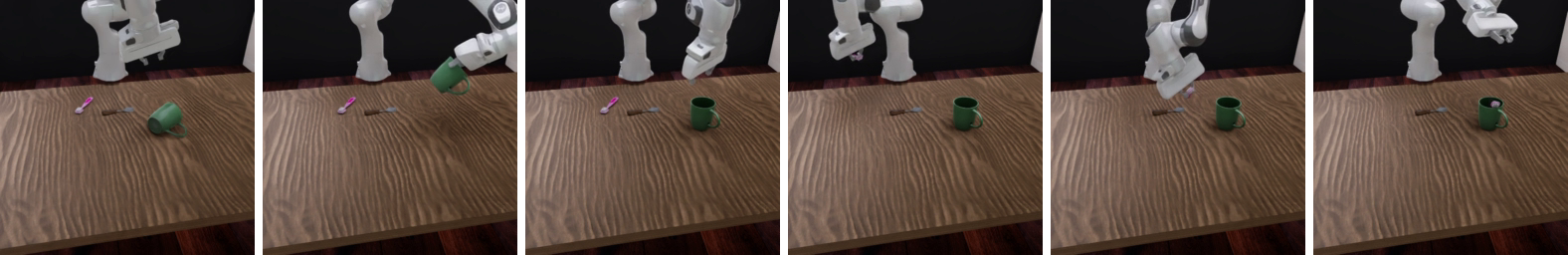}\\[1pt]
    {\footnotesize ``put the toothbrush in the green cup''}
}\par\vspace{6pt}
\nopagebreak
\captionof{figure}{Collected trajectories for \textbf{Long-horizon} tasks.}\label{fig:traj_long_horizon}\par
\vspace{10pt}

\noindent\parbox{\linewidth}{\centering
    \includegraphics[width=\linewidth]{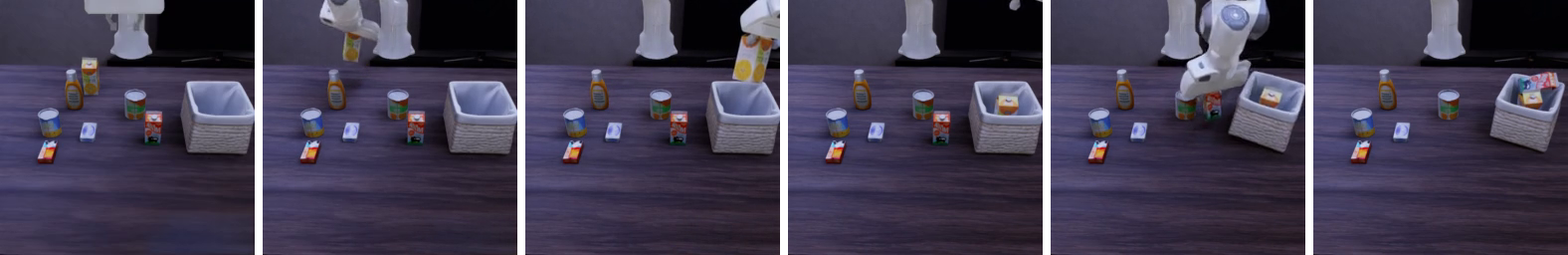}\\[1pt]
    {\footnotesize ``put both the milk and the orange juice in the basket''}
}\par\vspace{6pt}
\noindent\parbox{\linewidth}{\centering
    \includegraphics[width=\linewidth]{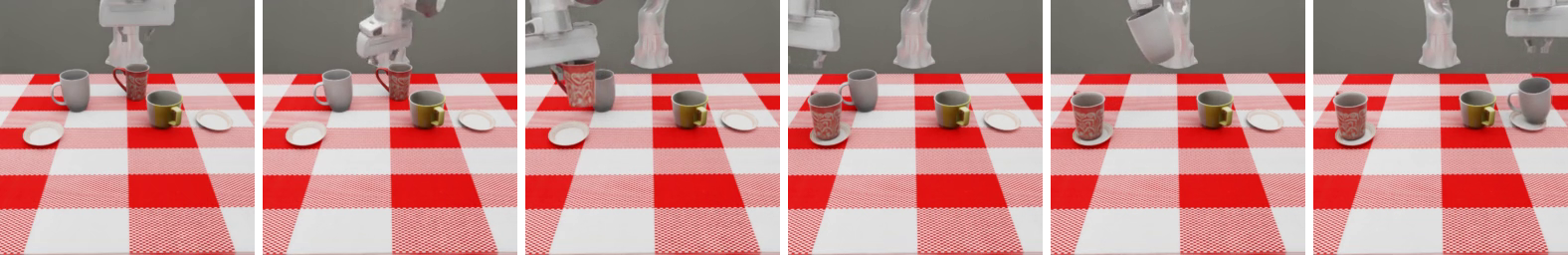}\\[1pt]
    {\footnotesize ``put the red coffee mug on the left plate and put the white mug on the right plate''}
}\par\vspace{6pt}
\noindent\parbox{\linewidth}{\centering
    \includegraphics[width=\linewidth]{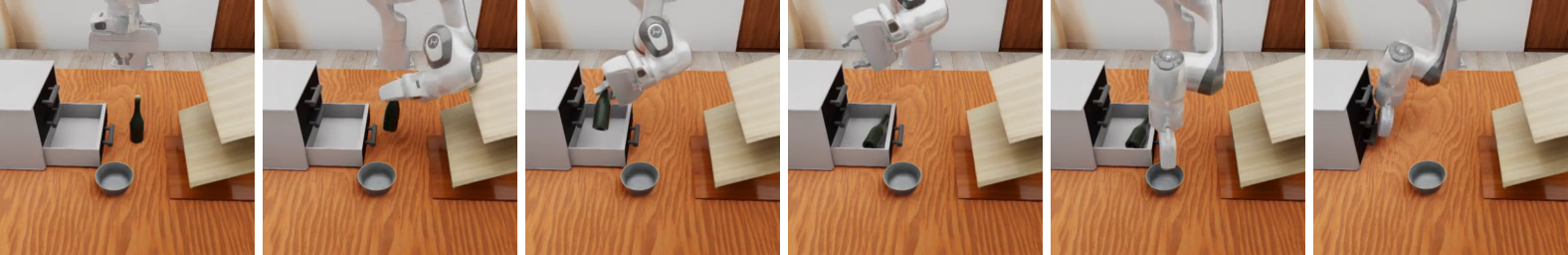}\\[1pt]
    {\footnotesize ``insert the bottle into the lower drawer and close it''}
}\par\vspace{6pt}
\noindent\parbox{\linewidth}{\centering
    \includegraphics[width=\linewidth]{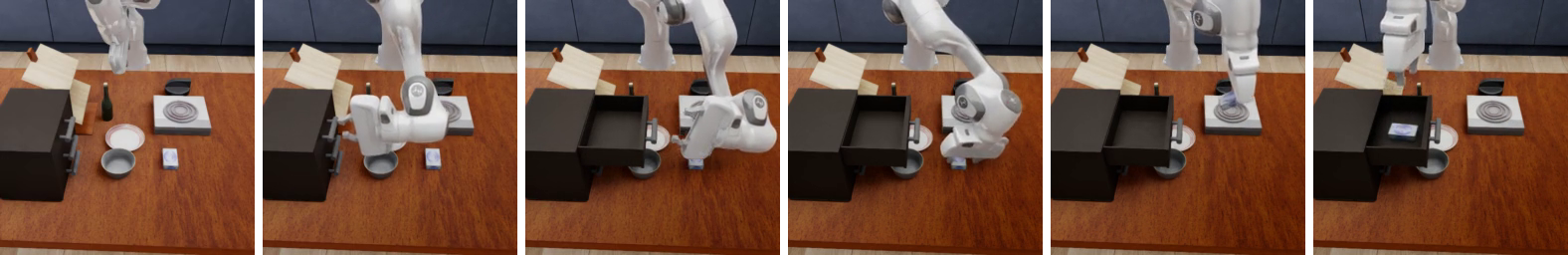}\\[1pt]
    {\footnotesize ``open the top drawer and place the cream cheese inside''}
}\par\vspace{6pt}
\nopagebreak
\captionof{figure}{Collected trajectories for \textbf{LIBERO-derived} tasks.}\label{fig:traj_libero}\par
\vspace{10pt}

\noindent\parbox{\linewidth}{\centering
    \includegraphics[width=\linewidth]{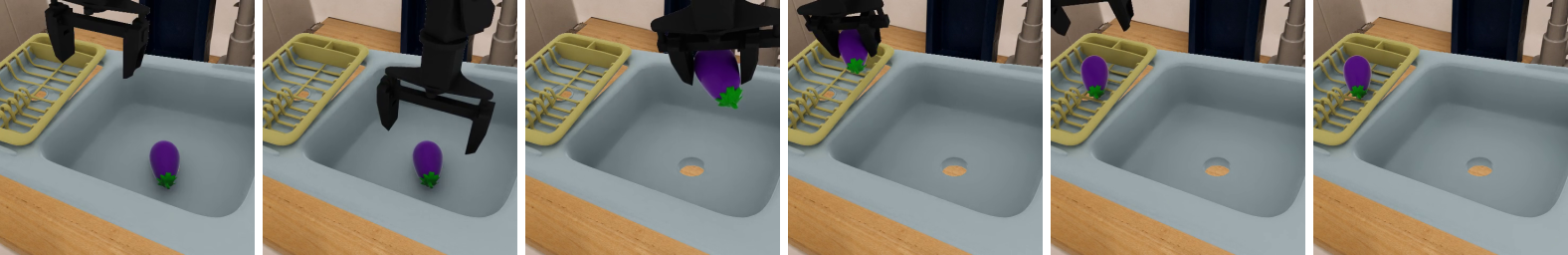}\\[1pt]
    {\footnotesize ``put eggplant into yellow basket''}
}\par\vspace{6pt}
\noindent\parbox{\linewidth}{\centering
    \includegraphics[width=\linewidth]{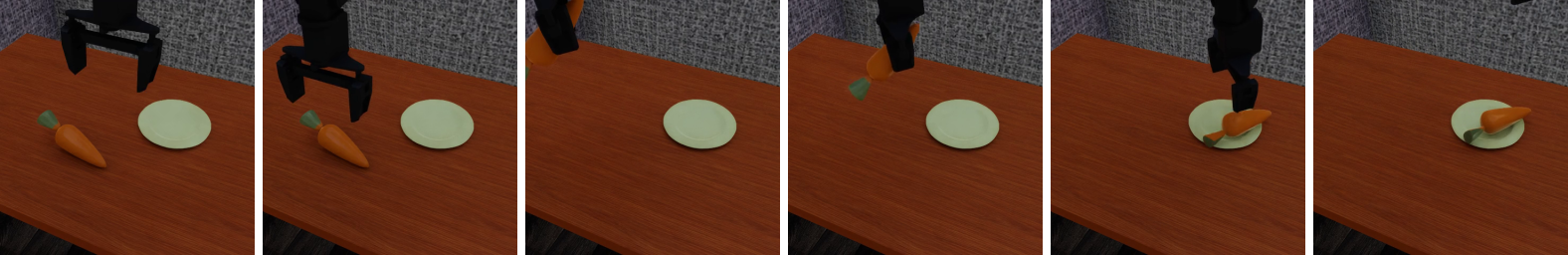}\\[1pt]
    {\footnotesize ``put carrot on plate''}
}\par\vspace{6pt}
\noindent\parbox{\linewidth}{\centering
    \includegraphics[width=\linewidth]{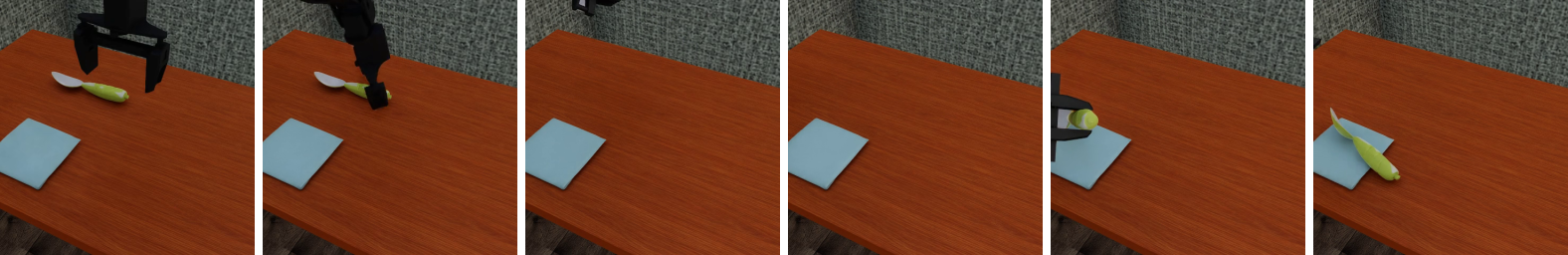}\\[1pt]
    {\footnotesize ``put the spoon on the towel''}
}\par\vspace{6pt}
\noindent\parbox{\linewidth}{\centering
    \includegraphics[width=\linewidth]{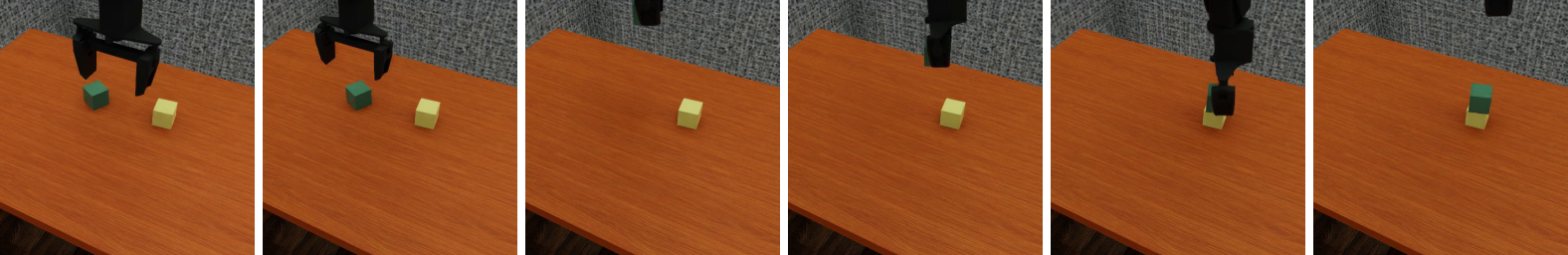}\\[1pt]
    {\footnotesize ``stack the green block on the yellow block''}
}\par\vspace{6pt}
\nopagebreak
\captionof{figure}{Collected trajectories for \textbf{SIMPLER-derived} tasks.}\label{fig:traj_simpler}\par
\vspace{10pt}

\FloatBarrier

\subsection{Qualitative Results for Axis Alignment}
\label{app:qualitative_align}

Fig.~\ref{fig:align_examples} visualizes the predicted axis alignment (App.~\ref{app:axis_align}) on three successful search trajectories.

\noindent\parbox{\linewidth}{\centering
    \includegraphics[width=\linewidth]{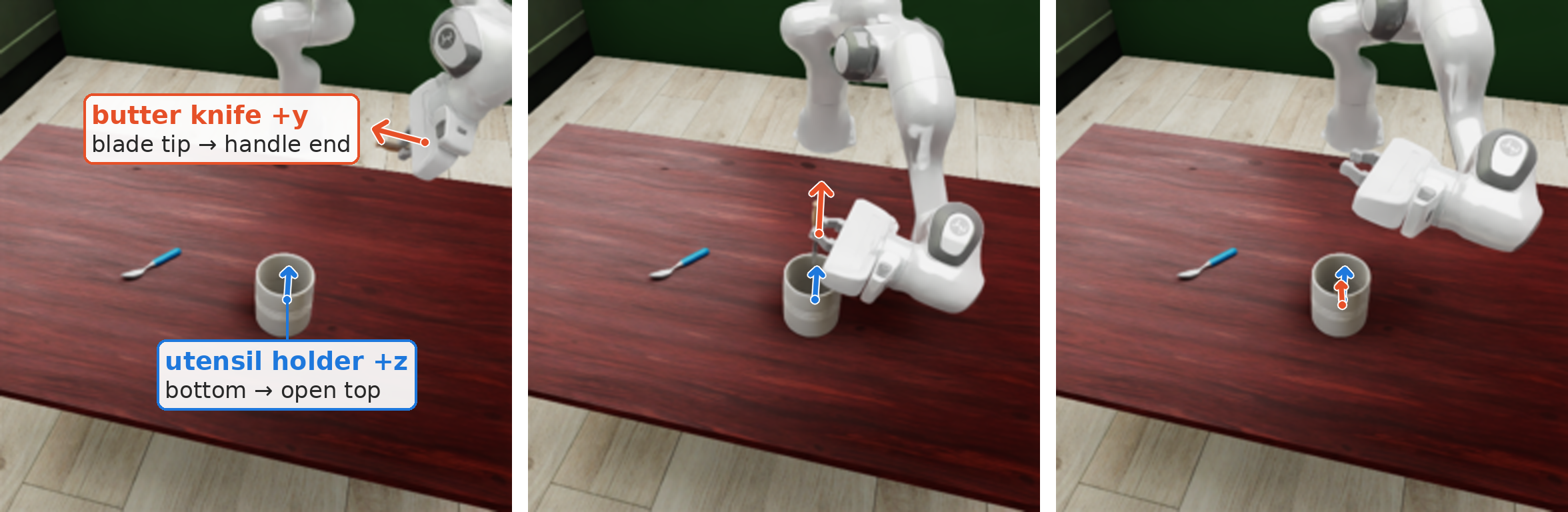}\\[2pt]
    {\footnotesize\raggedright
    \textbf{Subgoal:} ``Place the butter knife into the utensil holder with a vertical orientation.'' \quad \textbf{Predicted alignment:} $(y, z, +)$\\
    The knife is rotated so that its long axis matches the vertical axis of the holder before insertion.\par}
}\par\vspace{8pt}
\noindent\parbox{\linewidth}{\centering
    \includegraphics[width=\linewidth]{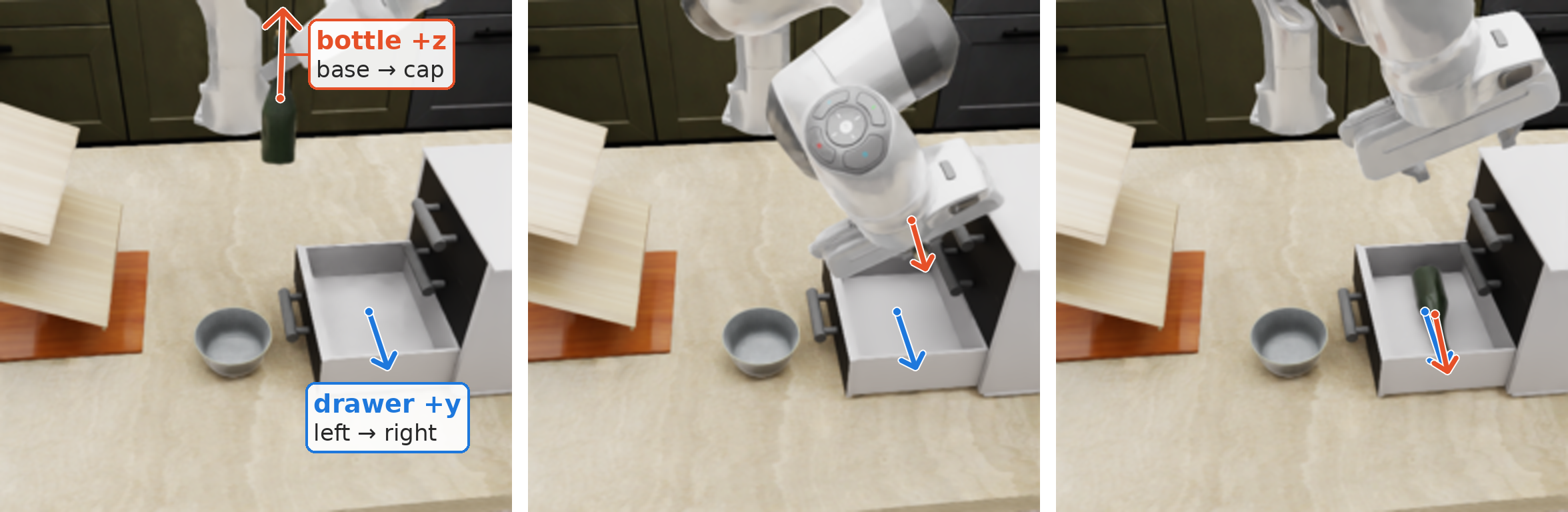}\\[2pt]
    {\footnotesize\raggedright
    \textbf{Subgoal:} ``Place the wine bottle on its side into the bottom drawer of the white cabinet.'' \quad \textbf{Predicted alignment:} $(z, y, +)$\\
    The bottle is laid along the left--right axis of the drawer so that it fits into the shallow drawer.\par}
}\par\vspace{8pt}
\noindent\parbox{\linewidth}{\centering
    \includegraphics[width=\linewidth]{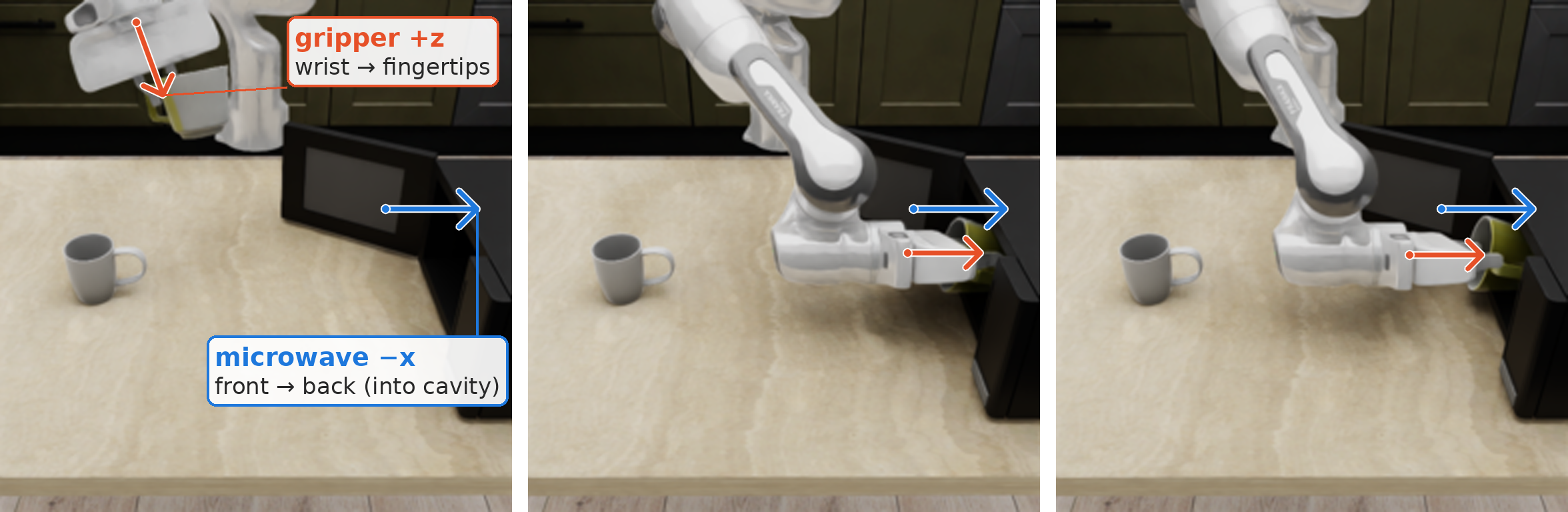}\\[2pt]
    {\footnotesize\raggedright
    \textbf{Subgoal:} ``Place the yellow and white mug into the microwave.'' \quad \textbf{Predicted alignment:} $(z, x, -)$\\
    The gripper reaches in against the opening direction of the microwave, so that the mug is inserted into the cavity from the front.\par}
}\par\vspace{8pt}
\nopagebreak
\captionof{figure}{Predicted axis alignment on successful trajectories. Each row shows the start of the alignment step, the aligned moment, and the end of the step. Red: aligned axis $a_h$ of the held object (top, middle) or of the gripper (bottom); blue: signed target axis $s\,a_t$. Labels in the first panel give the semantic meaning of each aligned axis from the canonical axis annotation.}\label{fig:align_examples}\par
\vspace{10pt}

\subsection{Qualitative Results for Real-world Deployment}
\label{app:qualitative_real}

We show real-world rollouts across our tasks. Each strip shows six frames uniformly sampled from a single trajectory, with the corresponding language instruction below it.
Large-scale simulation pre-training plus target-oriented post-training enables robust zero-shot sim-to-real transfer across various tasks and distribution shifts, including object swaps and background changes (Figs.~\ref{fig:real_orig}--\ref{fig:real_bg}).
We further increase the complexity of the real-world scenes by introducing \emph{diverse, unseen} distractor objects
and by varying the object layout across trials (Fig.~\ref{fig:real_distract}).
The policy still adapts \emph{zero-shot} to these cluttered scenes, grounding the target object among objects never seen during training.

\noindent\parbox{\linewidth}{\centering
    \includegraphics[width=\linewidth]{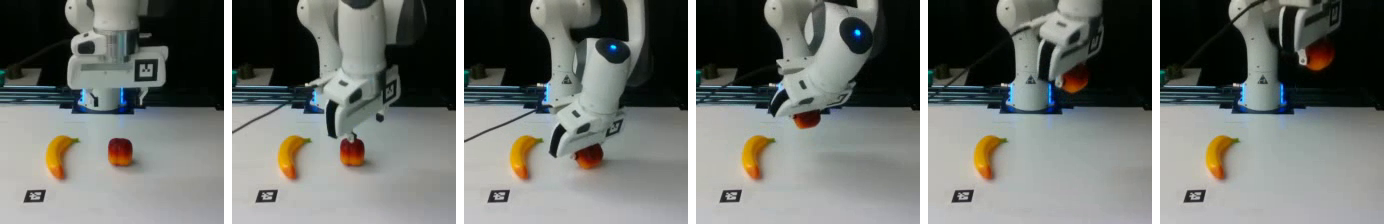}\\[1pt]
    {\footnotesize ``pick up the apple''}
}\par\vspace{6pt}
\noindent\parbox{\linewidth}{\centering
    \includegraphics[width=\linewidth]{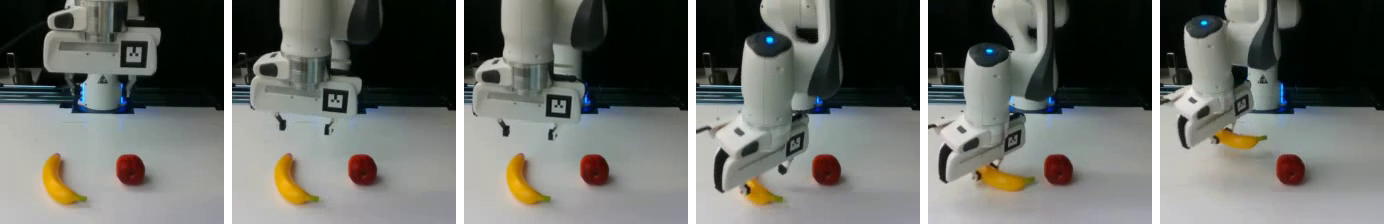}\\[1pt]
    {\footnotesize ``pick up the banana''}
}\par\vspace{6pt}
\noindent\parbox{\linewidth}{\centering
    \includegraphics[width=\linewidth]{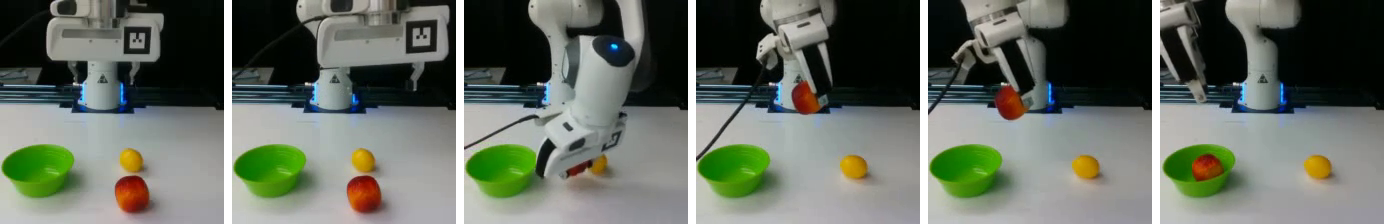}\\[1pt]
    {\footnotesize ``place the apple in the bowl''}
}\par\vspace{6pt}
\noindent\parbox{\linewidth}{\centering
    \includegraphics[width=\linewidth]{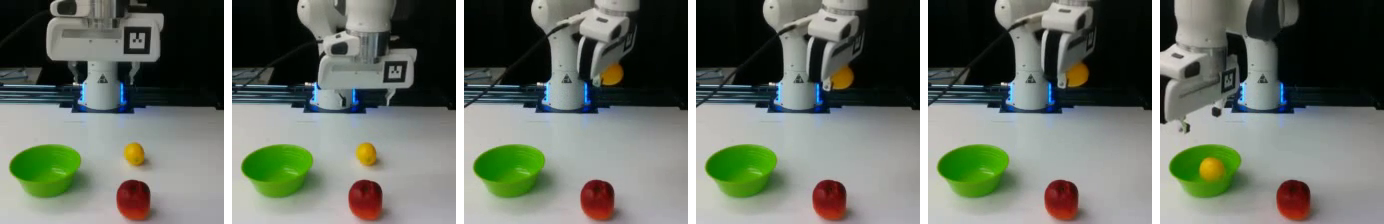}\\[1pt]
    {\footnotesize ``place the lemon in the bowl''}
}\par\vspace{6pt}
\nopagebreak
\captionof{figure}{Real-world rollouts on the original setup.}\label{fig:real_orig}\par
\vspace{10pt}

\noindent\parbox{\linewidth}{\centering
    \includegraphics[width=\linewidth]{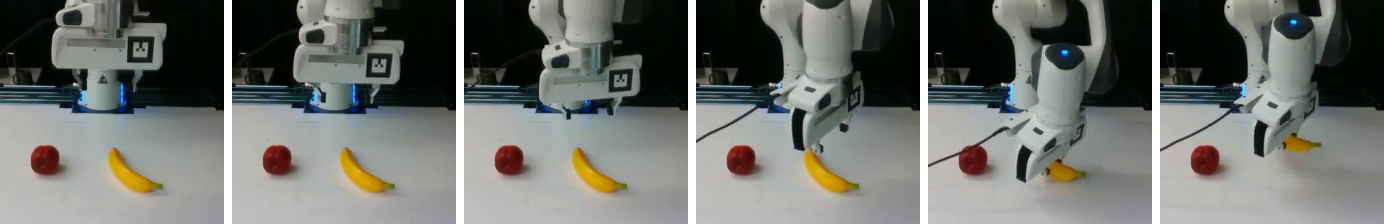}\\[1pt]
    {\footnotesize ``pick up the banana''}
}\par\vspace{6pt}
\noindent\parbox{\linewidth}{\centering
    \includegraphics[width=\linewidth]{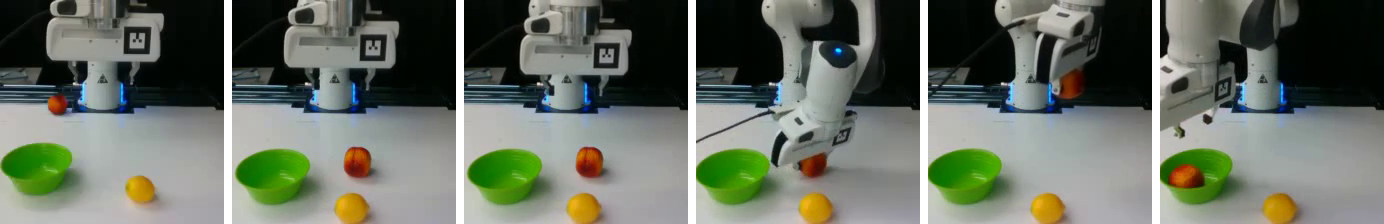}\\[1pt]
    {\footnotesize ``place the apple in the bowl''}
}\par\vspace{6pt}
\nopagebreak
\captionof{figure}{Real-world rollouts under object swaps.}\label{fig:real_swap}\par
\vspace{10pt}

\noindent\parbox{\linewidth}{\centering
    \includegraphics[width=\linewidth]{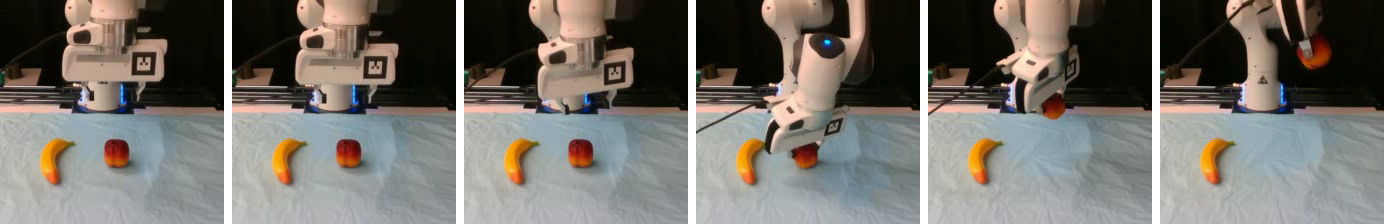}\\[1pt]
    {\footnotesize ``pick up the apple''}
}\par\vspace{6pt}
\noindent\parbox{\linewidth}{\centering
    \includegraphics[width=\linewidth]{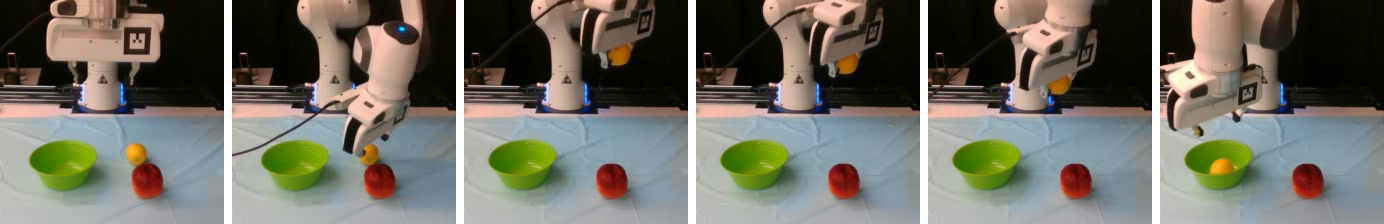}\\[1pt]
    {\footnotesize ``place the lemon in the bowl''}
}\par\vspace{6pt}
\nopagebreak
\captionof{figure}{Real-world rollouts under background changes.}\label{fig:real_bg}\par
\vspace{10pt}

\noindent\parbox{\linewidth}{\centering
    \includegraphics[width=\linewidth]{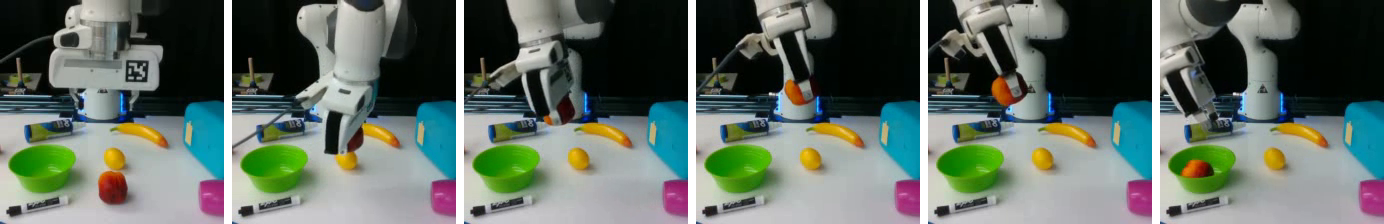}\\[1pt]
    {\footnotesize ``place the apple in the bowl''}
}\par\vspace{6pt}
\noindent\parbox{\linewidth}{\centering
    \includegraphics[width=\linewidth]{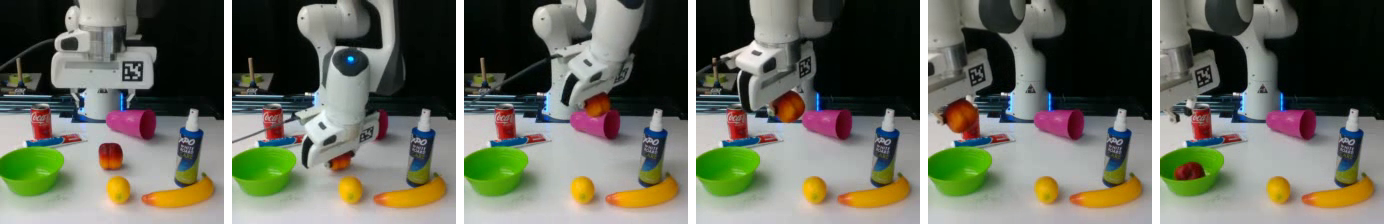}\\[1pt]
    {\footnotesize ``place the apple in the bowl''}
}\par\vspace{6pt}
\noindent\parbox{\linewidth}{\centering
    \includegraphics[width=\linewidth]{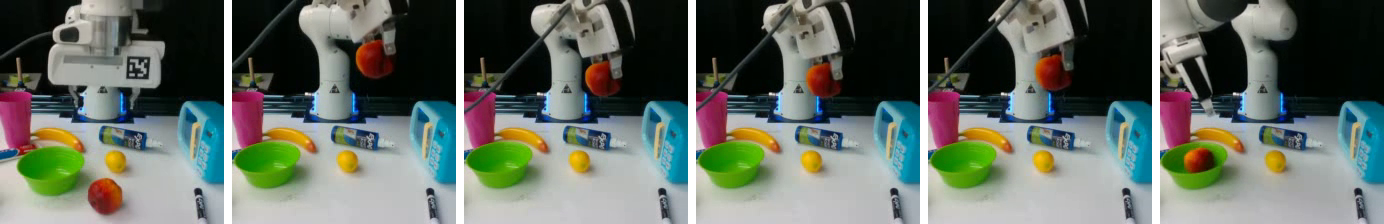}\\[1pt]
    {\footnotesize ``place the apple in the bowl''}
}\par\vspace{6pt}
\noindent\parbox{\linewidth}{\centering
    \includegraphics[width=\linewidth]{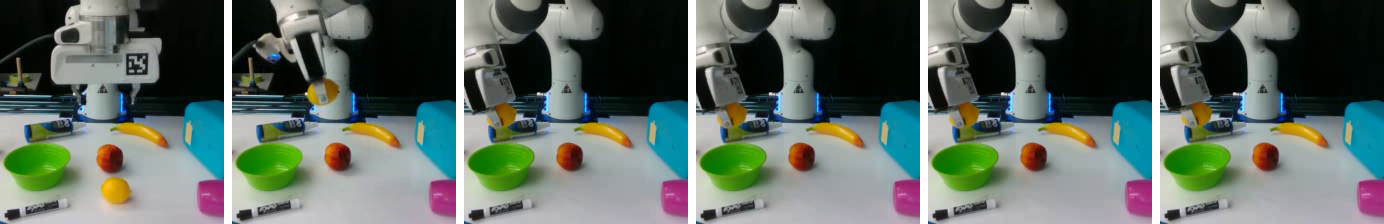}\\[1pt]
    {\footnotesize ``place the lemon in the bowl''}
}\par\vspace{6pt}
\nopagebreak
\captionof{figure}{Real-world rollouts with \emph{diverse, unseen} distractor objects and varied layouts, among objects that never appear in training.}\label{fig:real_distract}\par
\vspace{10pt}

\FloatBarrier

\bibliography{bib/newbib,bib/bib2,bib/refs,bib/act3d,bib/custom,bib/3DDArefs,bib/darpa}

\end{document}